\documentclass[10pt,journal,compsoc]{IEEEtran}
\usepackage{epsfig}
\usepackage{graphicx}
\usepackage{amsmath}
\usepackage{amssymb}
\usepackage{amsthm}
\usepackage[ruled,vlined]{algorithm2e}
\usepackage{colortbl,booktabs}
\usepackage{leftidx}
\usepackage{multirow}
\usepackage{subfigure}

\usepackage{amsmath}
\usepackage[ruled]{algorithm2e}
\usepackage{multirow}
\usepackage{array}
\usepackage{bbding}
\usepackage{epsfig}
\usepackage{graphicx}
\usepackage{subfigure}
\usepackage{epsfig}
\usepackage{amsfonts}       
\usepackage{helvet}
\usepackage{courier}
\usepackage{epsfig}
\usepackage{graphicx}
\usepackage{amsmath,algorithmic}
\usepackage{amssymb,amsthm,amsfonts,threeparttable,caption,tabularx,subfigure,graphicx,epstopdf,booktabs}
\usepackage{dsfont}

\newtheorem{definition}{Definition}[section]
\newtheorem{theorem}{Theorem}[section]

\newtheorem{prop}[theorem]{Proposition}

\newtheorem{pf}{Proof}[section]
\newcommand*{\QEDB}{\hfill\ensuremath{\square}}  

\usepackage[breaklinks=true,bookmarks=false]{hyperref}
\ifCLASSOPTIONcompsoc
  \usepackage[nocompress]{cite}
\else
  \usepackage{cite}
\fi
\ifCLASSINFOpdf
\else
\fi
\begin{document}
%
\title{Deep Partial Multi-View Learning}
%
%
%
%

\author{Changqing~Zhang,
		Yajie~Cui,
        Zongbo~Han,
        Joey Tianyi~Zhou,\\
        Huazhu Fu and
        Qinghua Hu
\IEEEcompsocitemizethanks{\IEEEcompsocthanksitem C. Zhang Y. Cui, Z. Han and Q. Hu are with the College of Intelligence and Computing, Tianjin University, Tianjin 300350, China (e-mail: zhangchangqing@tju.edu.cn; cuiyajie@tju.edu.cn; zongbo@tju.edu.cn; huqinghua@tju.edu.cn).\protect


\IEEEcompsocthanksitem J. T. Zhou is with the Institute of High Performance Computing (IHPC), A*STAR, Singapore (e-mail: zhouty@ihpc.a-star.edu.sg) (corresponding author).

\IEEEcompsocthanksitem H. Fu is with the Inception Institute of Artificial Intelligence, Abu Dhabi, United Arab Emirates (e-mail: hzfu@ieee.org).
}

}

%
%

\markboth{IEEE Transactions on Pattern Analysis and Machine Intelligence}%
{Shell \MakeLowercase{\textit{et al.}}: Bare Demo of IEEEtran.cls for Computer Society Journals}
%



\IEEEtitleabstractindextext{%
\begin{abstract}
Although multi-view learning has made significant progress over the past few decades, it is still challenging due to the difficulty in modeling complex correlations among different views, especially under the context of view missing. To address the challenge, we propose a novel framework termed Cross Partial Multi-View Networks (CPM-Nets), which aims to fully and flexibly take advantage of  multiple partial views. We first provide a formal definition of completeness and versatility for multi-view representation and then theoretically prove the versatility of the learned latent representations. For completeness, the task of learning latent multi-view representation is specifically translated to a degradation process by mimicking data transmission, such that the optimal tradeoff between consistency and complementarity across different views can be implicitly achieved. Equipped with adversarial strategy, our model stably imputes missing views, encoding information from all views for each sample to be encoded into latent representation to further enhance the completeness. Furthermore, a nonparametric classification loss is introduced to produce structured representations and prevent overfitting, which endows the algorithm with promising generalization under view-missing cases. Extensive experimental results validate the effectiveness of our algorithm over existing state of the arts for classification, representation learning and data imputation.
\end{abstract}
}

\maketitle
\IEEEdisplaynontitleabstractindextext


%
\IEEEpeerreviewmaketitle

\section{Introduction}
In real-word applications, data is usually represented with different views, including multiple modalities or various types of features.
Several studies \cite{Baltrusaitis2019PAMI,xu2013survey,dhillon2011multi} have empirically demonstrated that different views can complement each other, leading to ultimate performance improvement. Unfortunately, the unknown and complex correlations among different views often disrupt the integration of different modalities in the model. Moreover, data with missing views further aggravates the modeling difficulty. Conventional multi-view learning usually holds the assumption that each sample is associated with a unified set of observed views and that all views are available for each sample. However, in practical applications, there are usually incomplete cases for multi-view data \cite{li2014partial,tran2017missing,liu2018late,liu2016diagnosis,trivedi2010multiview}. For example, in medical applications, different types of examinations are usually conducted for different subjects, and in Web analysis, some websites may contain texts, pictures and videos, but others may only contain one or two types, which produces data with missing views. The view-missing patterns (\emph{i.e}., combinations of available views) become even more complex for the data with more views.

Projecting different views into a common space (\emph{e.g.}, CCA: Canonical Correlation Analysis and its variants \cite{hotelling1936relations,akaho2006kernel,andrew2013deep}) is impeded by view-missing issue. Several methods have been proposed to keep on exploiting the correlations of different views. One straightforward strategy is to complete the missing views, after which off-the-shelf multi-view learning algorithms can be adopted. The missing views are basically  blockwise and thus low-rank based completion \cite{cai2010singular,mazumder2010spectral} is not applicable, as has been widely recognized in \cite{tran2017missing,cai2018deep}. Missing modality imputation methods \cite{ngiam2011multimodal,tran2017missing} usually require samples with two paired modalities to train the networks which can predict the missing modality from the observed one. To explore the complementarity among multiple views, another natural strategy is to manually group samples according to the availability of data sources \cite{yuan2012multi}, and to subsequently learn multiple models on these groups for late fusion. Although it is more effective than learning on each individual view, the grouping strategy is not flexible, especially for data with a large number of views. 

\begin{figure*}[!ht]
	\centering
		\begin{minipage}[t]{0.9\linewidth}
			\centering
			\includegraphics[height=2.3in,width=0.8\linewidth]{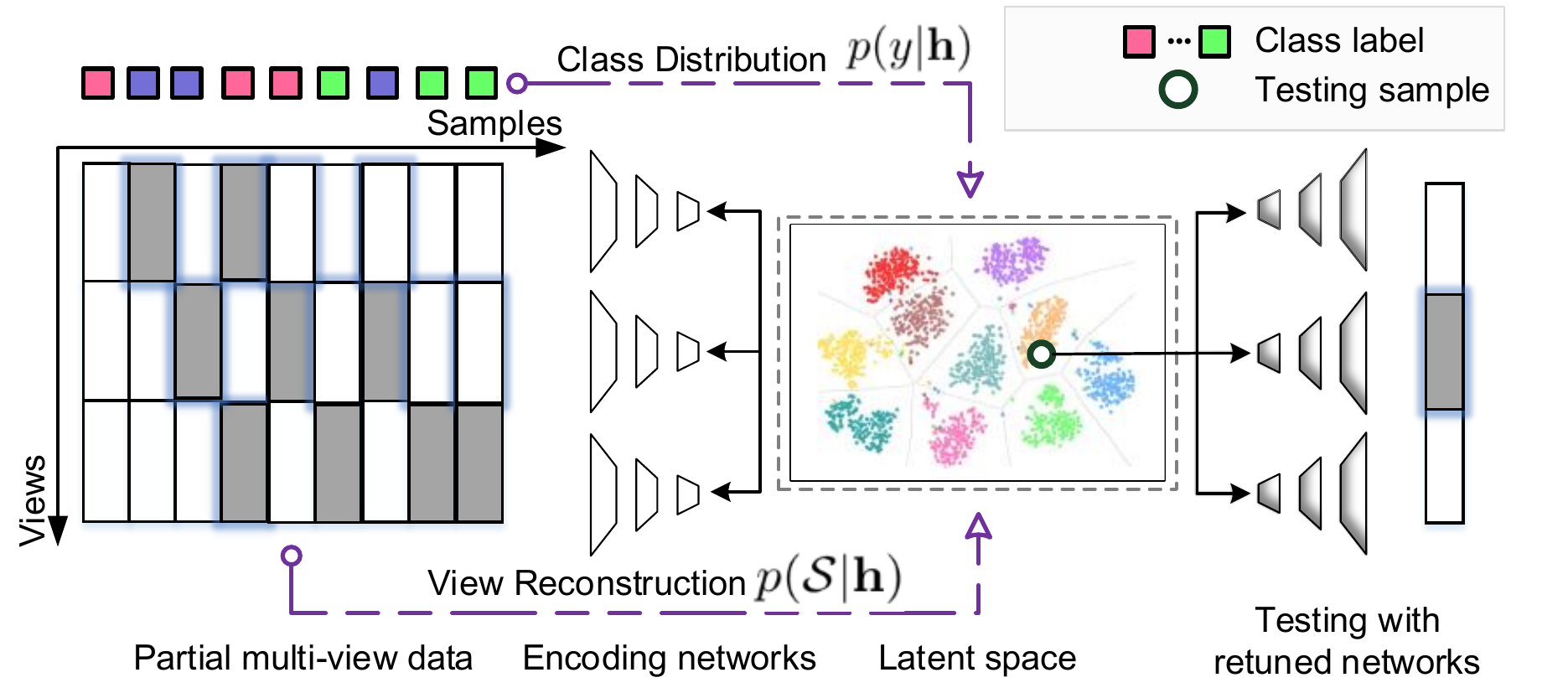}
	\end{minipage}
	\caption{Illustration of Cross Partial Multi-View Networks for classification. Given multi-view data with missing
views (black blocks), the encoding \protect\footnotemark networks degrade the complete latent representation into the available views (white blocks). Learning multi-view representation according to the distributions of observations and classes has the promising potential to encode complementary information, as well as provides accurate predictions.}
	\label{fig:framework}
\end{figure*}

Therefore, it is more challenging and important to endow the algorithm with high effectiveness and adaptive capacity under complex view-missing cases. In this paper, we advocate focusing on \textit{completeness and flexibility} for partial multi-view representation learning to fully exploit multiple views. We propose a novel deep partial multi-view learning algorithm, \emph{i.e.}, Cross Partial Multi-View Networks (CPM-Nets), as shown in Fig.~\ref{fig:framework}. The model comprehensively encodes information from different views into a latent representation which is versatile with theoretical guarantee compared to each single view and adaptive to complex correlations among different views. The model is \textit{flexible} for arbitrary view-missing patterns with latent representations and the \textit{completeness} is further enhanced with adversarial strategy. For classification, we introduce a non-parametric loss to learn clustering-structured representations introducing simple bias for generalization especially important for the partial view setting. Specifically, benefiting from the learned common latent representation from the encoding networks, all samples and all views can be jointly exploited regardless of view-missing patterns. For the multi-view representation, CPM-Nets jointly considers completeness and structure, making them mutually improve each other to obtain the representation reflecting the underlying patterns. Theoretical analysis and empirical results validate the effectiveness of the proposed CPM-Nets in exploiting partial multi-view data. The contributions of this paper are summarized as follows:
\begin{itemize}
\item We propose a novel framework - CPM-Nets - to conduct partial multi-view learning, which jointly considers completeness and structure to learn a unified latent representation, endowing the algorithm with high flexibility and generalization for partial multi-view data.
\item The latent representation encoded from observations is complete and versatile, and thus enhances the prediction performance, while the clustering-like classification schema in turn enhances the separability of the latent representation. Theoretical analysis and empirical results in classification validate the effectiveness of the proposed CPM-Nets in exploiting partial multi-view data.
\item For unsupervised learning, using degradation strategy, the information from observed views is flexibly encoded into the learned representation. Meanwhile, benefiting from the adversarial strategy, the imputation is further stabilized, which in turn improves the latent representation.
\item Extensive experiments on diverse multi-view data validate that the proposed CPM-Nets can improve the unified representation, classification and data imputation compared with the current state-of-the-art results.
\end{itemize}

\subsection{Related Work}
\textbf{Multi-View Learning (MVL)} aims to jointly utilize information from different views. Multi-view clustering algorithms \cite{kumar2011coReg,kumar2011coTrain,zhang2015low,zhang2017latent,yang2019split,BDR2019,GongHW19} usually search for consistent clustering hypotheses across different views, where the representative methods include co-regularization based \cite{kumar2011coReg}, co-training based \cite{kumar2011coTrain} and high-order multi-view clustering \cite{zhang2015low}. Under the metric learning framework, multi-view classification methods \cite{zhang2011heterogeneous,zhang2017hierarchical} jointly learn multiple metrics for multiple views. The representative multi-view representation learning methods are CCA based, including kernelized CCA \cite{akaho2006kernel}, deep neural network based CCA \cite{andrew2013deep,Wang2016On}, DVCCA \cite{WangLL16a} and semi-paired and semi-supervised generalized correlation analysis (S$^2$GCA) \cite{chen2012unified}. DVCCA has the advantages in learning nonlinear correlations and disentanglement of variables instead of handling complex missing patterns.

\textbf{Cross-View Learning (CVL)} essentially searches for mappings between two views, where the similarity between the samples from different modalities can be measured directly. It has been widely applied in real applications \cite{rasiwasia2010new,chung2018unsupervised,castrejon2016learning,zhou2019multi,zhou2019deep,street2019}. With adversarial training, the embedding spaces of two individual views are learned and aligned simultaneously \cite{chung2018unsupervised}. The cross-modal convolutional neural networks are regularized to obtain a shared representation which is agnostic to the modality for cross-modal scene images \cite{castrejon2016learning}. In \cite{street2019}, an analogical embedding space is learned for both views and then information is distilled across views. Meanwhile the cross-view learning can be also utilized for missing view imputation \cite{shang2017vigan,cai2018deep}.

\textbf{Learning on Incomplete Data.} A straightforward method of learning on incomplete data is to conduct data imputation and then existing models (e.g., clustering/classification) can be applied. The representative global imputation algorithms include SVD imputation (SVDimpute) \cite{SVD2001} and Bayesian principal component analysis (BPCA) \cite{BPCA2003}. In contrast to global strategy, local imputation algorithms exploit local similarity structure in the dataset for missing value imputation \cite{KNNimpute2001,LSimpute2004,LLSimpute2006}. The hybrid strategy captures both global and local correlation information \cite{LinCmb2005}.  Recently proposed imputation methods \cite{tran2017missing,shang2017vigan} complete the missing views by leveraging the power of deep neural networks. There are also a few algorithms which can directly conduct learning without imputation. The grouping strategy \cite{yuan2012multi} divides all samples according to the availability of data sources, and then multiple classifiers are learned for late fusion. This strategy cannot scale well for data with a large number of views or small-sample-size cases. Another line \cite{XueDDRL19} incorporates semi-supervised deep matrix factorization, correlated subspace learning, and multi-view label prediction into a unified framework. Further, \cite{YangZS018} learns the classifiers by leveraging the intrinsic view consistency and extrinsic unlabeled information. 

\section{Cross Partial Multi-View Networks}
In this work, we focus on classification based on data with missing views termed \emph{Partial Multi-View Classification} (see definition \ref{def:cpmvc}), where samples with different view-missing patterns are involved. The proposed \emph{cross partial multi-view networks} enable the comparability for samples with different combinations of views rather than only  two views, which generalizes the concept of cross-view learning. There are three main challenges for partial multi-view representation learning: (1) project samples with arbitrary view-missing patterns (flexibility) into a comprehensive latent space (completeness) for comparability (in section~\ref{sec:complete-representation}); (2) make the learned representation reflect class distribution (structured representation) for separability (in section~\ref{sec:classification}); (3) reduce the gap between representations obtained in test stage and training stage for consistency (in section~\ref{sec:testing}). For clarification, we first provide the formal definition of partial multi-view classification as follows:
\begin{definition} (\textbf{Partial Multi-View Classification (PMVC)})~
\label{def:cpmvc}
     Given the training set $\{\mathcal{S}_n,y_n\}_{n=1}^N$, where $\mathcal{S}_n$ is a subset of the complete observations $\mathcal{X}_n=\{\mathbf{x}_n^{(v)}\}_{v=1}^V$ (i.e., $\mathcal{S}\subseteq \mathcal{X}$) and $y_n$ is the class label with $N$ and $V$ being the number of samples and views, respectively, PMVC trains a classifier by using training data containing view-missing samples, to classify a new instance $\mathcal{S}$ with an arbitrary view-missing pattern.
 \end{definition}

\subsection{Multi-View Complete Representation}
\label{sec:complete-representation}

Considering the first challenge, where the desired latent representation should encode the information from observed views, we provide the definition of completeness for multi-view representation (inspired by the reconstruction point of view \cite{lee1996image}) as follows:
\begin{definition} (\textbf{Completeness for Multi-View Representation})~
\label{def:completeness}
     A multi-view representation $\mathbf{h}$ is complete if each observation, {i.e.}, $\mathbf{x}^{(v)}$ from  $\{\mathbf{x}^{(1)},...,\mathbf{x}^{(V)}\}$, can be reconstructed from a mapping $f_v(\cdot)$, i.e., $\mathbf{x}^{(v)} = f_v(\mathbf{h})$.
\end{definition}
Intuitively, we can reconstruct each view from a complete representation in a numerically stable way. Furthermore, we show that the completeness is achieved under the assumption that each view is conditionally independent given the shared multi-view representation \cite{white2012convex}. Similar to each view from $\mathcal{X}$, the class label $y$ can also be considered as one (semantic) view, in which case we have
\begin{equation}
\begin{aligned}
p(y,\mathcal{S}|\mathbf{h})= p(y|\mathbf{h})p(\mathcal{S}|\mathbf{h}),
\label{equ:self-sub}
\end{aligned}
\end{equation}
where $p(\mathcal{S}|\mathbf{h}) = p(\mathbf{x}^{(1)}|\mathbf{h})p(\mathbf{x}^{(2)}|\mathbf{h})...p(\mathbf{x}^{(V)}|\mathbf{h})$. We can obtain the common representation by maximizing $p(y,\mathcal{S}|\mathbf{h})$.

\footnotetext{Since we employ a backward encoding strategy which encodes the information from multiple views into latent representations, it can be considered as an ‘encoding’ process although it is different from the well-known encoder-decoder structure.}

Based on multiple views in $\mathcal{S}$, we model the likelihood with respect to $\mathbf{h}$ given observations $\mathcal{S}$ as
\begin{equation}
\begin{aligned}
p(\mathcal{S}|\mathbf{h})\propto e^{-\Delta(\mathcal{S},f(\mathbf{h};\boldsymbol{\Theta}_r))},
\label{equ:self-sub}
\end{aligned}
\end{equation}
where $\boldsymbol{\Theta}_r$ are parameters governing the reconstruction mapping $f(\cdot)$ from common representation $\mathbf{h}$ to partial observations $\mathcal{S}$, with $\Delta(\mathcal{S},f(\mathbf{h};\boldsymbol{\Theta}_r))$ being the reconstruction loss. From the view of class label, we model the likelihood with respect to $\mathbf{h}$ given class label $y$ as
\begin{equation}
\begin{aligned}
p(y|\mathbf{h})\propto e^{-\Delta(y,g(\mathbf{h};\boldsymbol{\Theta}_c))},
\label{equ:self-sub}
\end{aligned}
\end{equation}
where $\boldsymbol{\Theta}_c$ are parameters of the classification function $g(\cdot)$ based on $\mathbf{h}$, and $\Delta(y,g(\mathbf{h};\boldsymbol{\Theta}_c))$ defines the classification loss.
Accordingly, assuming data are independent and identically
distributed (IID), the log-likelihood function is
\begin{equation}
\begin{aligned}
&\mathcal{L}(\{\mathbf{h}_n\}_{n=1}^N,\boldsymbol{\Theta}_r,\boldsymbol{\Theta}_c) = \sum_{n=1}^N\text{ln}~p(y_n, \mathcal{S}_n|\mathbf{h}_n) \\
&\propto -\big(\sum_{n=1}^N\Delta(\mathcal{S}_n,f(\mathbf{h}_n;\boldsymbol{\Theta}_r)) + \Delta(y_n,g(\mathbf{h}_n;\boldsymbol{\Theta}_c))\big),
\label{equ:self-sub}
\end{aligned}
\end{equation}
where $\mathcal{S}_n$ denotes the available views for the $n$th sample. On one hand, we encode the information from available views into a latent representation $\mathbf{h}_n$ and denote the encoding loss as $\Delta(\mathcal{S}_n,f(\mathbf{h}_n;\boldsymbol{\Theta}_r))$. On the other hand, the learned representation should be consistent with the class distribution, which is implemented by minimizing the loss $\Delta(y_n,g(\mathbf{h}_n;\boldsymbol{\Theta}_c))$ to penalize any disagreement with the class label.

Effectively encoding information from different views is the key requirement for multi-view representation, and thus we seek a common representation that can recover the partial (available) observations. Accordingly, the following loss is induced
\begin{equation}
\begin{aligned}
&\Delta(\mathcal{S}_n,f(\mathbf{h}_n;\boldsymbol{\Theta}_r)) = \ell_r(\mathcal{S}_n,\mathbf{h}_n) \\&
= \sum_{v=1}^V s_{nv}||f_v(\mathbf{h}_n; \boldsymbol{\Theta}^{(v)}_r)-\mathbf{x}_n^{(v)}||^2,
\label{equ:reconstruction}
\end{aligned}
\end{equation}
where $\Delta(\mathcal{S}_n,f(\mathbf{h}_n;\boldsymbol{\Theta}_r))$ is implemented with the reconstruction loss $\ell_r(\mathcal{S}_n,\mathbf{h}_n)$. $s_{nv}$ indicates the availability of the $n$th sample in the $v$th view, \emph{i.e.}, $s_{nv}=1$ and $0$ indicate available and unavailable views, respectively. $f_v(\cdot; \boldsymbol{\Theta}^{(v)}_r)$ is the reconstruction network for the $v$th view parameterized by $\boldsymbol{\Theta}^{(v)}_r$. In this way, $\mathbf{h}_n$ encodes  information from available views, and different samples (regardless of their missing patterns) are associated with representations in a common space, making them comparable.

Ideally, minimizing Eq.~(\ref{equ:reconstruction}) will induce a complete representation. Since the complete representation encodes information from different views, it should be more versatile compared with using any single view. We provide the definition of versatility for multi-view representation as follows:
\begin{definition} (\textbf{Versatility for Multi-View Representation})~
     Given the observations $\mathbf{x}^{(1)},...,\mathbf{x}^{(V)}$ from $V$ views, the multi-view representation $\mathbf{h}$ is of versatility if~$\forall~v~\text{and}~\forall~\text{mapping}~\varphi(\cdot)$ with $y^{(v)} = \varphi(\mathbf{x}^{(v)})$, there exists a mapping ~$\psi(\cdot)$ satisfying $y^{(v)} = \psi(\mathbf{h})$, where $\mathbf{h}$ is the corresponding multi-view representation for sample $\mathcal{S} = \{\mathbf{x}^{(1)},...,\mathbf{x}^{(V)}\}$.
 \end{definition}
Accordingly, we have the following theoretical result:
\begin{prop} (\textbf{Versatility for the Multi-View Representation from Eq.~(\ref{equ:reconstruction})})~
     There exists a solution (with respect to latent representation $\mathbf{h}$) to Eq. (\ref{equ:reconstruction}) which holds the versatility.
     \label{prop:completeness}
 \end{prop}

 \begin{pf}
The proof for proposition \ref{prop:completeness} is as follows. Ideally, according to Eq.~(\ref{equ:reconstruction}), $\forall~v$, there exists $\mathbf{x}^{(v)}=f_v(\mathbf{h}; \boldsymbol{\Theta}^{(v)}_r)$, where $f_v(\cdot)$ is the mapping from $\mathbf{h}$ to $\mathbf{x}^{(v)}$. Hence, $\forall~\varphi(\cdot)$ with $y^{(v)} = \varphi(\mathbf{x}^{(v)})$, there exists a mapping $\psi(\cdot)$ satisfying $y^{(v)} = \psi(\mathbf{h})$ by defining $\psi(\cdot) = \varphi(f_v(\cdot))$. This proves the versatility of the latent representation $\mathbf{h}$ based on multi-view observations $\{\mathbf{x}^{(1)},...,\mathbf{x}^{(V)}\}$.

{In practical cases, it is usually difficult to guarantee the exact versatility for latent representation, so the goal is to minimize the error $e_y=\sum_{v=1}^V ||\psi(\mathbf{h})-\varphi(\mathbf{x}^{(v)})||^2$ (\emph{i.e.}, $\sum_{v=1}^V ||\varphi(f_v(\mathbf{h;\boldsymbol{\Theta}}^{(v)}))-\varphi(\mathbf{x}^{(v)})||^2$) which is inversely proportional to the degree of versatility. Fortunately, it is easy to show that $Ke_r$ with $e_r = \sum_{v=1}^V ||f_v(\mathbf{h}; \boldsymbol{\Theta}^{(v)}_r)-\mathbf{x}^{(v)}||^2$ from Eq.~(\ref{equ:reconstruction}) is the upper bound of $e_y$ if $\varphi(\cdot)$ is Lipschitz continuous, with $K$ being the Lipschitz constant.}
\QEDB
\end{pf}
Although the proof is inferred under the condition that all views are available, it is intuitive and easy to generalize the results for view-missing cases.

\subsection{Classification on Structured Latent Representation}
\label{sec:classification}
Multiclass classification remains challenging due to possible confusing classes \cite{liu2017easy}. For the second challenge, we thus aim to ensure that the learned representation is structured for separability, using a clustering-like loss. Specifically, we should minimize the following classification loss
\begin{equation}
\begin{aligned}
\Delta({y}_n,y)  = \Delta(y_n, g(\mathbf{h}_n;\boldsymbol{\Theta}_c)),
\label{equ:self-sub}
\end{aligned}
\end{equation}
where $g(\mathbf{h}_n;\boldsymbol{\Theta}_c) = {\arg\max}_{y\in \mathcal{Y}}\mathbb{E}_{\mathbf{h} \sim \mathcal{T}(y)}F(\mathbf{h},\mathbf{h}_n)$ and $F(\mathbf{h},\mathbf{h}_n) = \phi(\mathbf{h};\boldsymbol{\Theta}_c)^T\phi(\mathbf{h}_n;\boldsymbol{\Theta}_c)$, with $\phi(\cdot;\boldsymbol{\Theta}_c)$ being the feature mapping function for $\mathbf{h}$, and $\mathcal{T}(y)$ being the set of latent representation from class $y$. In our implementation, we set $\phi(\mathbf{h};\boldsymbol{\Theta}_c) = \mathbf{h}$ for simplicity and effectiveness. By jointly considering classification and representation learning, the misclassification loss is specified as
\begin{equation}
\begin{aligned}
\ell_c({y}_n,y,\mathbf{h}_n) = \max \bigg(0, &\Delta(y_n, y)+\mathbb{E}_{\mathbf{h} \sim \mathcal{T}(y)} F(\mathbf{h},\mathbf{h}_n)\\
&-\mathbb{E}_{\mathbf{h} \sim \mathcal{T}(y_n)} F(\mathbf{h},\mathbf{h}_n) \bigg).
\label{equ:self-sub}
\end{aligned}
\end{equation}
Compared with the cross entropy loss, which is most commonly used in parametric classification, the clustering-like loss not only penalizes the misclassification but also guarantees a structured representation. $\Delta(y_n, y)$ represents a part of loss and a conditional margin between corrected and incorrected classifications as well. Specifically, for a correctly classified sample (\emph{i.e.}, $y = y_n$), we have $\Delta(y_n, y)$ = 0. For an incorrectly classified sample (\emph{i.e.}, $y\neq y_n$), we have $\Delta(y_n, y)$ = 1, indicating the similarity between $\mathbf{h}_n$ and the centroid corresponding to class $y_n$ is larger than that between $\mathbf{h}_n$ and the centroid corresponding to class $y$ (wrong label) by a margin of $\Delta(y_n, y)$. In this way, $\Delta(y_n, y)$ acts as the indicator of correct classification, as well as the margin between correct and incorrect classifications. Hence, the proposed non-parametric loss naturally leads to a representation with clustering structure.

Based on the above considerations, the overall objective function is induced as
\begin{equation}
\begin{aligned}
{\min}_{\{\mathbf{h}_n\}_{n=1}^{N},\boldsymbol{\Theta}_r}\frac{1}{N}\sum_{n=1}^N\ell_r(\mathcal{S}_n,\mathbf{h}_n;\boldsymbol{\Theta}_r) + \lambda \ell_c({y}_n,y,\mathbf{h}_n),
\label{equ:self-sub}
\end{aligned}
\end{equation}
where $\lambda >0$ balances the degree of completeness from multiple views and structure according to class labels. 

\subsection{Test: Towards Consistency with Training Stage}
\label{sec:testing}
The last challenge lies in narrowing the gap between training and test stages in representation learning. To classify a test sample with incomplete views $\mathcal{S}$, a straightforward way is to optimize the objective, $\min_{\mathbf{h}}\ell_r(\mathcal{S},f(\mathbf{h}; \boldsymbol{\Theta}_{r}))$, to encode the information from $\mathcal{S}$ into $\mathbf{h}$. However, in this way, the gap originates from the difference between the objectives corresponding to training and test stages. To address this issue, we introduce a re-tuning strategy. In training stage, we obtain $\mathbf{h}_n$ which encodes information from partial multi-view features $\mathcal{S}_n$. As for fine-tuning procedure, we optimize equation (5) (instead of the overall objective function (8)) with $\{\mathcal{S}_n, \mathbf{h}_n\}_{n=1}^N$, and accordingly the networks $\{f_v(\mathbf{h}; \boldsymbol{\Theta}^{(v)}_r)\}_{v=1}^V$ can be retuned into $\{f_v^{'}(\mathbf{h}; \boldsymbol{\Theta}^{(v)}_{rt})\}_{v=1}^V$. In this way, $\{f_v^{'}(\mathbf{h}; \boldsymbol{\Theta}^{(v)}_{rt})\}_{v=1}^V$ ensures the  consistency of obtaining latent representations between training and testing stages. Subsequently, in test stage, we can solve the following objective - $\min_{\mathbf{h}}\ell_r(\mathcal{S},f^{'}(\mathbf{h}; \boldsymbol{\Theta}_{rt}))$ to obtain the latent representations which are consistent with those in training stage. The optimization of the proposed CPM-Nets and the test procedure are summarized in Algorithm~\ref{alg:alg1}.

\begin{algorithm}[t]
\SetAlgoLined
\caption{Algorithm for CPM-Nets}
\textbf{/*Training*/}\\
\KwIn{Partial multi-view dataset: $\mathcal{D} = \{\mathcal{S}_n, y_n\}_{n=1}^N$, hyperparameter $\lambda$.}
\textbf{Initialize:} {Initialize $\{\mathbf{h}_n\}_{n=1}^N$ and $\{\boldsymbol{\Theta}_r^{(v)}\}_{v=1}^V$ with random values}.\\
\While{not converged}{
\For{$v=1:V$ }
{
Update the network parameters

$\mathbf{\Theta}_r^{(v)}$ with gradient descent: \\
 \begin{small}
$\boldsymbol{\Theta}_r^{(v)} \leftarrow \boldsymbol{\Theta}_r^{(v)} - \alpha   \partial \frac{1}{N}\sum_{n=1}^N \ell_r(\mathcal{S}_n,\mathbf{h}_n;\boldsymbol{\Theta}_r)/\partial \boldsymbol{\Theta}_r^{(v)}$;
\end{small}
\\
}
\For{$n=1:N$ }
{
Update the latent representation $\mathbf{h}_n$ with gradient descent: \\
 \begin{small}
$\mathbf{h}_n \leftarrow \mathbf{h}_n-\alpha \partial\frac{1}{N} \sum_{n=1}^N ({\ell_r(\mathcal{S}_n,\mathbf{h}_n;\boldsymbol{\Theta}_r) +
\lambda \ell_c({y}_n,y,\mathbf{h}_n)})/\partial \mathbf{h}_n$;\\
\end{small}
}
}
\KwOut{networks parameters $\{\mathbf{\Theta}_r^{(v)}\}_{v=1}^V$ and latent representation $\{\mathbf{h}_n\}_{n=1}^N$.}
\textbf{/*Test*/}\\
Train the retuned networks ($\{\mathbf{\Theta}_{rt}^{(v)}\}_{v=1}^V$) for test;\\
Calculate the latent representation with the retuned networks for test instance;\\
Classify the test instance with $y = {\arg\max}_{y\in \mathcal{Y}}\mathbb{E}_{\mathbf{h} \sim \mathcal{T}(y)}F(\mathbf{h},\mathbf{h}_{test})$.
\label{alg:alg1}
\end{algorithm}

\begin{figure*}[h]
	\centering
		\begin{minipage}[t]{0.9\linewidth}
			\centering
			\includegraphics[height=2.3in,width=0.8\linewidth]{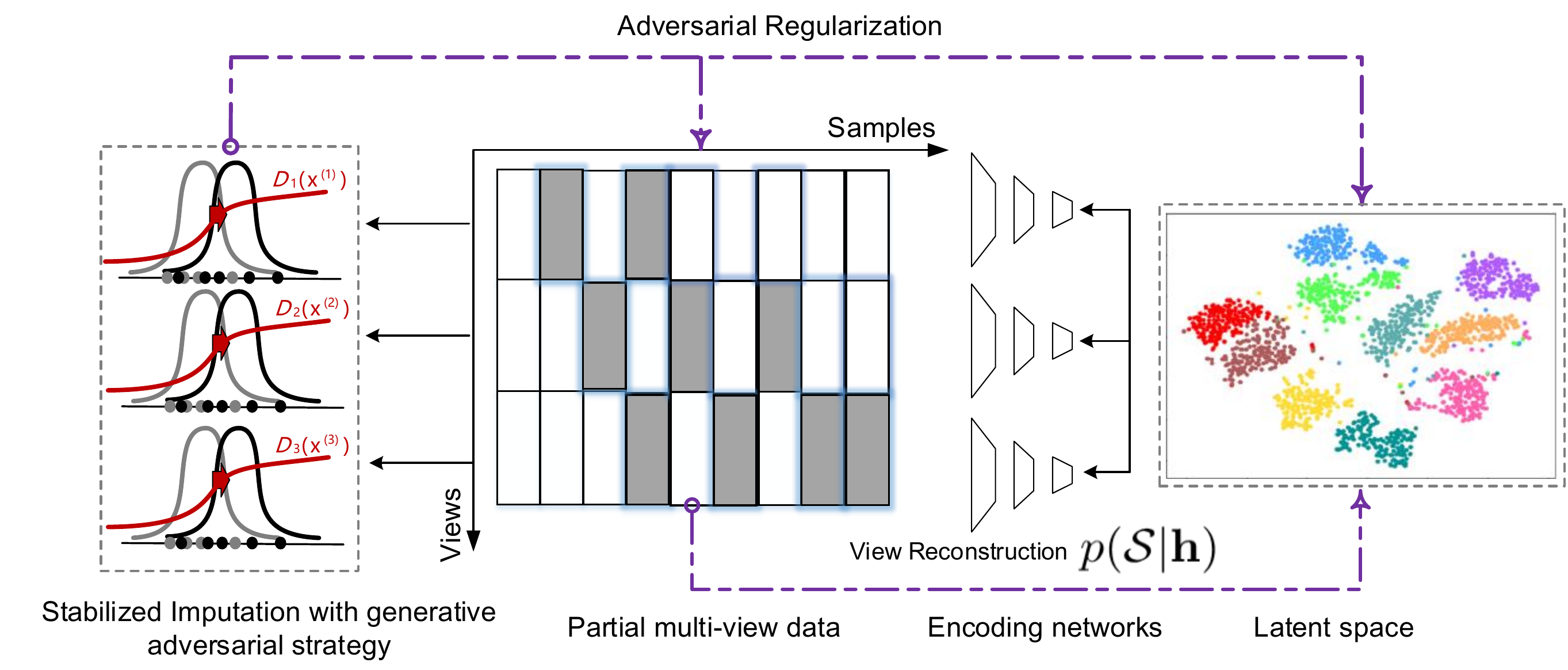}
	\end{minipage}
	\caption{Illustration of CPM-Nets for unsupervised learning. The latent representation learning and missing data imputation equipped with adversarial strategy (CPM-GAN) are jointly conducted to improve each other.}
	\label{fig:framework2}
\end{figure*}
\noindent\textbf{Discussion on Key Components.}~
The CPM-Nets is composed of two key components, \emph{i.e.}, encoding networks and clustering-like classification. As these are different from conventional components, detailed explanations are provided.
\begin{itemize}
\item{\textbf{Encoding schema.} To encode the information from multiple views into a common representation, there is an alternative route, \emph{i.e.}, $\ell_r(\mathcal{S}_n,\mathbf{h}_n) = \sum_{v=1}^V s_{nv}||f(\mathbf{x}_n^{(v)}; \boldsymbol{\Theta}^{(v)})-\mathbf{h}_n||^2$. This is different from the schema used in our model shown in Eq. (\ref{equ:reconstruction}), \emph{i.e.}, $\ell_r(\mathcal{S}_n,\mathbf{h}_n) = \sum_{v=1}^V s_{nv}||f(\mathbf{h}_n; \boldsymbol{\Theta}^{(v)})-\mathbf{x}_n^{(v)}||^2$. The underlying assumption in our model is that information from different views originates from a latent representation $\mathbf{h}$, and hence it can be mapped to each individual view. Whereas for the alternative, it assumes that the latent representation can be obtained from (mapping) each single view, which is often not the case in real applications. Further for the alternative, ideally, minimizing the loss will force the representations of different views to be the same, which is not reasonable especially for highly independent views. 
The theoretical results in subsection \ref{sec:complete-representation} further support this analysis.
}
\item{\textbf{Classification model.} For classification, the widely used strategy is to learn a classification function based on $\mathbf{h}$, \emph{i.e.}, $y=f(\mathbf{h};\boldsymbol{\Theta})$ parameterized with $\boldsymbol{\Theta}$. Compared with this manner, the reasons for using the clustering-like classifier instead in our model are as follows. First, jointly learning the latent representation and parameterized classifier is likely an under-constrained problem which may find representation that can well fit the training data but not well reflect the underlying patterns, and thus the generalization ability may be affected \cite{le2018nips}. Second, the clustering-like classification produces compactness within the same class and separability between different classes for the learned representation, making the classifier interpretable. Finally, the non-parametric strategy reduces the load of parameter tuning and reflects a simpler inductive bias, which is especially beneficial to small-sample-size problems \cite{Jake2017}.}
\end{itemize}

\section{Enhanced Completeness with Adversarial Strategy}
\label{sec:cpm-gam}
Generative adversarial networks (GANs) \cite{GoodfellowPMXWOCB14} have shown remarkable performance in lots of tasks. A typical GAN model consists of two modules: a generative model $G(\cdot)$ to learn the distribution $p_{data}(\mathbf{x})$ over data $\mathbf{x}$, and a discriminative model $D(\cdot)$ to recognize whether a sample is from training data or from $G(\cdot)$. The minimax objective function is
\begin{equation}
\begin{aligned}
\min _{G} \max _{D} V(D, G)=\mathbb{E}_{\boldsymbol{x} \sim p_{\text {data }}(\boldsymbol{x})}[\log D(\boldsymbol{x})] \\
+\mathbb{E}_{\boldsymbol{z} \sim p_{\boldsymbol{z}}(\boldsymbol{z})}[\log (1-D(G(\boldsymbol{z})))].
\end{aligned}
\end{equation}
We will introduce the adversarial strategy to promote our model in missing view imputation and latent representation learning as well.

In classification, learning latent representation is simultaneously guided by the observed data and labels. The unsupervised setting is more challenging, especially for extensive missing cases and lack of labels. Therefore, we aim to effectively utilize both observed and unobserved views for learning latent representations. Specifically, this is achieved by improving the imputation for missing views and thus promoting latent representations. To ensure the rationality of the imputation, we introduce adversarial strategy to enforce the generated data for missing ones in each view to obey distribution of the observed data, as shown in Fig.~\ref{fig:framework2}. Then, the adversarial loss is induced as:
\begin{equation}
\begin{aligned}
\mathcal{L}_{adv} &=\sum_{n=1}^{N}\sum_{v=1}^{V}\sum_{i=1}^{I} \left( 1-s_{nv} \right)
\left[ \log D_v\left(   \mathbf{x}^{(v)}_{i};\boldsymbol{\Theta}^{(v)}_d  \right) +\right.\\
&\phantom{=\;\;}\left.\log \left( 1-D_v\left( G_v\left( \mathbf{h}_n;\boldsymbol{\Theta}^{(v)}_g\right);\boldsymbol{\Theta}^{(v)}_d\right)\right)\right] ,\\
\label{equ:gan}
\end{aligned}
\end{equation}
where $(1-s_{nv})$ indicates whether or not the $n$th sample in the $v$th view needs to be imputed. $G_v(\cdot;\boldsymbol{\Theta}^{(v)}_g)$ and $D_v(\cdot; \boldsymbol{\Theta}^{(v)}_d)$ are the generator (degradation networks)  and  discriminator for the $v$th view parameterized by $\boldsymbol{\Theta}^{(v)}_g$ and $\boldsymbol{\Theta}^{(v)}_d$, respectively.
 $\{\mathbf{x}^{(v)}_{i}\}_{i}^I$ are the available data in the $v$th view, with the total number of samples being $I$. In this way, all the available data act as positive samples to train the discriminators.
The missing data $G_v\left( \mathbf{h}_n;\boldsymbol{\Theta}^{(v)}_g\right)$ generated by the generator are updated iteratively to approximate the distribution of observed data.

Accordingly, the overall objective function for unsupervised CPM-Nets is induced as:

\begin{equation}
\begin{aligned}
&\mathcal{L}=\min _{G}\max _{D}\min _{\mathbf{h}} \mathcal{L}_{adv}+\mathcal{L}_{rec}\\
\text{with}~
&\mathcal{L}_{rec} = \sum ^{N}_{n=1}\sum ^{V}_{v=1}s_{nv}\left|\left| G_{v}\left( \mathbf{h}_n;\boldsymbol{\Theta}^{(v)}_g\right) -\mathbf{x}^{(v)}_{n}\right|\right|^{2},
\label{equ:self-sub1}
\end{aligned}
\end{equation}
where $\mathcal{L}_{rec}(\cdot)$ is the reconstruction loss which is actually supervised by the observed data. In this way, the missing data are considered as variables to be jointly optimized with the latent representations, and thus the latent representation and the missing views are updated alternatively to improve each other.
\begin{algorithm}[t]
\SetAlgoLined
\caption{Algorithm for CPM-GAN}

\KwIn{Partial multi-view dataset: $\mathcal{D} = \{\mathcal{S}_1, ..., S_N$\}.}
\textbf{Initialize:} {Initialize $\{\mathbf{h}_n\}_{n=1}^N$ , $\{\boldsymbol{\Theta}^{(v)}_{g}\}_{v=1}^V$ and $\{\boldsymbol{\Theta}^{(v)}_{d}\}_{v=1}^V$ with random values}.\\
\While{not converged}{
\For{$v=1:V$ }
{
Update the discriminator parameters $\mathbf{\Theta}^{(v)}_{d}$ with gradient descent: \\
$\boldsymbol{\Theta}^{(v)}_{d} \leftarrow \boldsymbol{\Theta}^{(v)}_{d} + \eta \partial \mathcal{L}/\partial \boldsymbol{\Theta}^{(v)}_{d}$;\\
}
\For{$v=1:V$ }
{
Update the generator parameters $\mathbf{\Theta}^{(v)}_{g}$ with gradient descent: \\
$\boldsymbol{\Theta}^{(v)}_{g} \leftarrow \boldsymbol{\Theta}^{(v)}_{g} - \eta \partial \mathcal{L}/\partial \boldsymbol{\Theta}^{(v)}_{g}$;\\
}
\For{$n=1:N$ }
{
Update the latent representation $\mathbf{h}_n$ with gradient descent: \\
$\mathbf{h}_n \leftarrow \mathbf{h}_n-\eta  \partial \mathcal{L}/\partial \mathbf{h}_n $;\\
}
}
\KwOut{
\begin{small}
networks parameters  $\{\boldsymbol{\Theta}^{(v)}_{g}\}_{v=1}^V$ and $\{\boldsymbol{\Theta}^{(v)}_{d}\}_{v=1}^V$ and latent representation $\{\mathbf{h}_n\}_{n=1}^N$.
 \end{small}
}
\label{alg:alg2}
\end{algorithm}

Although both use adversarial strategy, there are several key differences between generative adversarial networks (GAN) and our model: (1) the inputs of GAN are fixed, which are sampled from one specific distribution, while in our algorithm the latent representations acting as inputs are variables to be optimized; (2) different from original GAN, our model of representation learning is performed in a sample-to-sample supervision manner \cite{DBLP:conf/icml/LarsenSLW16},\cite{DBLP:conf/nips/LiuBK17}; (3) in our model, there are multiple discriminators (each one for a view), which are used to make the generated missing data obey the distribution of the observed data to stabilize the imputation and thus enhance the representation learning. The optimization of the proposed CPM-GAN is summarized in Algorithm~\ref{alg:alg2}.

\begin{figure*}[!ht]
\centering
\subfigure[Animal]{
\begin{minipage}[t]{0.32\linewidth}
\centering
\includegraphics[width=2.4in,height = 1.9in]{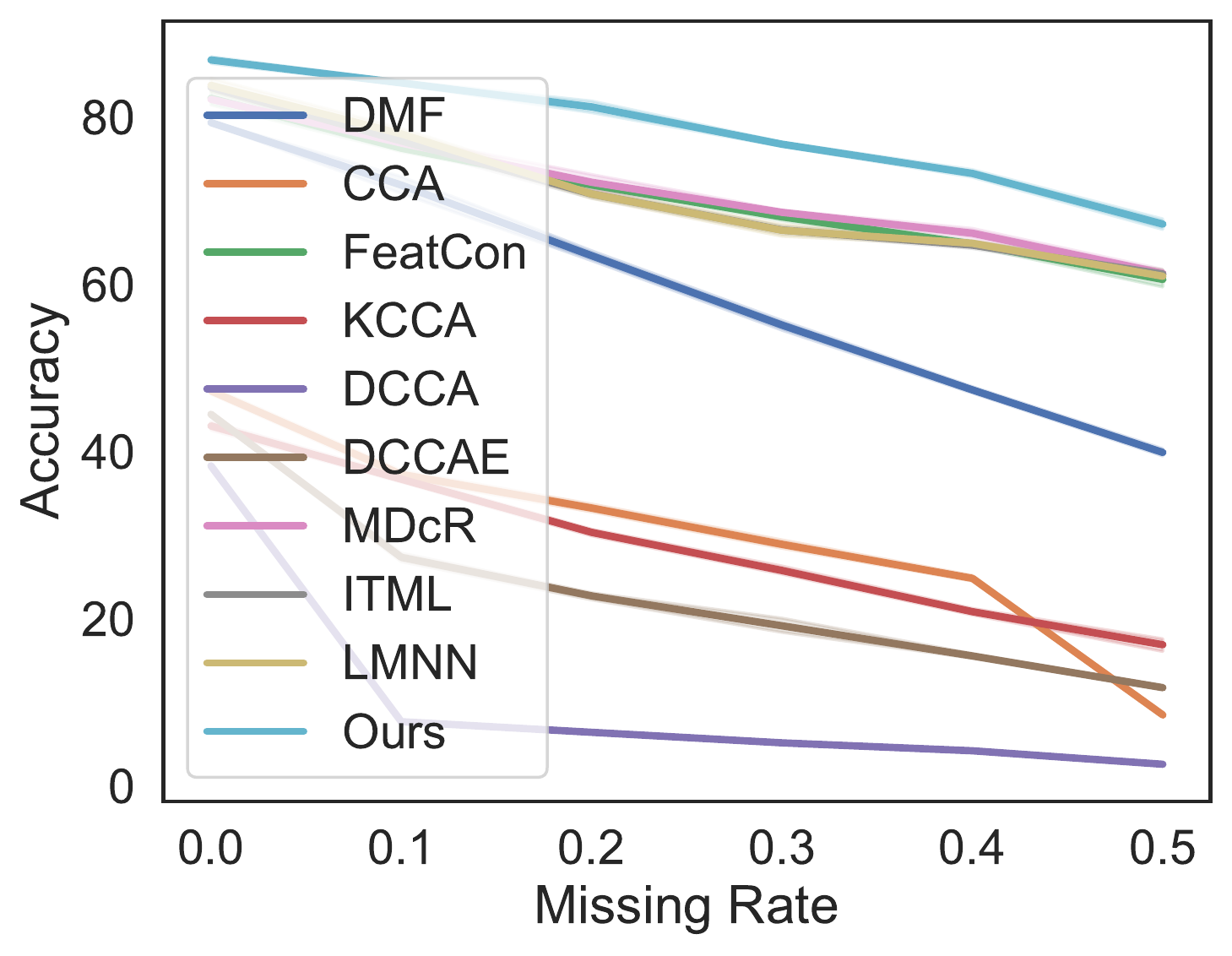}
\end{minipage}}
\subfigure[Handwritten]{
\begin{minipage}[t]{0.32\linewidth}
\centering
\includegraphics[width=2.4in,height = 1.9in]{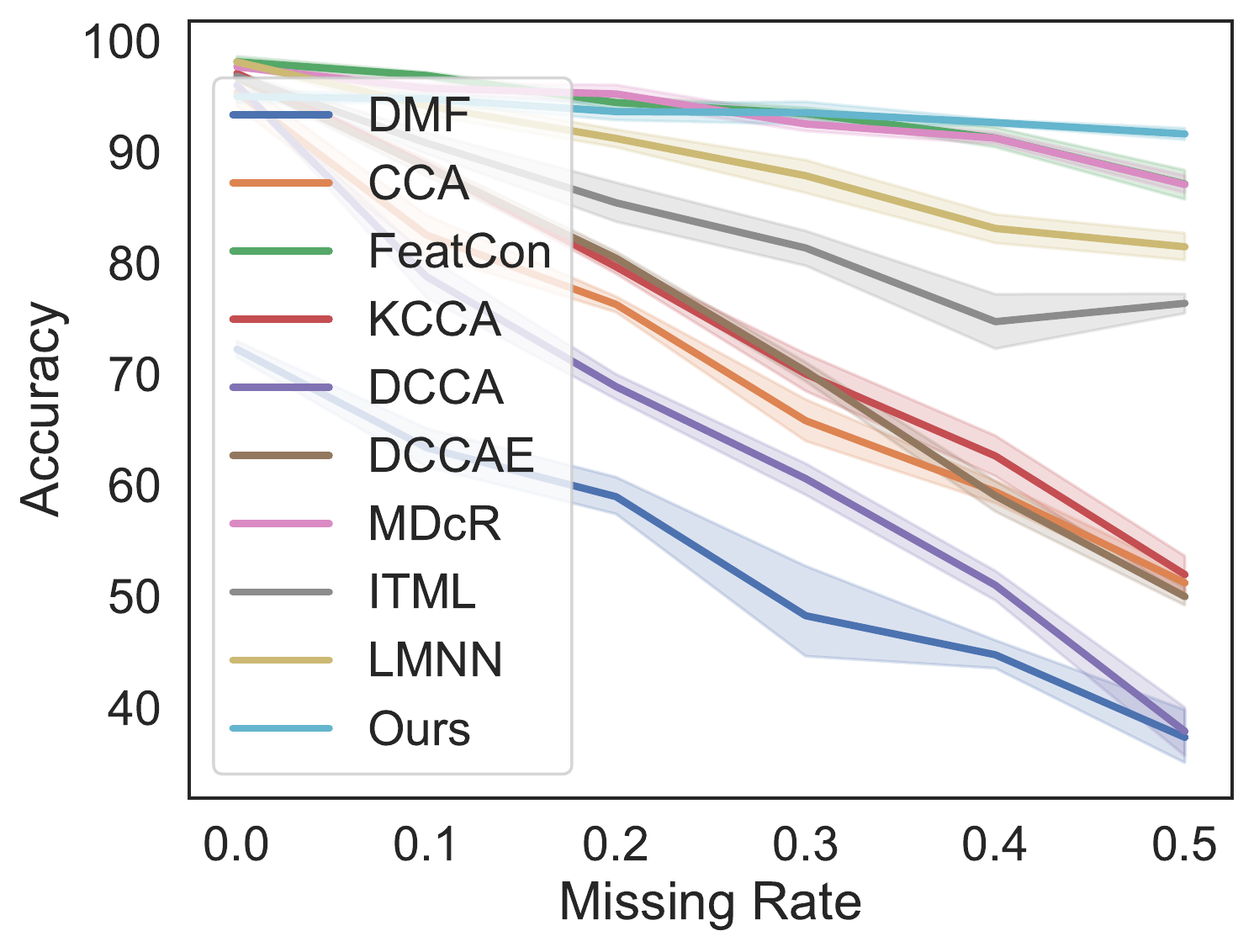}
\end{minipage}}
\centering
\subfigure[CUB]{
\begin{minipage}[t]{0.32\linewidth}
\centering
\includegraphics[width=2.4in,height = 1.9in]{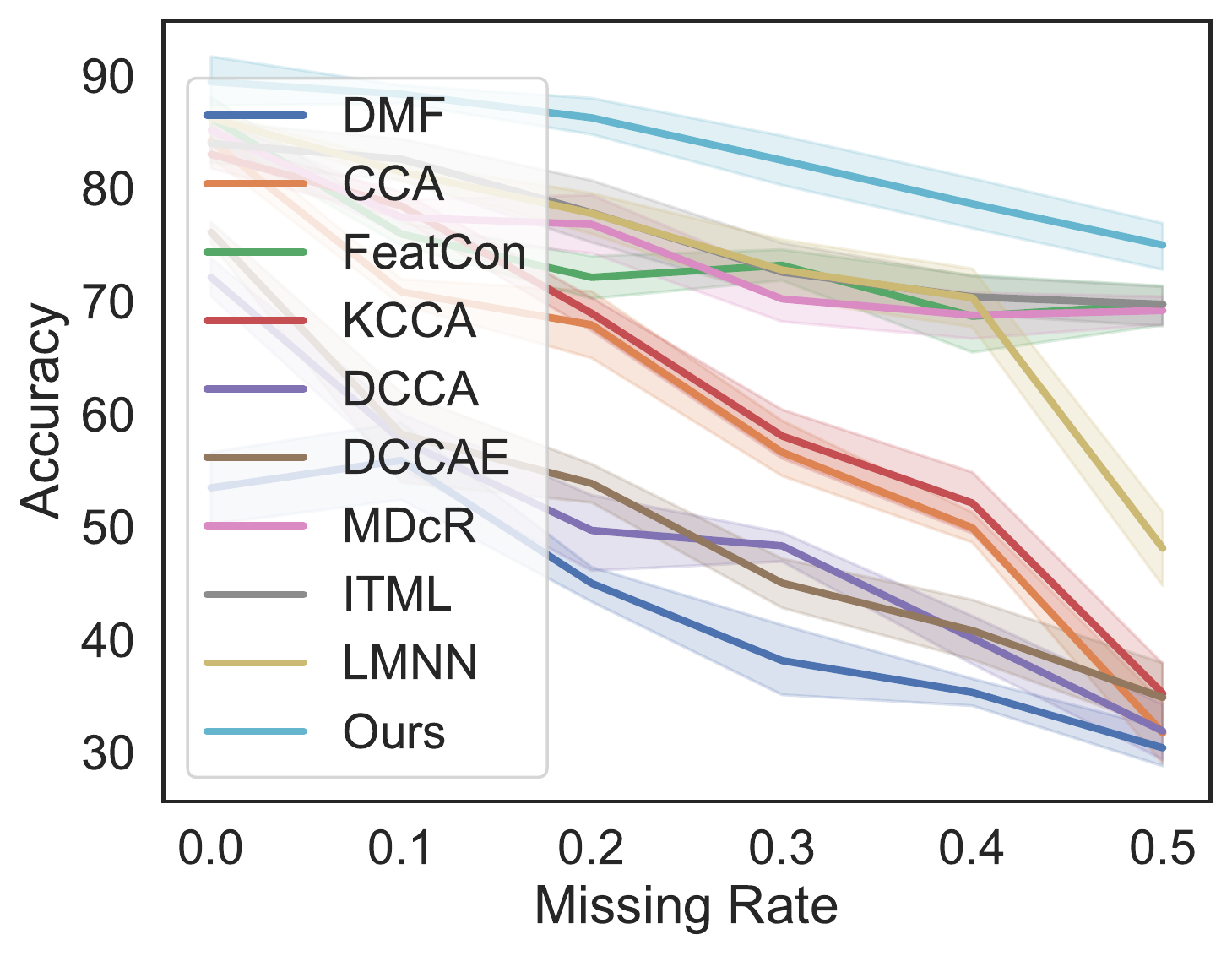}
\end{minipage}}
\centering
\subfigure[3Sources-complete]{
\begin{minipage}[t]{0.32\linewidth}
\centering
\includegraphics[width=2.4in,height = 1.9in]{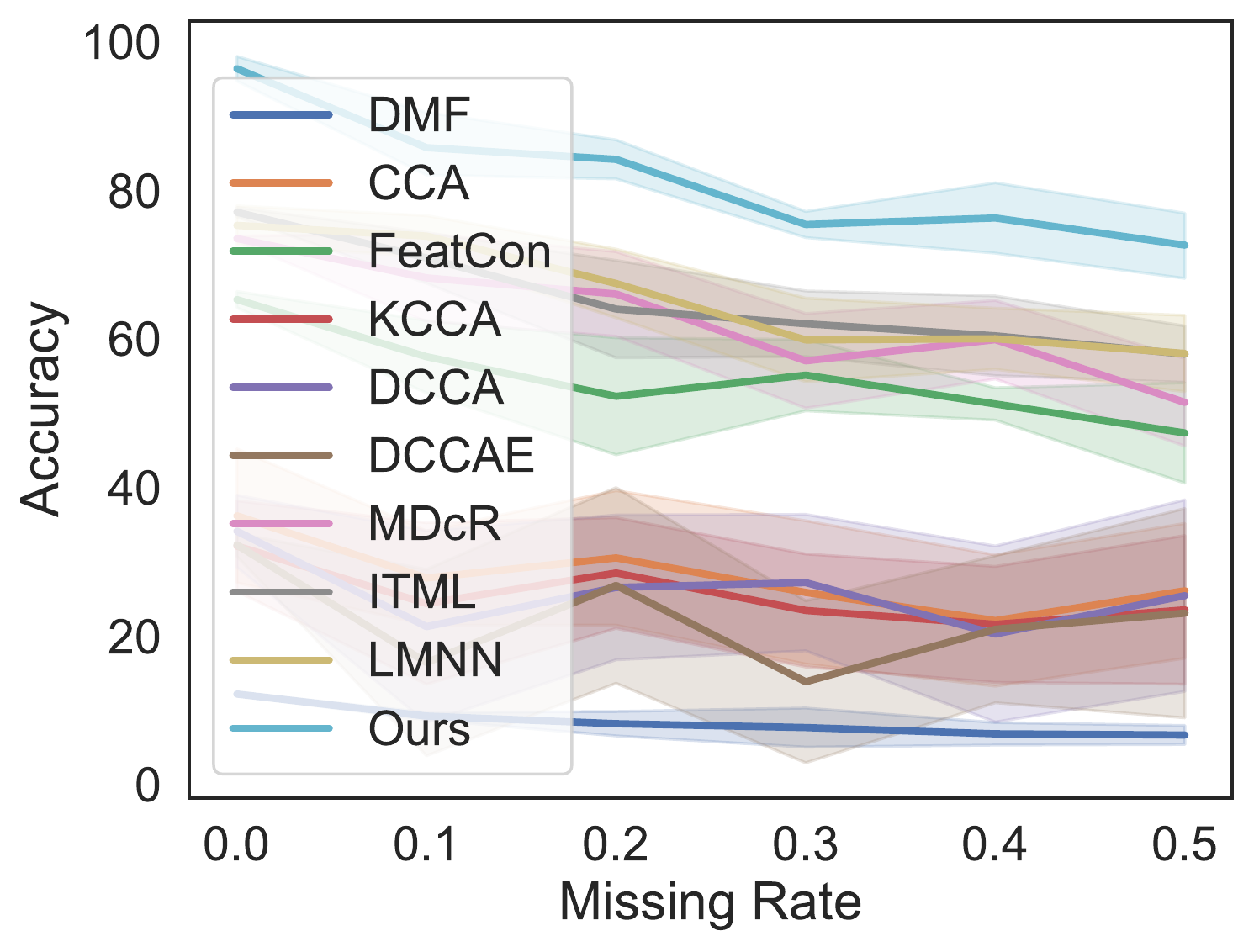}
\end{minipage}}
\centering
\subfigure[Football]{
\begin{minipage}[t]{0.32\linewidth}
\centering
\includegraphics[width=2.4in,height = 1.9in]{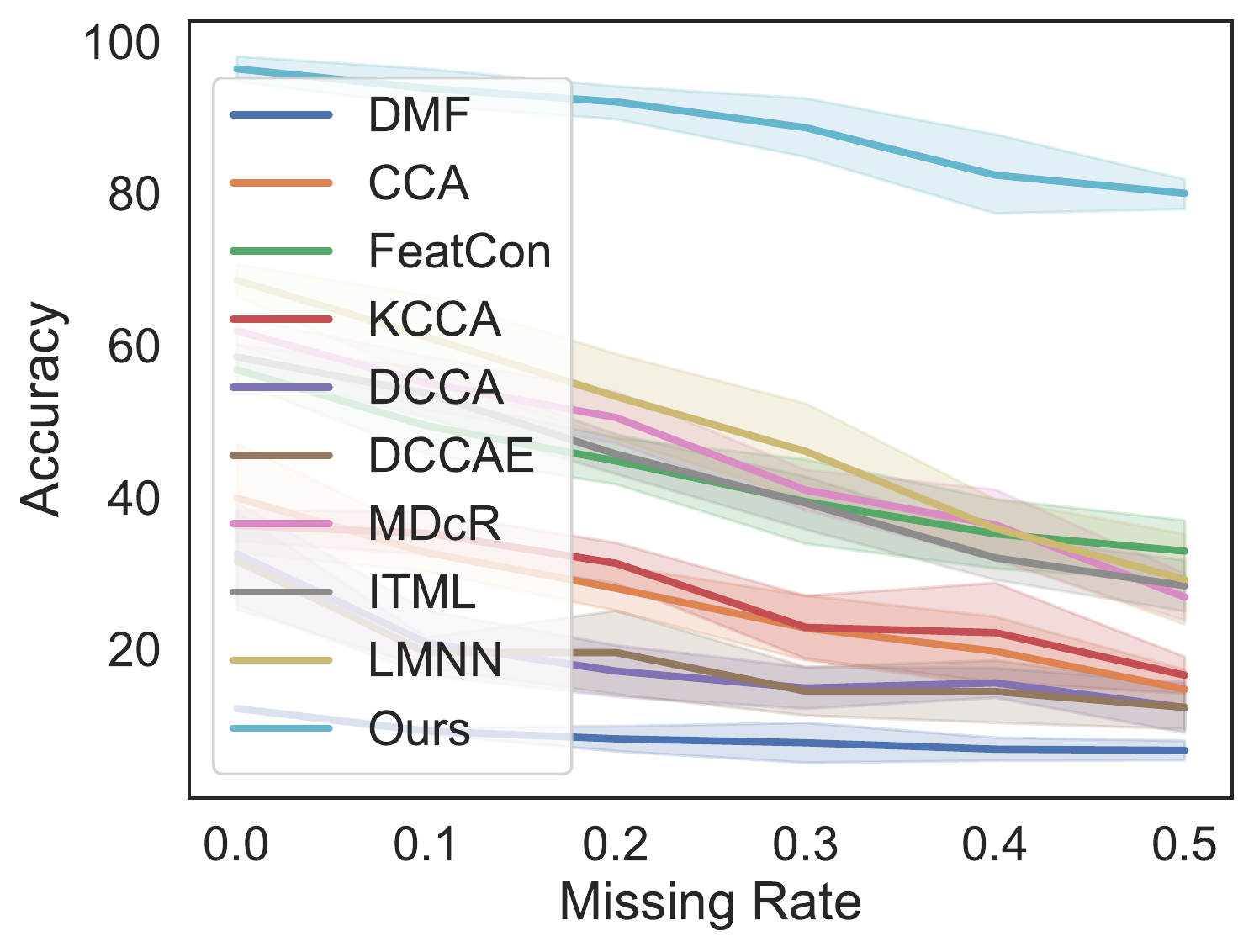}
\end{minipage}}
\centering
\subfigure[Politics]{
\begin{minipage}[t]{0.32\linewidth}
\centering
\includegraphics[width=2.4in,height = 1.9in]{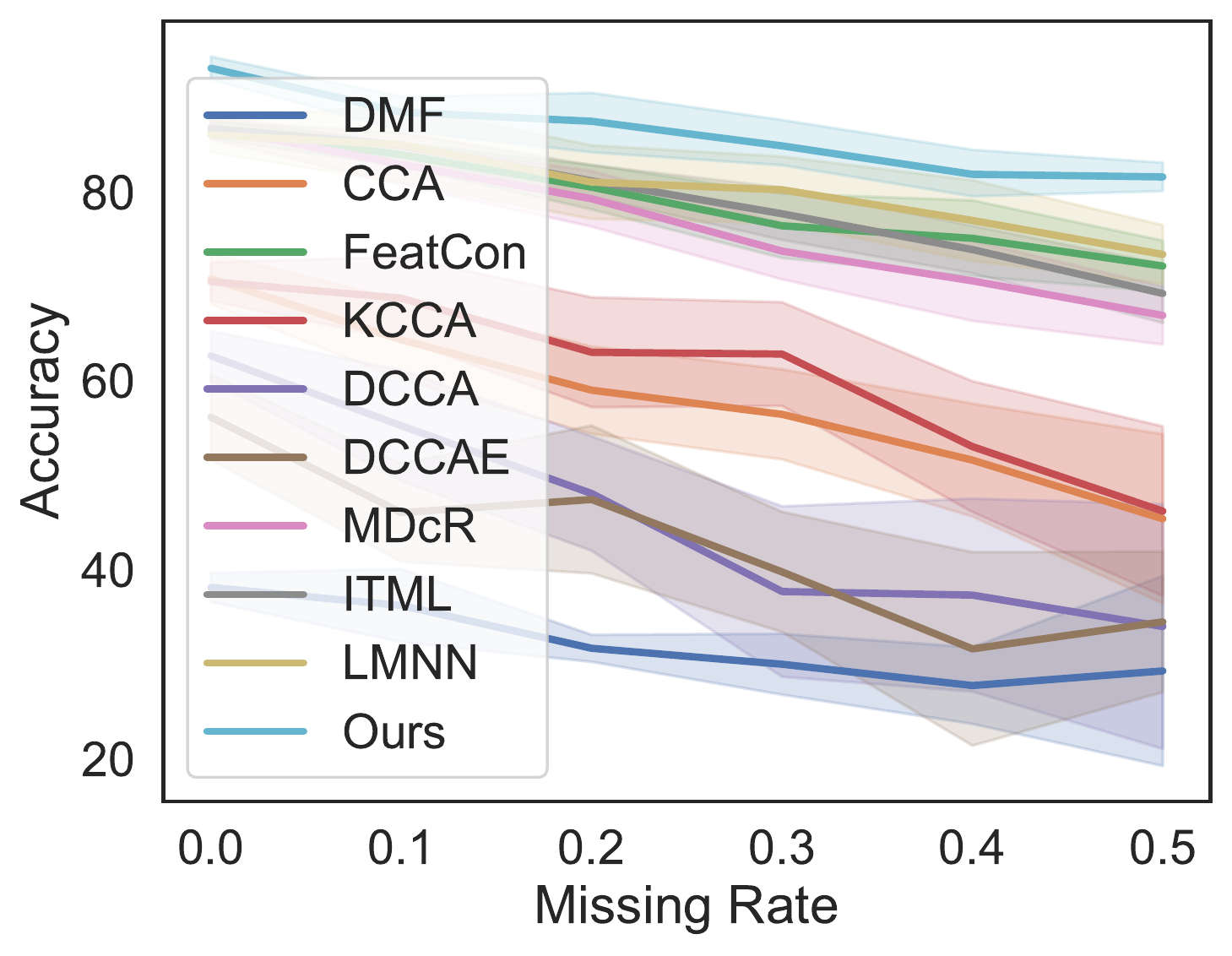}
\end{minipage}}
\caption{Classification performance comparison under different missing rates ($\eta$).}
\label{fig:CPMNets}
\end{figure*}
\begin{figure*}[t]
\center
{\includegraphics[width=0.9\linewidth,height = 2.2in]{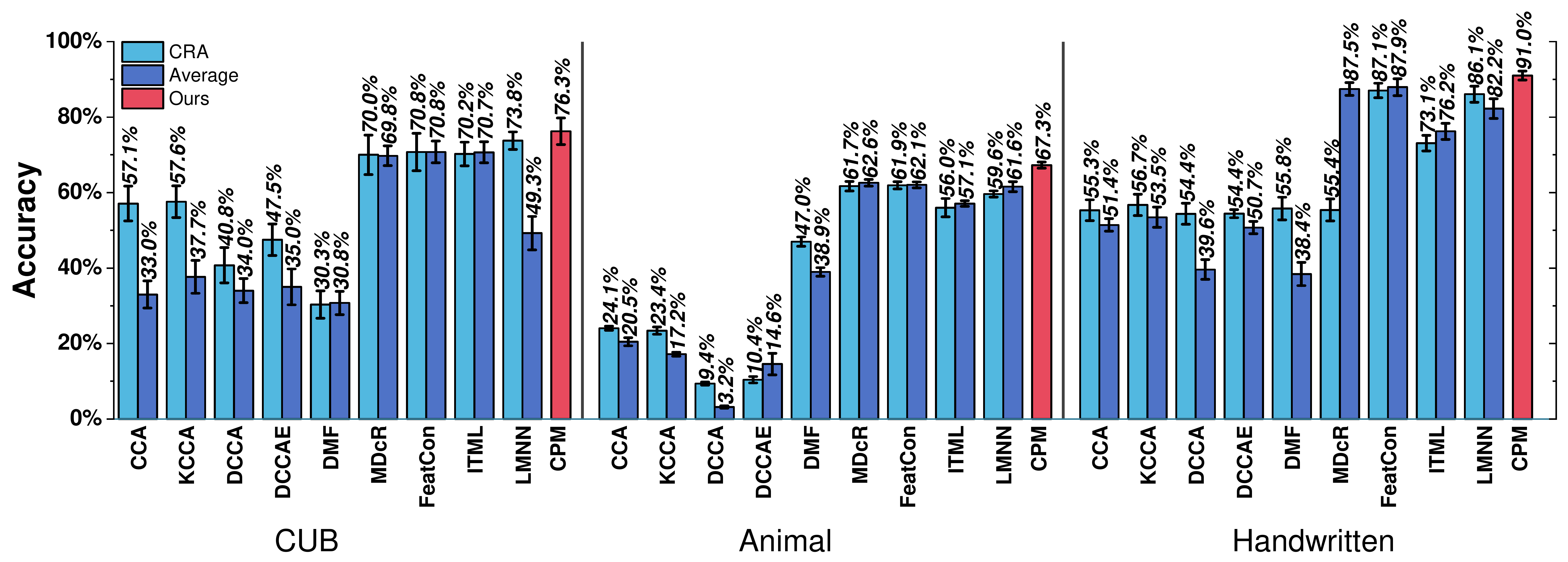}}
{\includegraphics[width=1.0\linewidth,height = 2.2in]{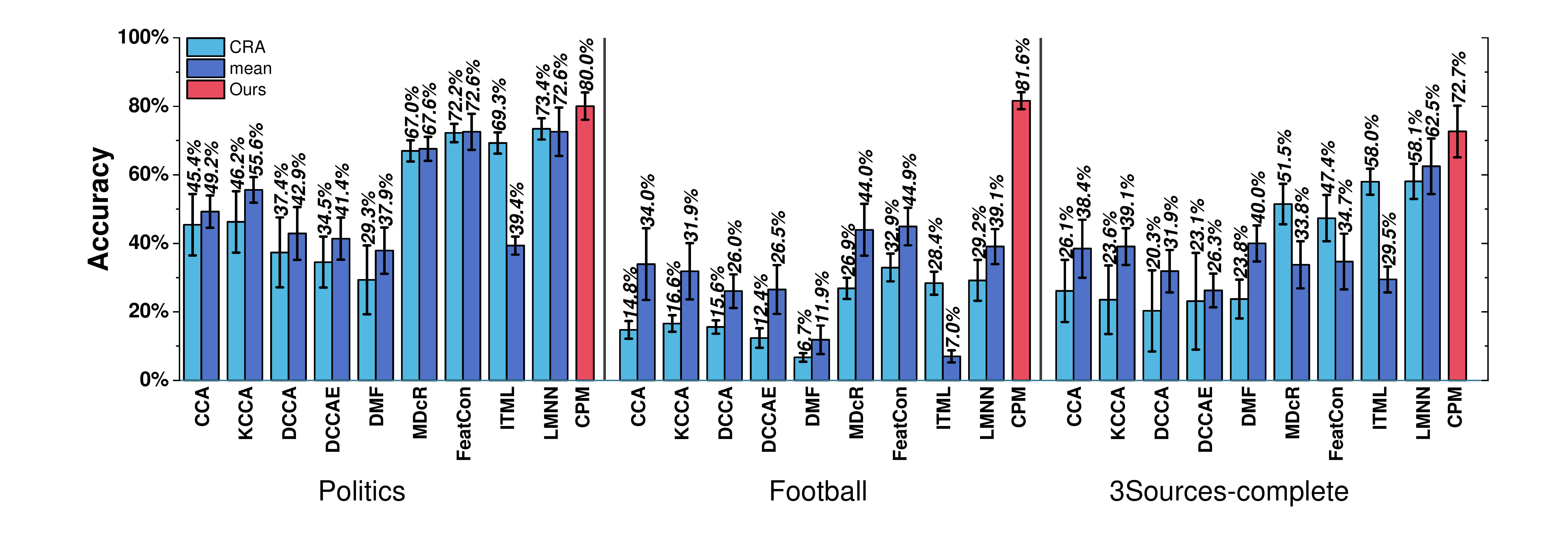}}
{\includegraphics[width=1.0\linewidth,height = 2.2in]{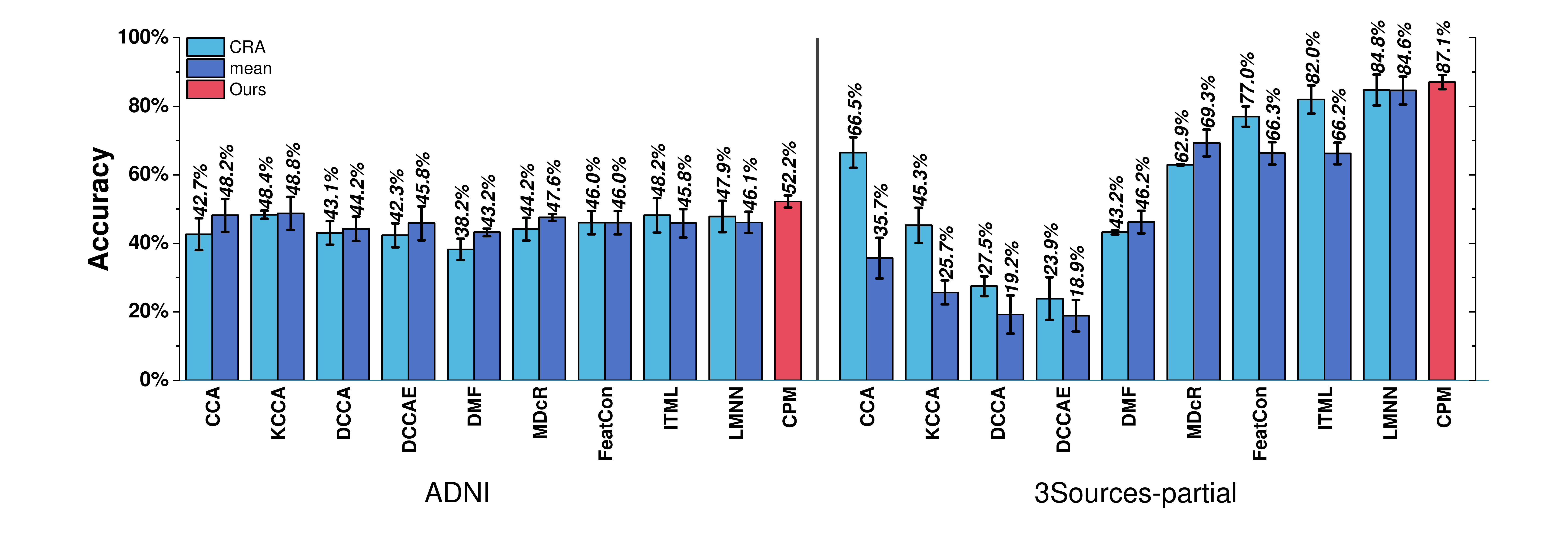}}
\caption{Classification comparison with view completion using average value and cascaded residual autoencoder (CRA) \cite{tran2017missing} (with missing rate $\eta = 0.5$). ADNI and 3Sources-partial are multi-modal datasets with naturally modality missing. }
\label{fig:mean-cra}
\end{figure*}
\begin{figure*}[!ht]
\centering
\subfigure[FeatCon (U)]{
\centering
\begin{minipage}[t]{0.25\linewidth}
\centering
\fbox{\includegraphics[width=1.5in,height = 1.3in]{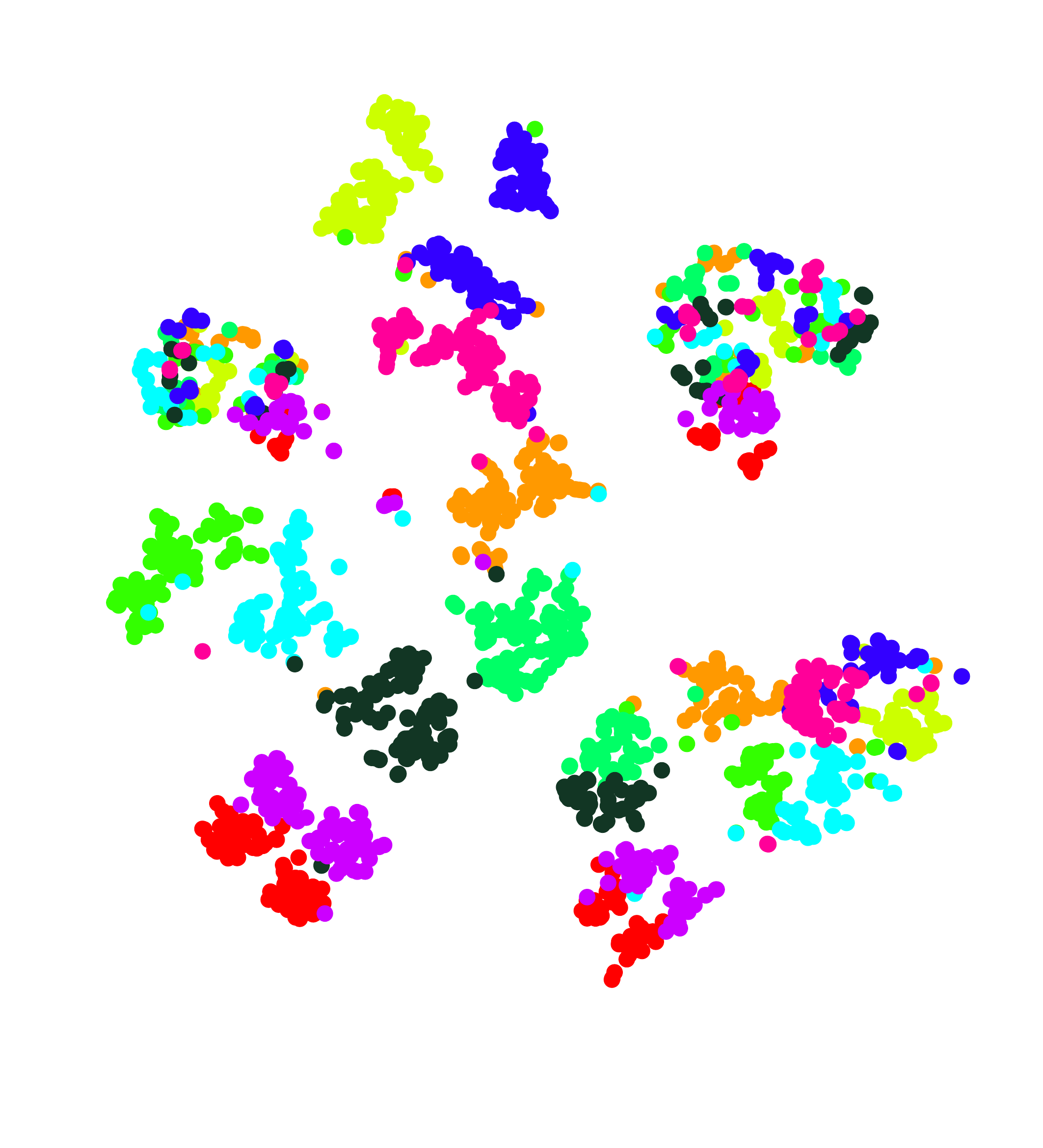}}
\centering
\end{minipage}}
\centering
\subfigure[DCCA (U)]{
\begin{minipage}[t]{0.25\linewidth}
\centering
\fbox{\includegraphics[width=1.5in,height = 1.3in]{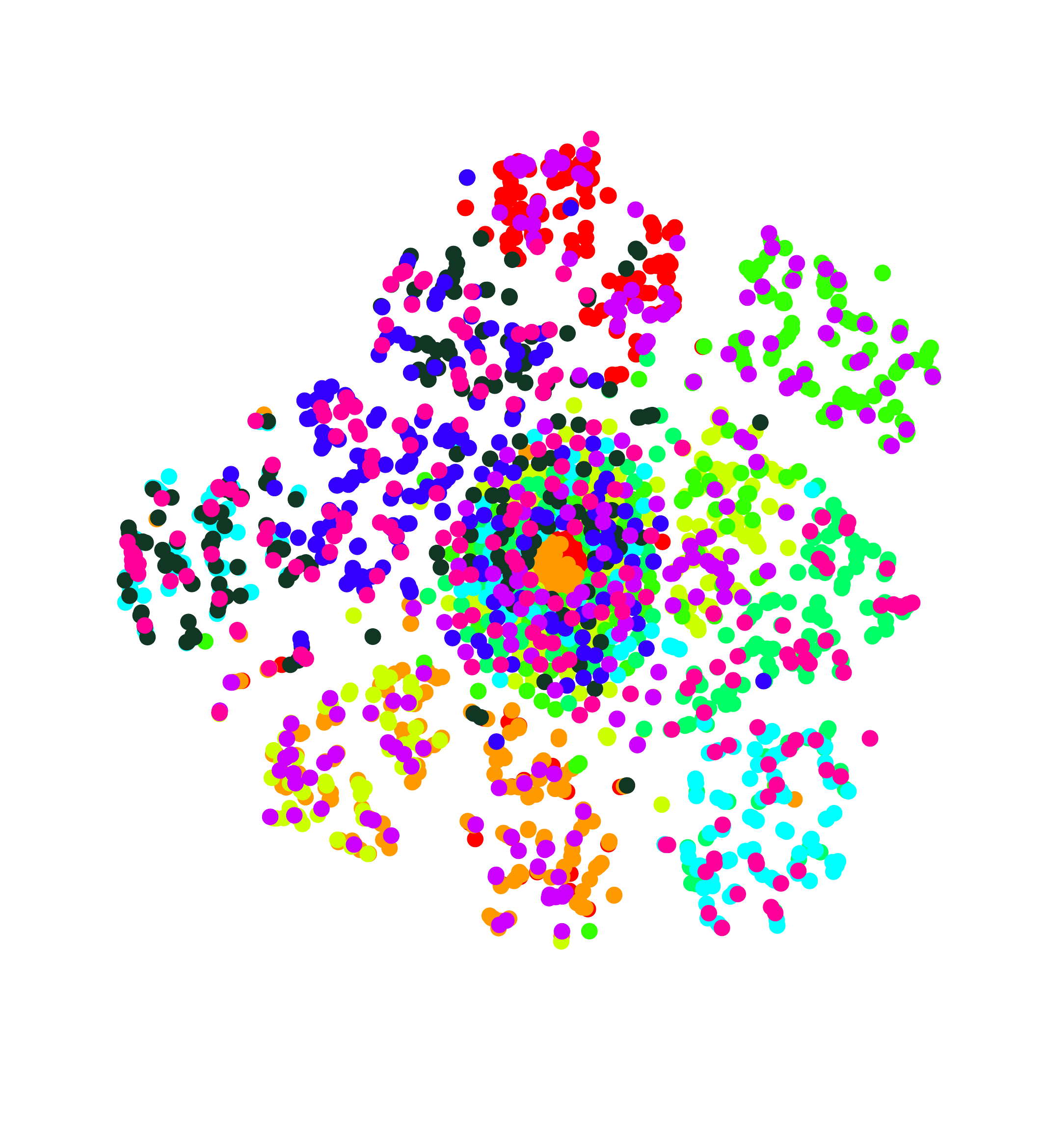}}
\end{minipage}}
\centering
\subfigure[Ours (U)]{
\begin{minipage}[t]{0.25\linewidth}
\centering
\fbox{\includegraphics[width=1.5in,height = 1.3in]{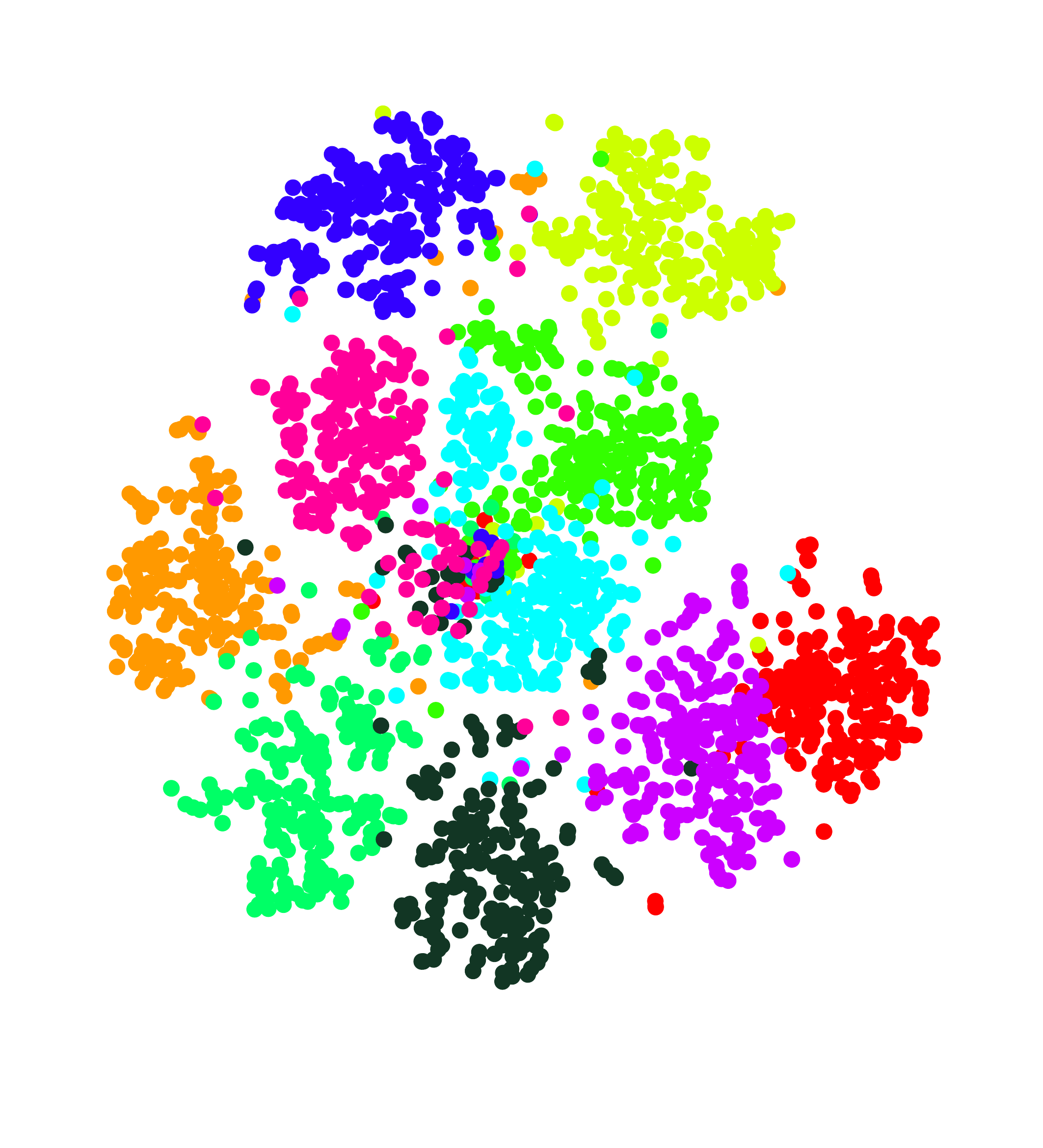}}
\end{minipage}}
\subfigure[LMNN (S)]{
\begin{minipage}[t]{0.25\linewidth}
\centering
\fbox{\includegraphics[width=1.5in,height = 1.3in]{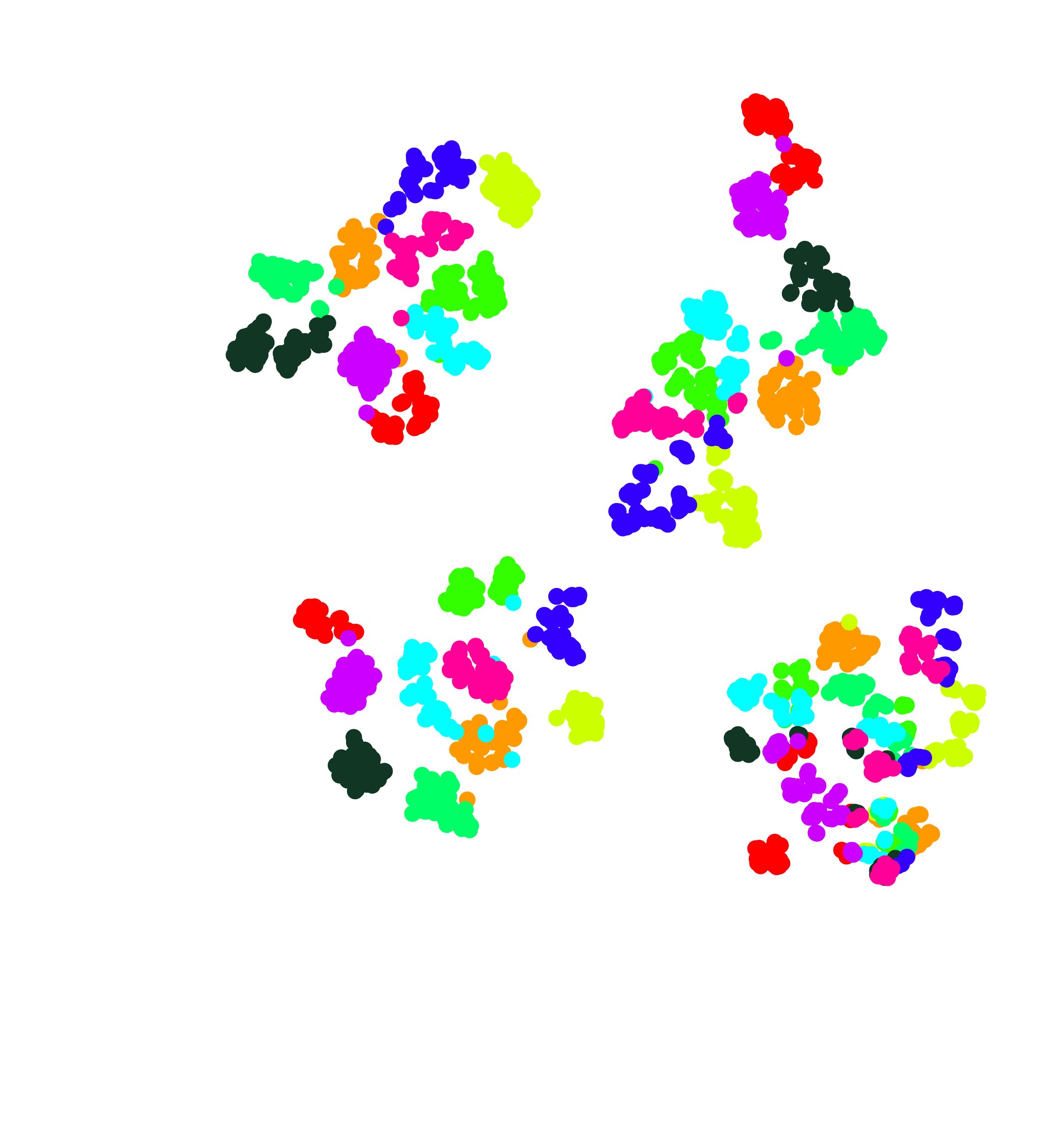}}
\end{minipage}}
\subfigure[ITML (S)]{
\begin{minipage}[t]{0.25\linewidth}
\centering
\fbox{\includegraphics[width=1.5in,height = 1.3in]{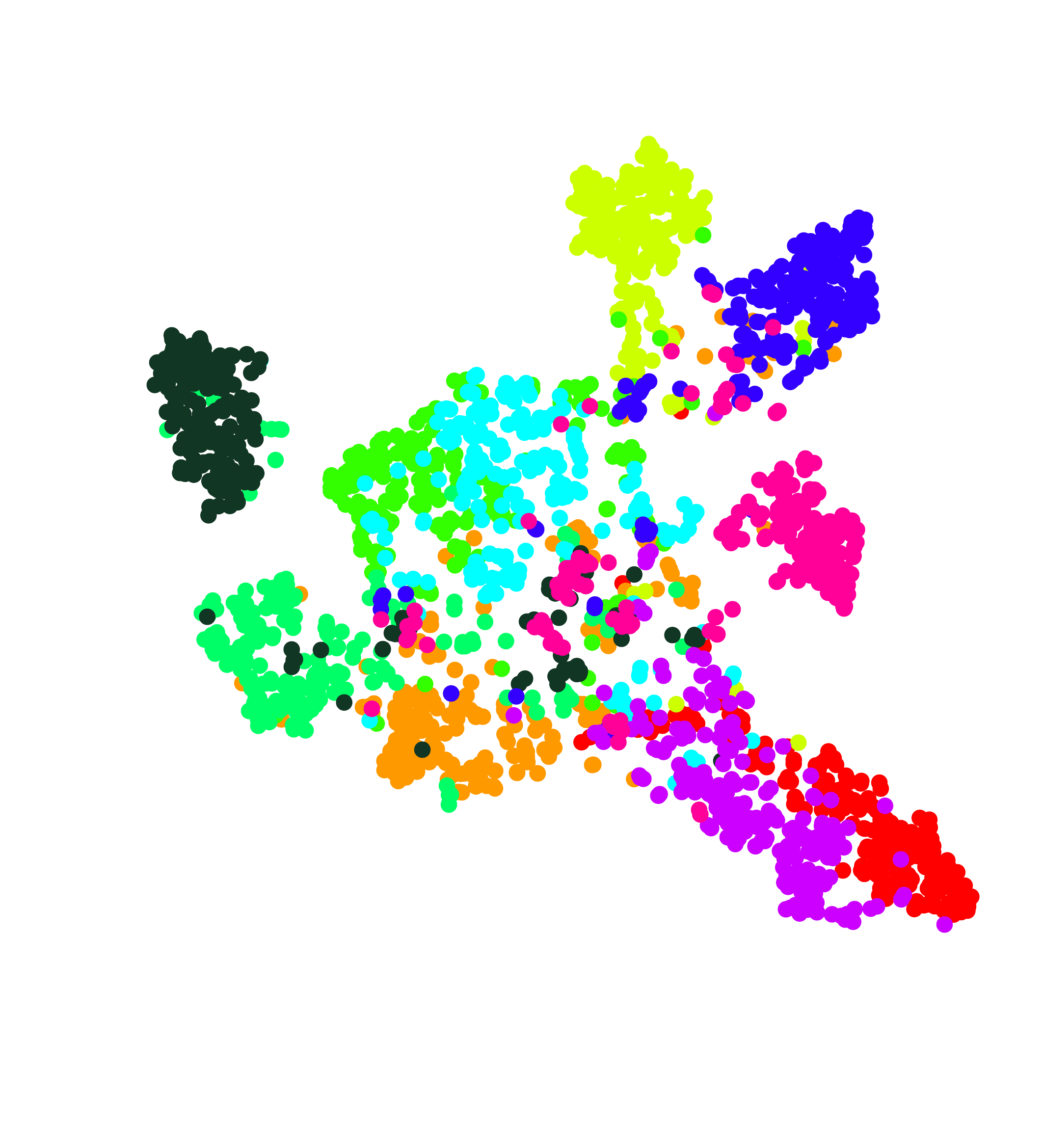}}
\end{minipage}}
\centering
\subfigure[Ours (S)]{
\begin{minipage}[t]{0.25\linewidth}
\centering
\fbox{\includegraphics[width=1.5in,height = 1.3in]{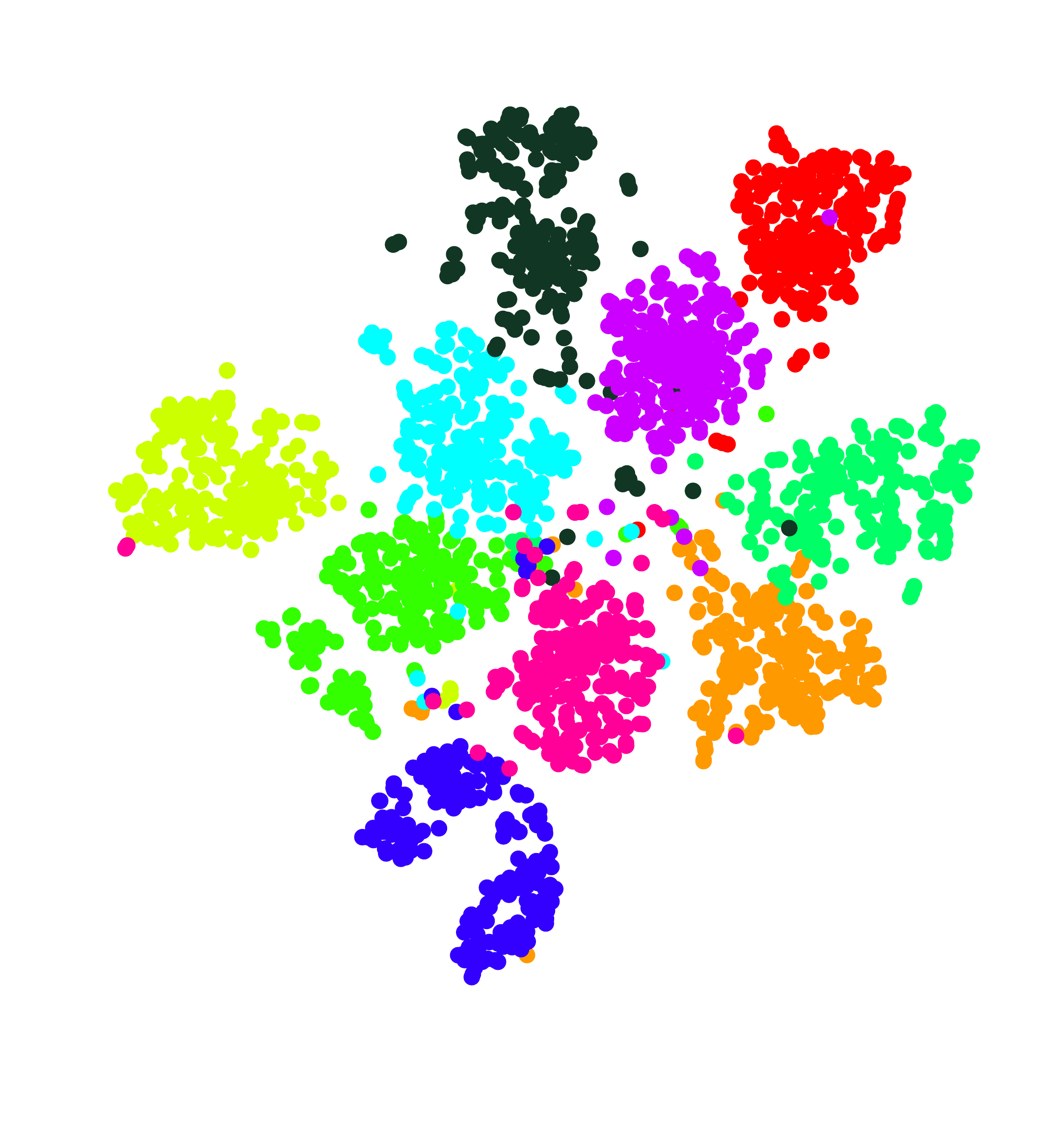}}
\end{minipage}}
\begin{minipage}[t]{0.0001\linewidth}
\centering
\includegraphics[width=0.2in,height = 1.3in]{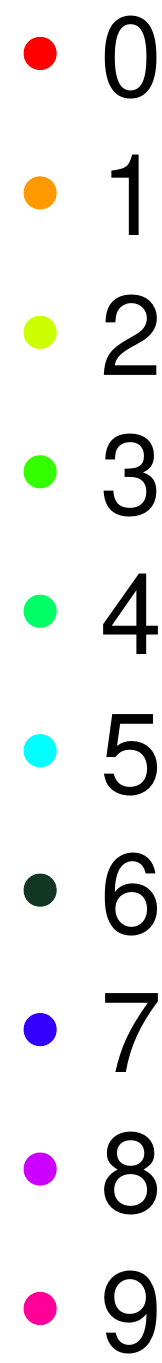}
\end{minipage}
\centering
\caption{Visualization of representations on Handwritten with missing rate $\eta = 0.5$, where `U' and `S' indicate `unsupervised' and `supervised' settings in representation learning. (Zoom in for best view).}
\label{fig:tsne}
\end{figure*}
\begin{figure}[!ht]
\centering
\subfigure[Handwritten]{
\begin{minipage}[t]{0.43\linewidth}
\centering
\includegraphics[width=1.6in,height = 1.5in]{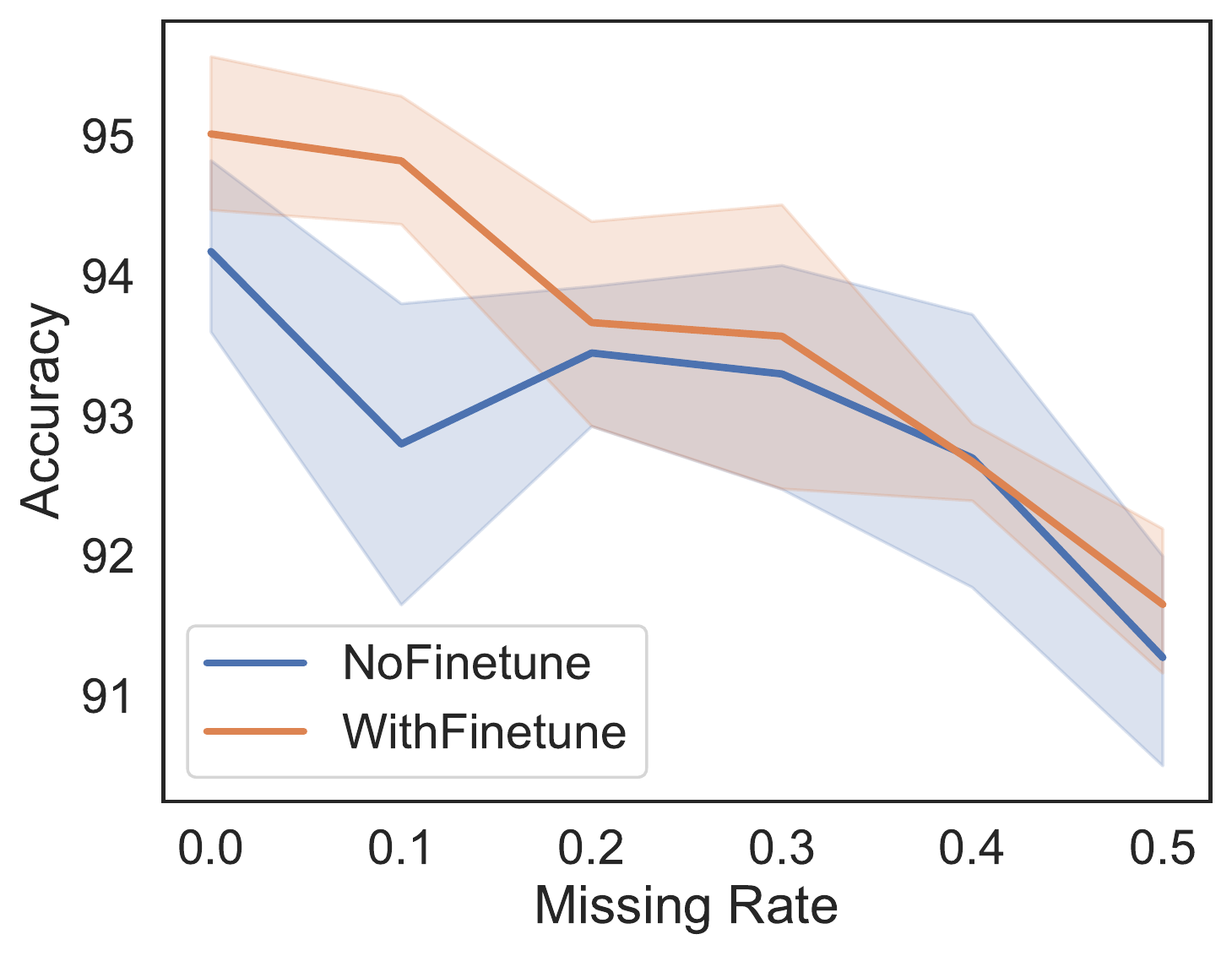}
\end{minipage}}
\centering
\subfigure[CUB]{
\begin{minipage}[t]{0.43\linewidth}
\centering
\includegraphics[width=1.6in,height = 1.5in]{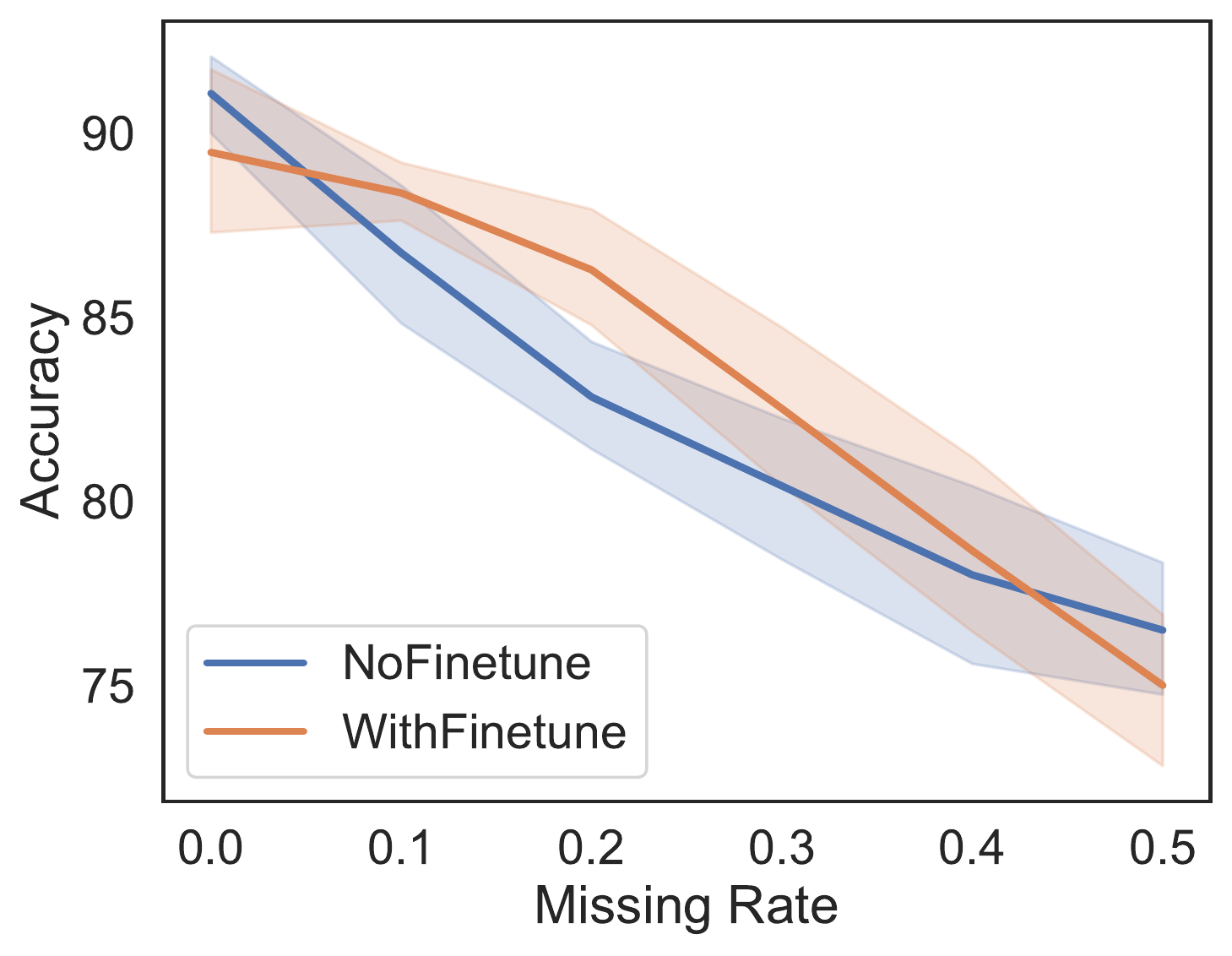}
\end{minipage}}
\centering
\caption{Fine-tuning evaluation.}
\label{fig:finetune}
\end{figure}

\begin{figure*}[htbp]
\centering
\subfigure[Animal]{
\begin{minipage}[t]{0.32\linewidth}
\centering
\includegraphics[width=2.4in,height = 1.8in]{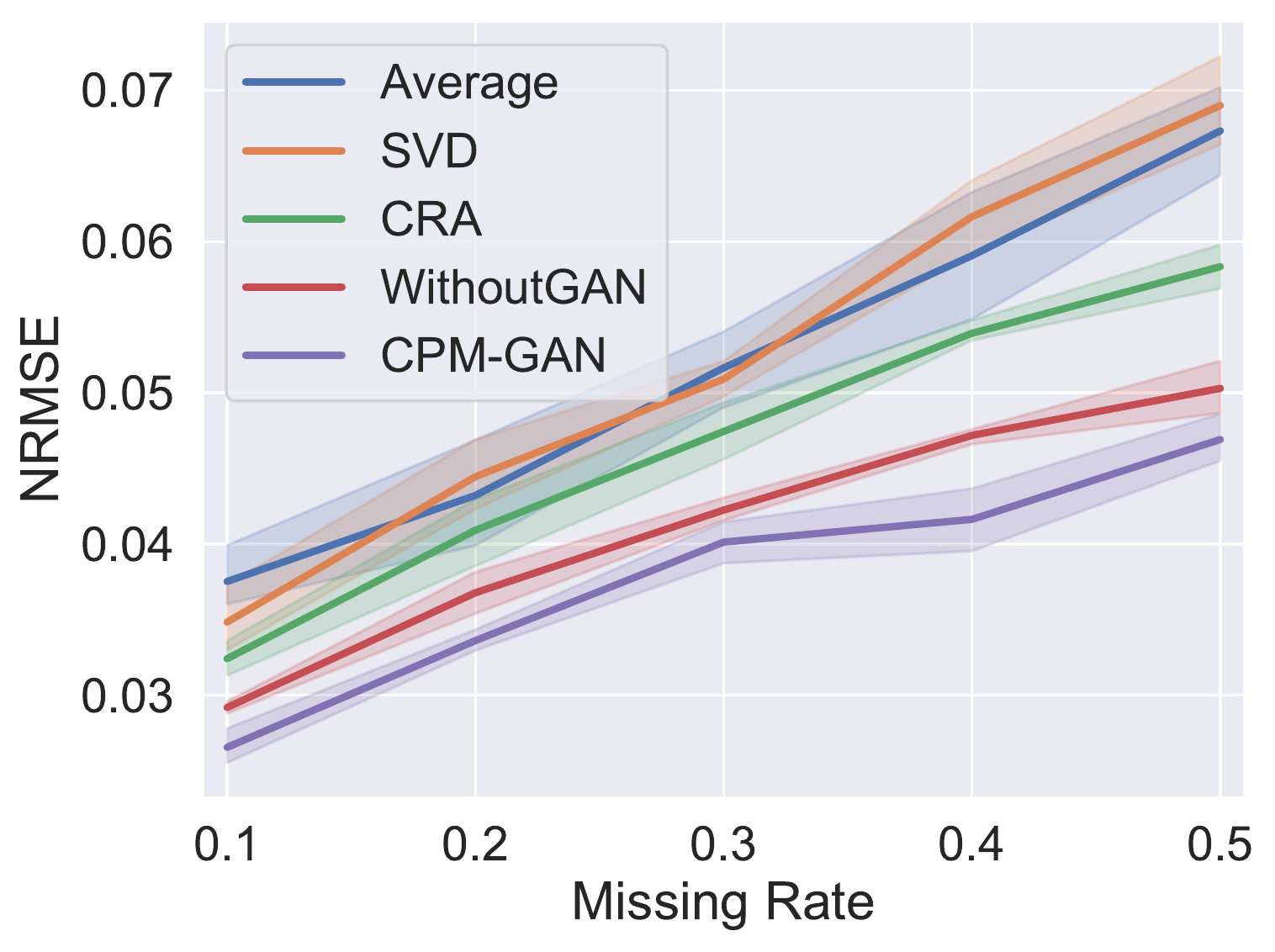}
\end{minipage}}
\centering
\subfigure[Handwritten]{
\begin{minipage}[t]{0.32\linewidth}
\centering
\includegraphics[width=2.4in,height = 1.8in]{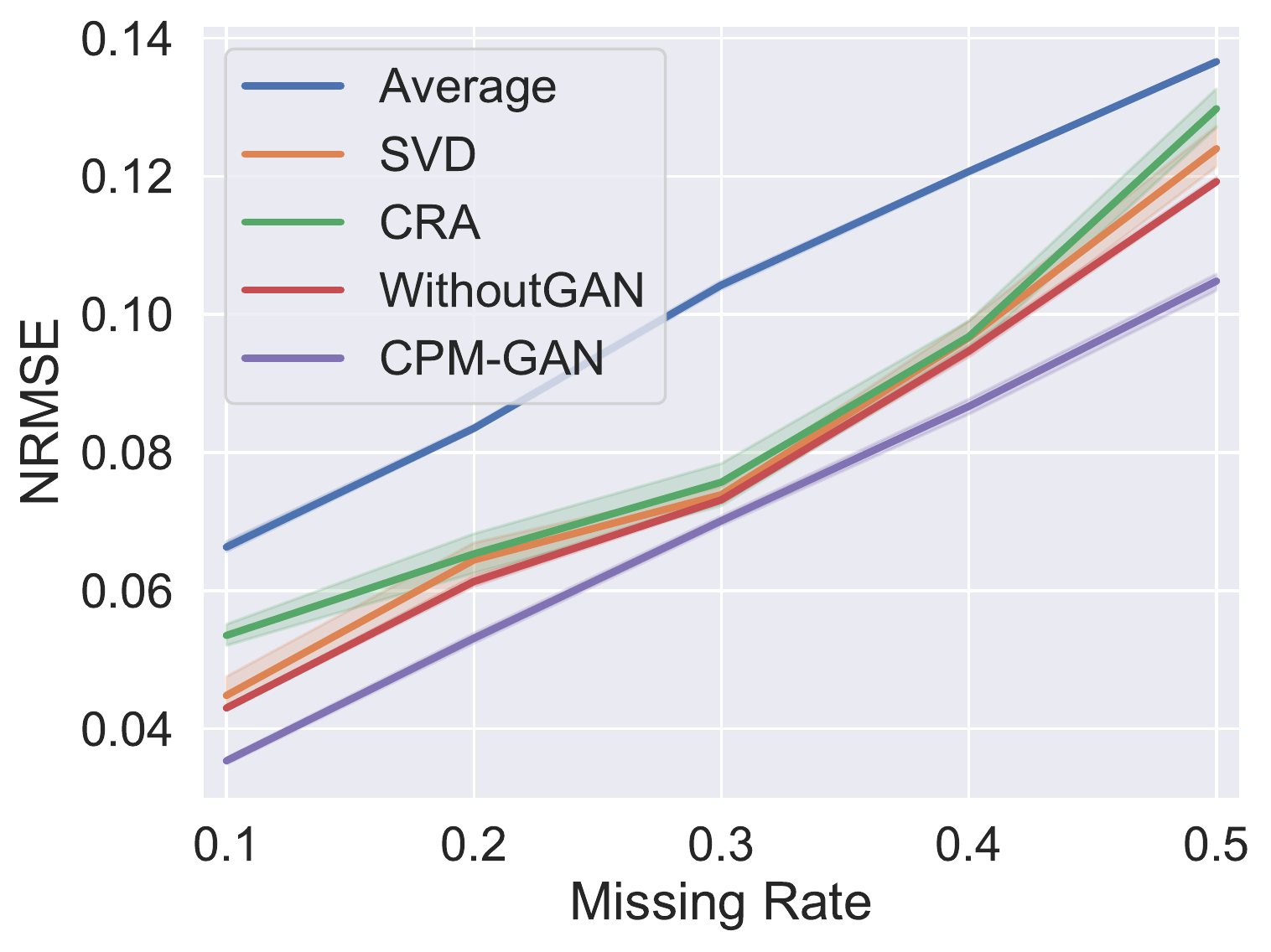}
\end{minipage}}
\subfigure[CUB]{
\begin{minipage}[t]{0.32\linewidth}
\centering
\includegraphics[width=2.4in,height = 1.8in]{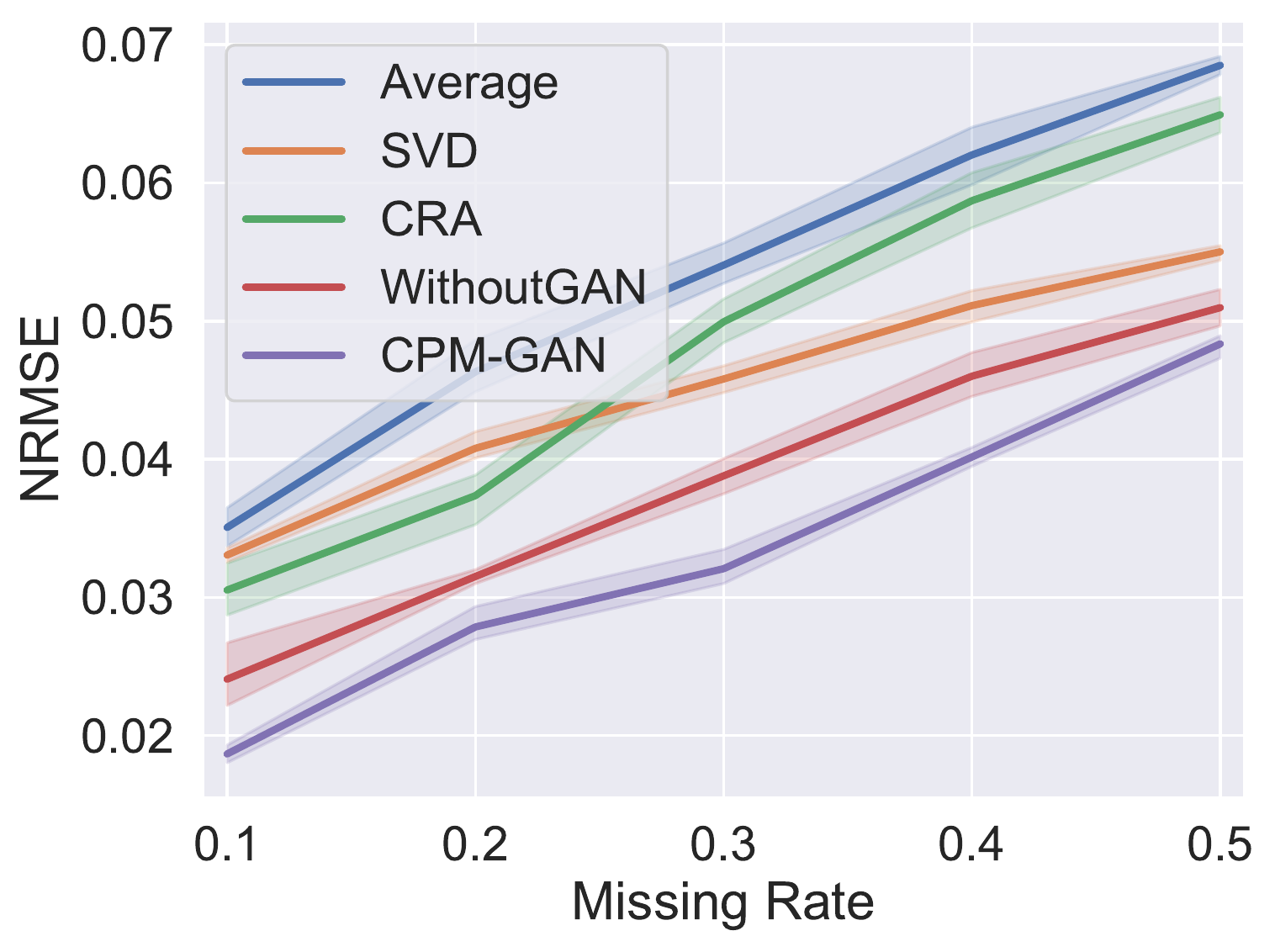}
\end{minipage}}

\centering
\subfigure[3Sources-complete]{
\begin{minipage}[t]{0.32\linewidth}
\centering
\includegraphics[width=2.4in,height = 1.8in]{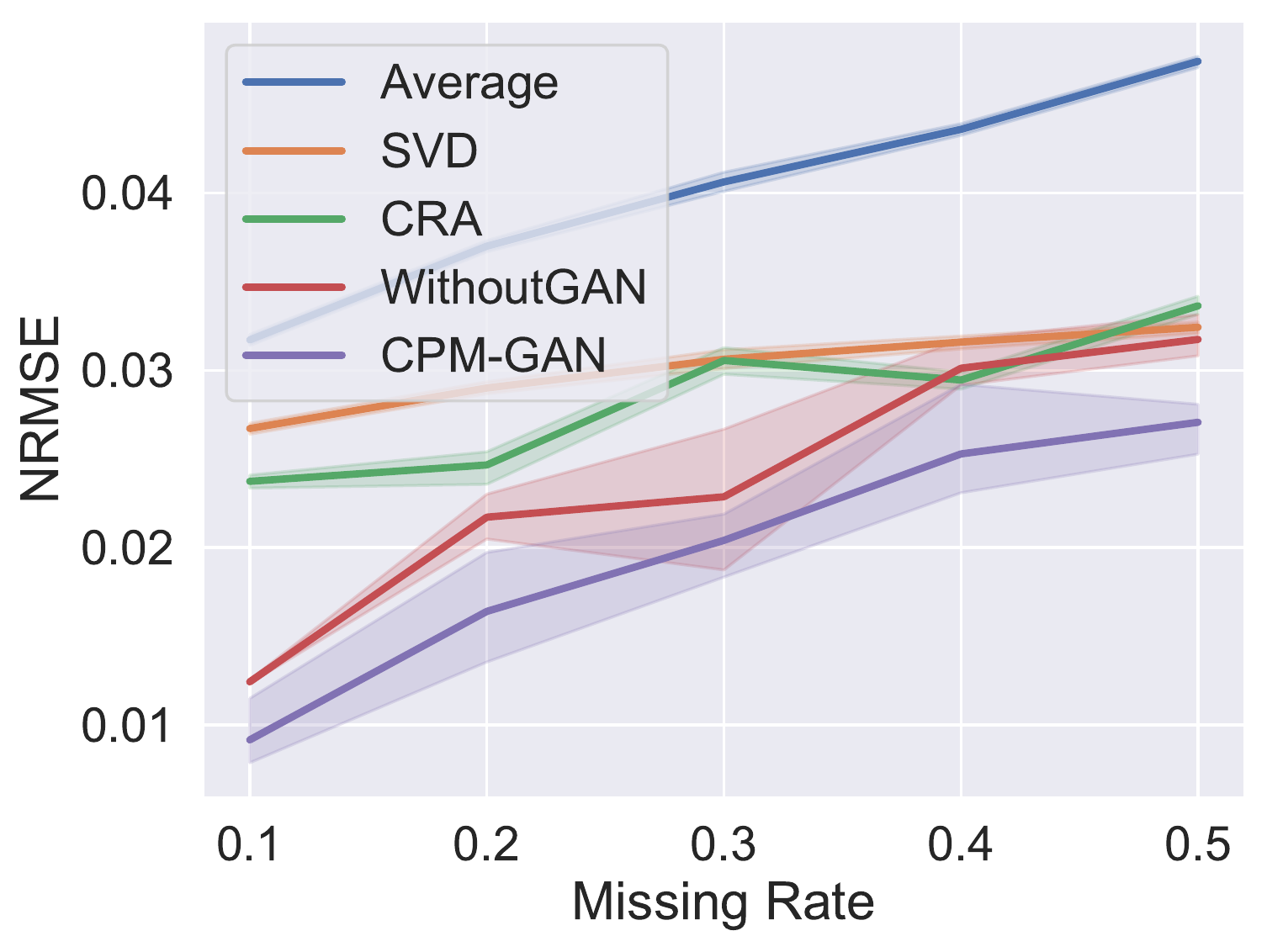}
\end{minipage}}
\centering
\subfigure[Football]{
\begin{minipage}[t]{0.32\linewidth}
\centering
\includegraphics[width=2.4in,height = 1.8in]{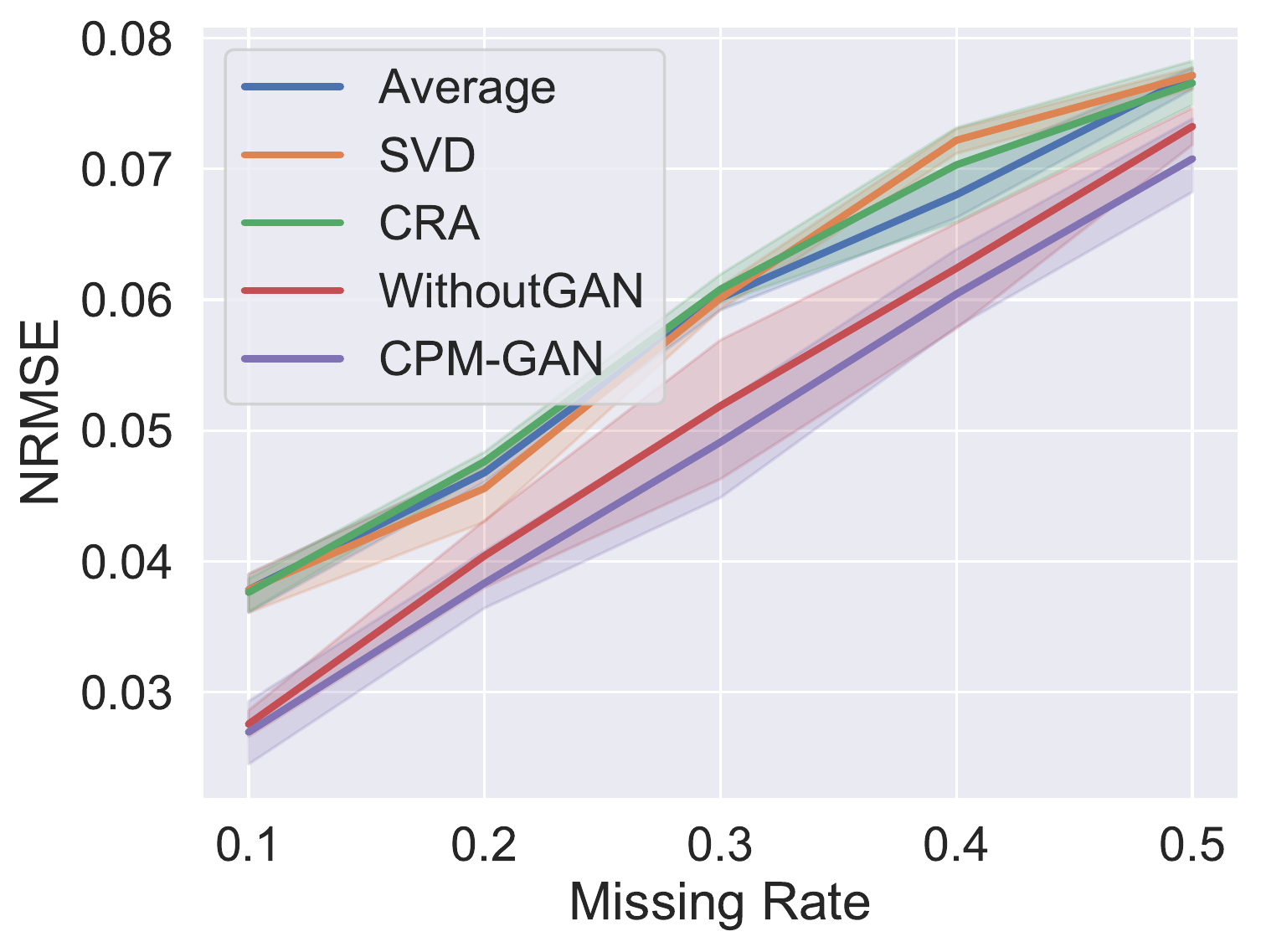}
\end{minipage}}
\centering
\subfigure[Politics]{
\begin{minipage}[t]{0.32\linewidth}
\centering
\includegraphics[width=2.4in,height = 1.8in]{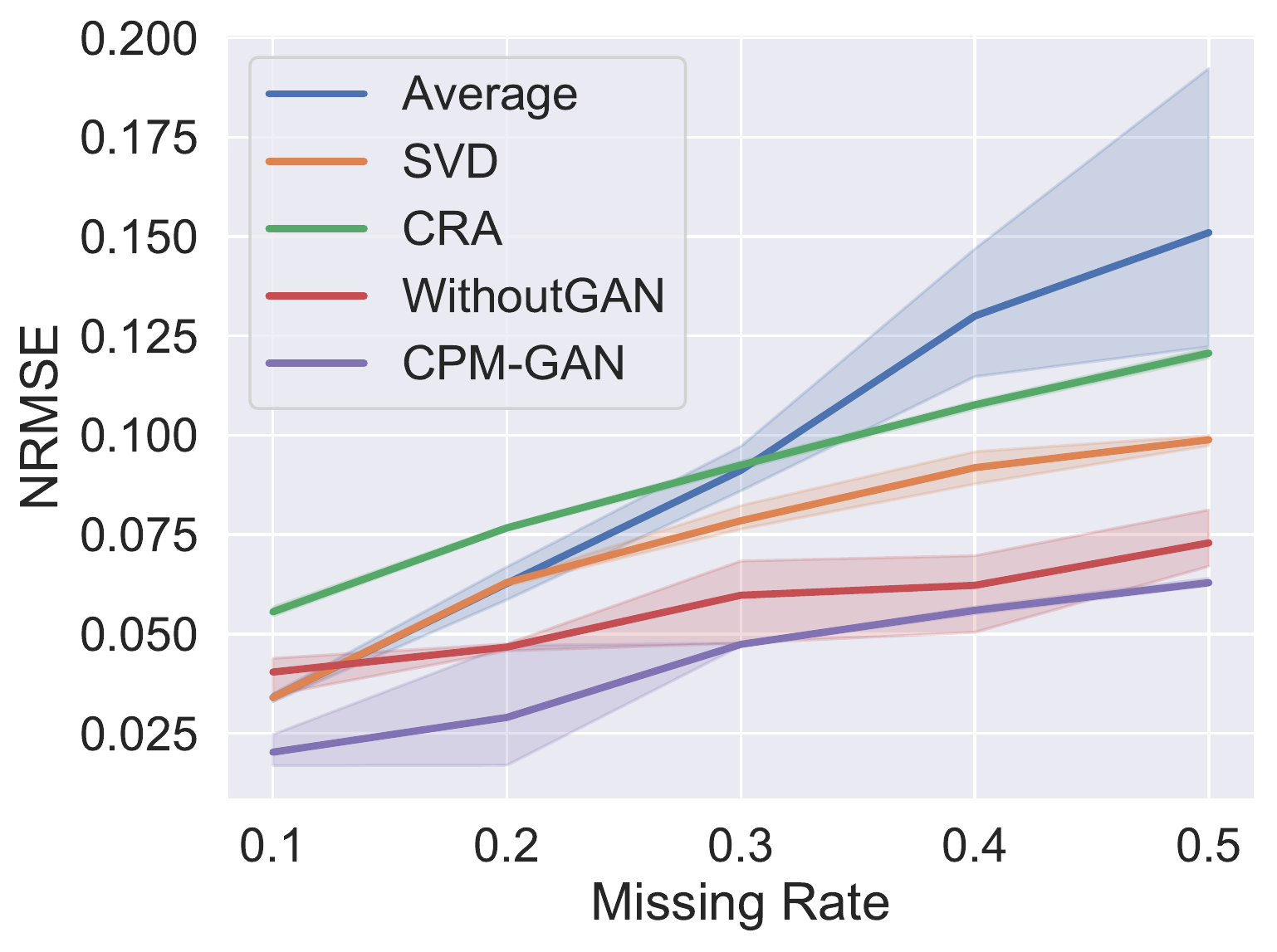}
\end{minipage}}
\caption{Imputation performance comparison with different missing rates ($\eta$).}
\label{fig:curves}
\end{figure*}

\section{Experiments}
\subsection{Experimental Settings}
We conduct experiments on the following datasets:
%
$\diamond$ \textbf{Handwritten}\footnote{https://archive.ics.uci.edu/ml/datasets/Multiple+Features}: This dataset contains 10 categories from digits `0' to `9', and 200 images in each category with six types of image features are used. $\diamond$ \textbf{Animal} \cite{LampertNH14} : The dataset consists of 10,158 images from 50 classes with two types of deep features extracted with DECAF \cite{krizhevsky2012imagenet} and VGG19 \cite{simonyan2014very}. $\diamond$ \textbf{CUB} \cite{wah2011caltech}: This dataset contains different categories of birds, where the first 10 categories are used. Deep visual features from GoogLeNet and text features using doc2vec \cite{le2014distributed} are employed as two views. $\diamond$ \textbf{3Sources-complete}\footnote{http://mlg.ucd.ie/datasets/3sources.html}: is collected from three online news sources: BBC, Reuters, and Guardian. In total 169 samples of stories are used, which are reported by all three sources. $\diamond$ \textbf{Football}\footnote{http://mlg.ucd.ie/aggregation/index.html}: A collection of 248 English Premier League football players and clubs active on Twitter. The disjoint ground truth communities correspond to 20 individual clubs in the league. $\diamond$ \textbf{Politics}\footnote{http://mlg.ucd.ie/aggregation/index.html}: A collection of Irish politicians and political organisations, assigned to seven disjoint ground truth groups according to their affiliation. For both football and politics datasets, 9 views are provided including follows, followedby, mentions, mentionedby, retweets, retweetedby, listmerged, lists, and tweets. $\diamond$ \textbf{ADNI}\footnote{http://www.loni.usc.edu/ADNI}: The dataset consists of 774 subjects from ADNI-1, including 226 normal controls (NC), 362 MCI and 186 AD subjects. There are only 379 subjects with complete MRI and PET data, including 101 NC, 185 MCI, and 93-AD, where the missing rate is up to 0.26. We use 93-dimensional ROI-based features from both MRI and PET data, respectively. $\diamond$ \textbf{3Sources-partial}\footnote{http://erdos.ucd.ie/datasets/3sources.html}: is collected from three well-known online news sources: BBC, Reuters, and Guardian, in which some stories are not reported by all three sources (i.e., view missing). The missing rate of 3Sources-partial is 0.24.

We compare the proposed CPM-Nets with the following methods: (1) \textbf{FeatConcate} simply concatenates multiple types of features from different views. (2) \textbf{CCA} \cite{hotelling1936relations} maps multiple types of features into one common space, and subsequently concatenates the low-dimensional features of different views. (3) \textbf{DCCA} (Deep Canonical Correlation Analysis) \cite{andrew2013deep} learns low-dimensional features with neural networks and concatenates them. (4) \textbf{DCCAE} (Deep Canonical Correlated AutoEncoders) \cite{Wang2016On} employs autoencoders for common representations, and then combines these projected low-dimensional features together. (5) \textbf{KCCA} (Kernelized CCA) \cite{akaho2006kernel} employs feature mappings induced by positive-definite kernels. (6) \textbf{MDcR} (Multi-view Dimensionality co-Reduction) \cite{zhang2017flexible} applies kernel matching to regularize the dependence across multiple views and projects each view into a low-dimensional space. (7) \textbf{DMF-MVC} (Deep Semi-NMF for Multi-View Clustering) \cite{zhao2017multi} utilizes a deep structure through semi-nonnegative matrix factorization to seek a common feature representation. (8) \textbf{ITML} (Information-Theoretic Metric Learning) \cite{davis2007information} characterizes the metric using a Mahalanobis distance function and solves the problem as a particular Bregman optimization. (9) \textbf{LMNN} (Large Margin Nearest Neighbors) \cite{weinberger2009distance} searches a Mahalanobis distance metric to optimize the $k$-nearest neighbours classifier. For metric learning methods, the original features of multiple views are concatenated, and then the new representation can be obtained with the projection induced by the learned metric matrix.

\begin{figure*}[!ht]
\centering
\subfigure[Animal]{
\begin{minipage}[t]{0.32\linewidth}
\centering
\includegraphics[width=2.4in,height = 1.8in]{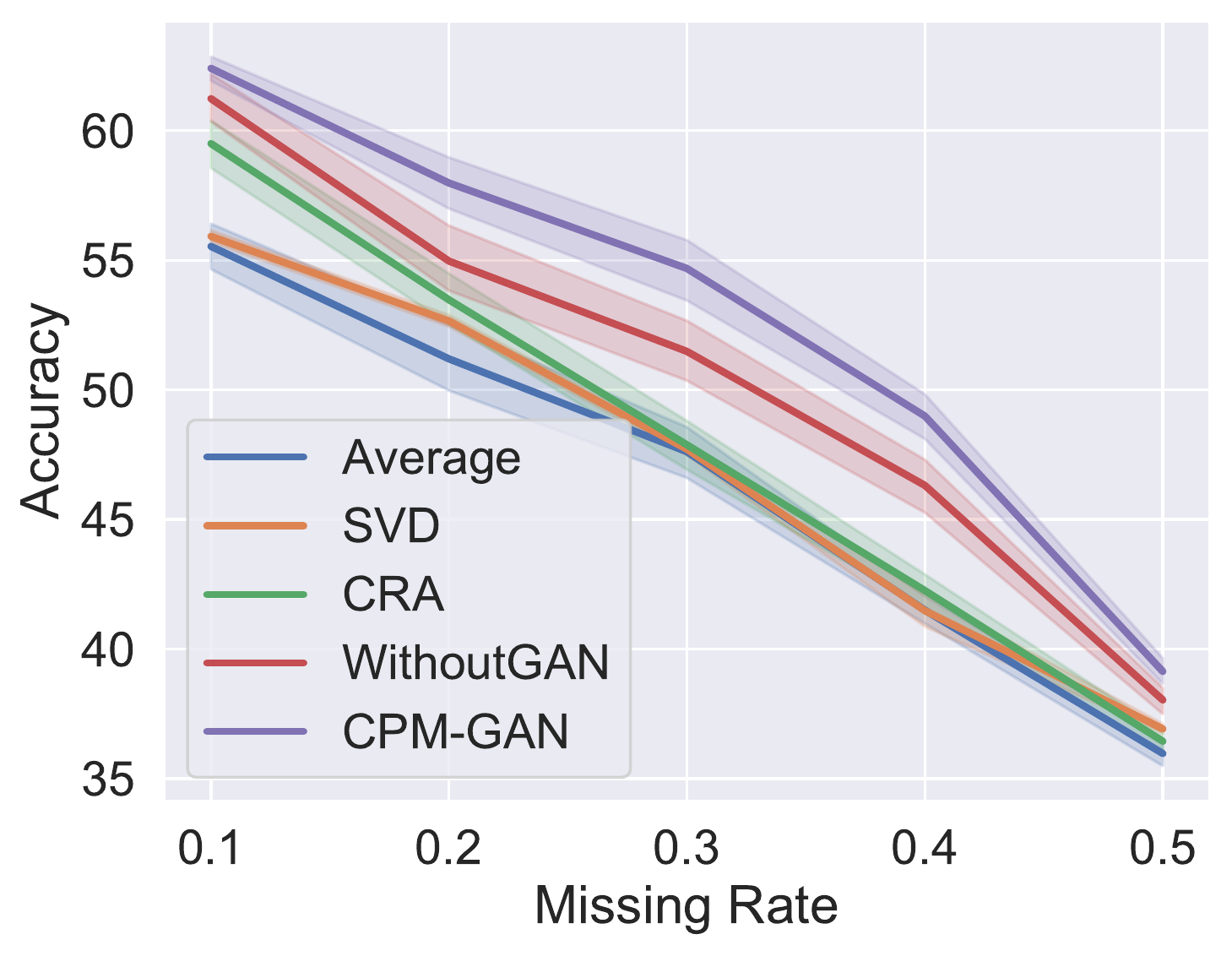}
\includegraphics[width=2.4in,height = 1.8in]{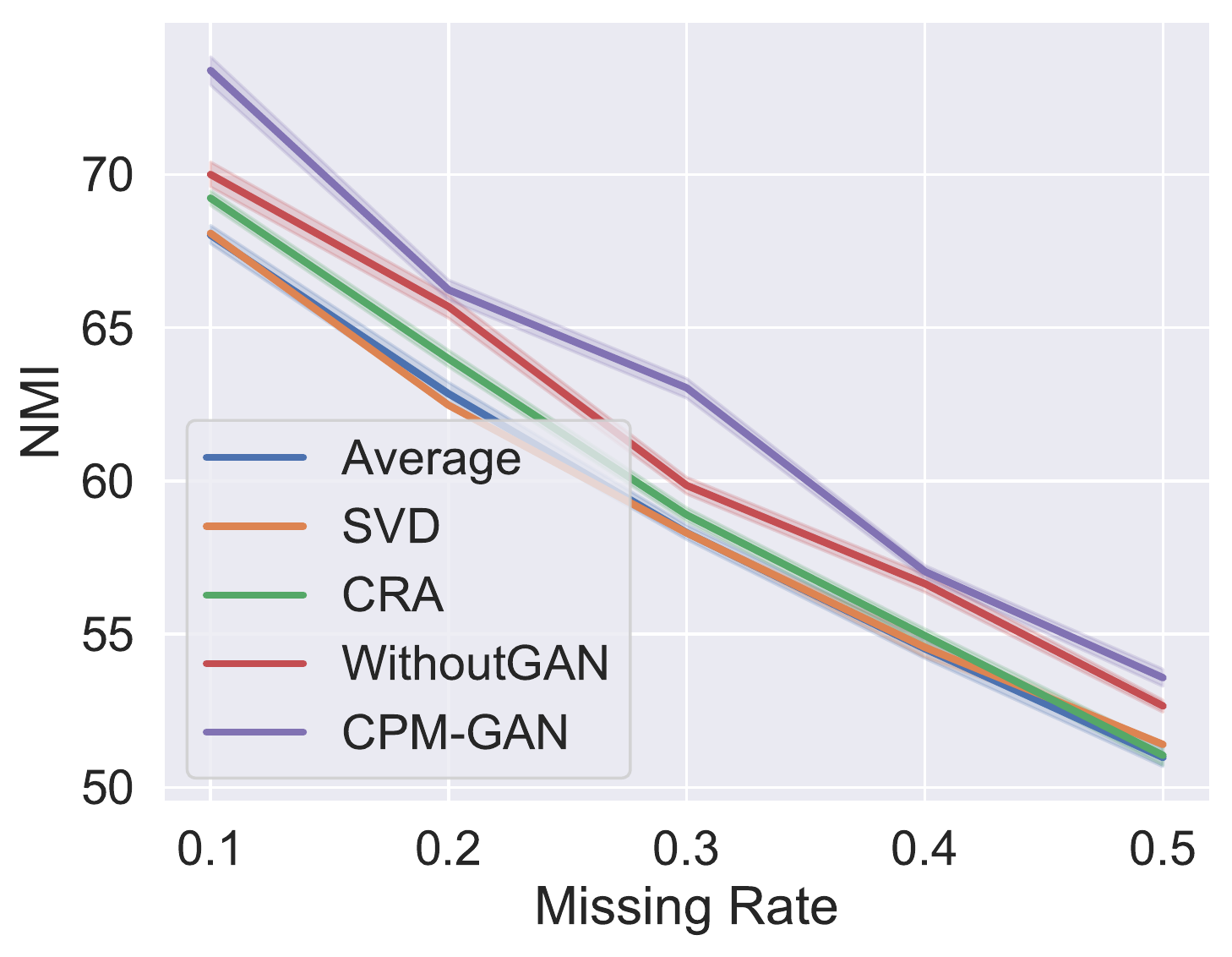}
\end{minipage}}
\centering
\subfigure[Handwritten]{
\begin{minipage}[t]{0.32\linewidth}
\centering
\includegraphics[width=2.4in,height = 1.8in]{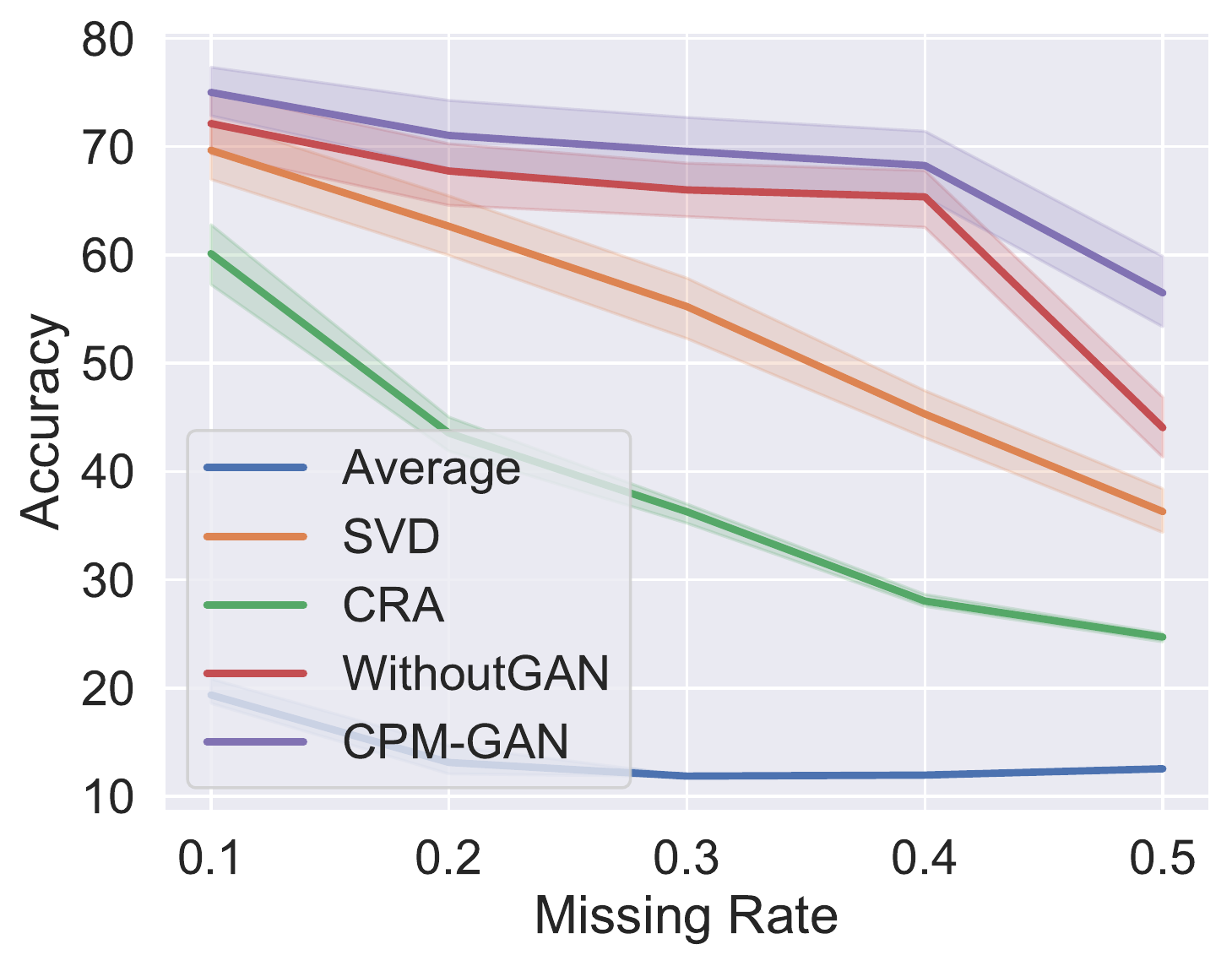}
\includegraphics[width=2.4in,height = 1.8in]{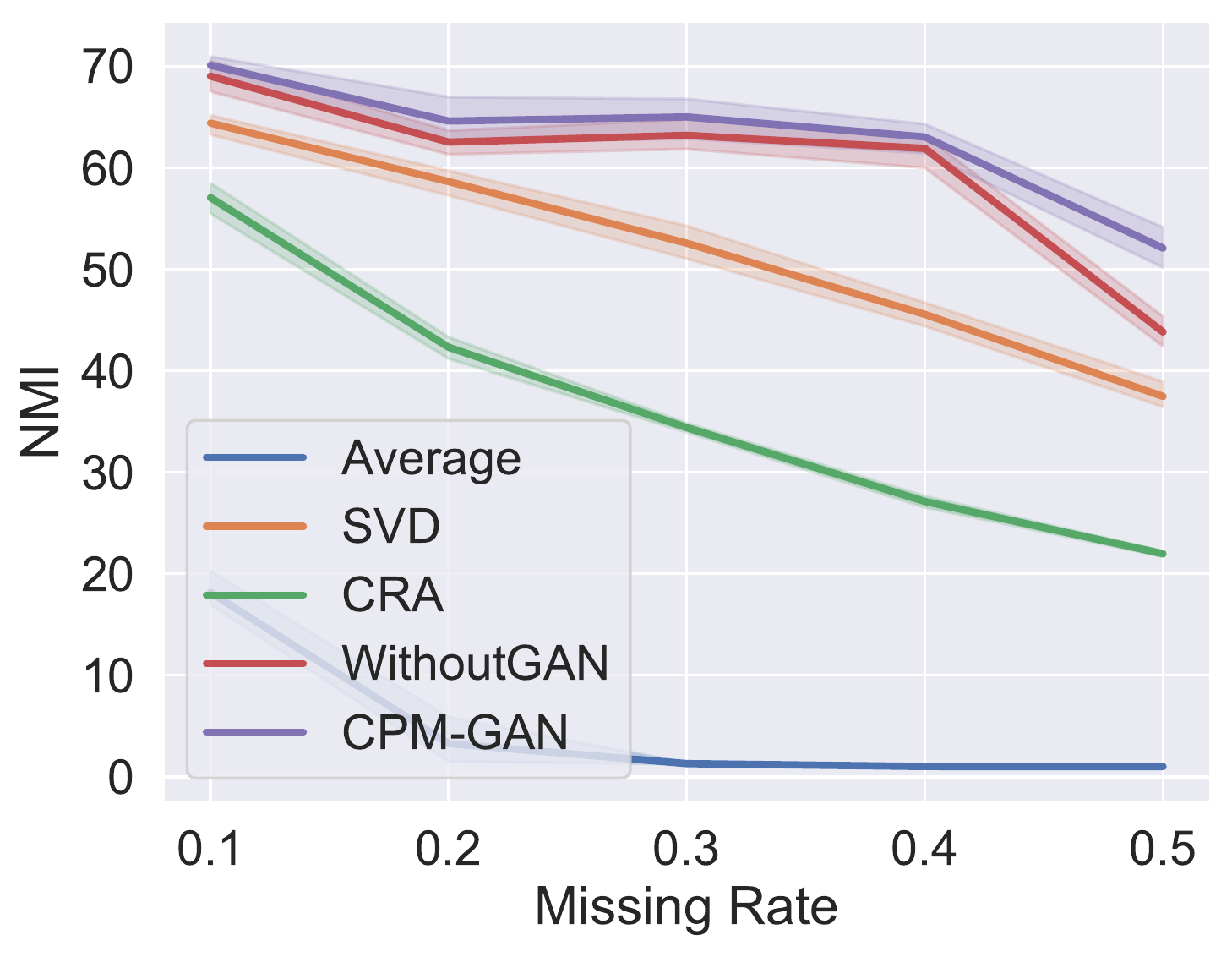}
\end{minipage}}
\subfigure[CUB]{
\begin{minipage}[t]{0.32\linewidth}
\centering
\includegraphics[width=2.4in,height = 1.8in]{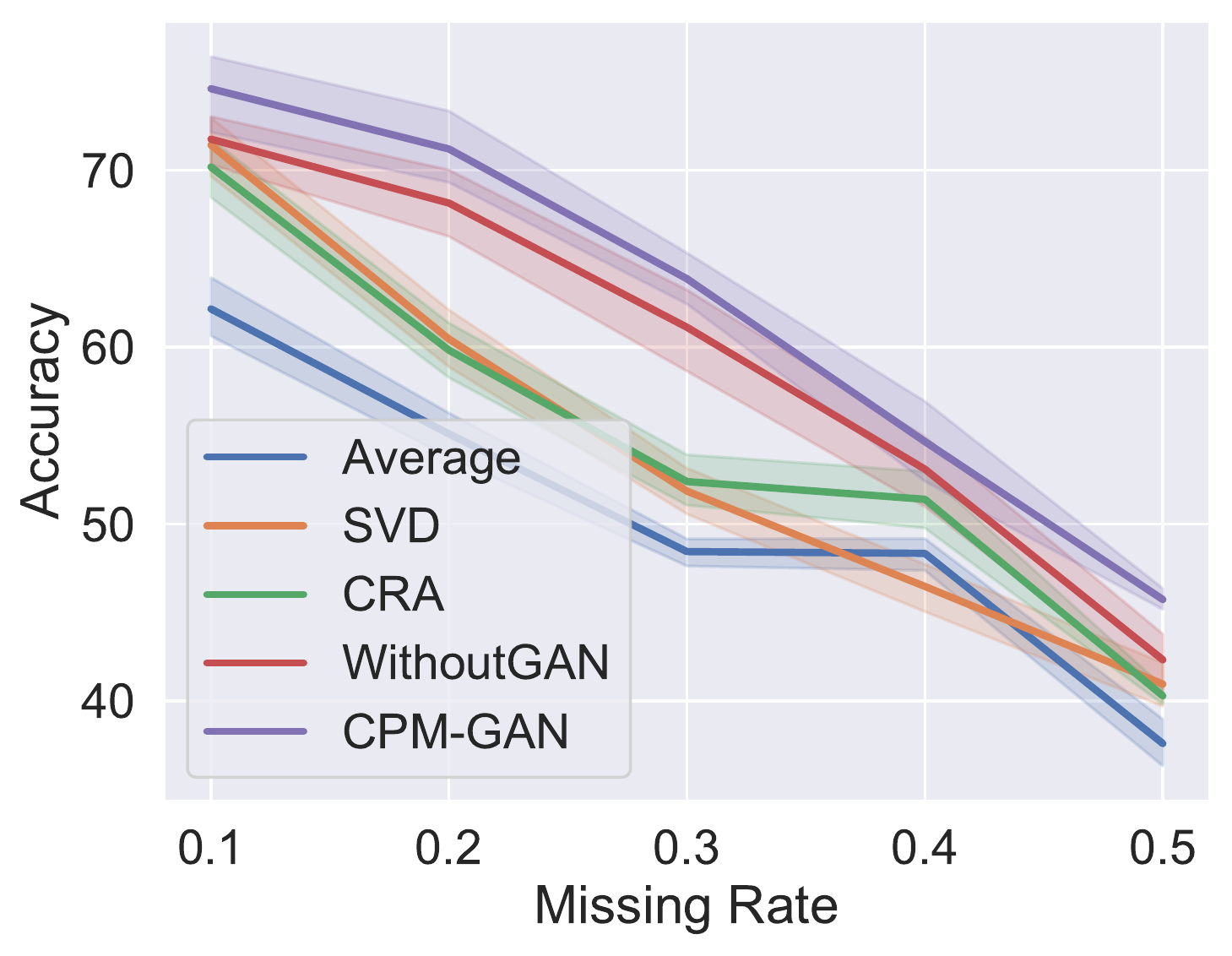}
\includegraphics[width=2.4in,height = 1.8in]{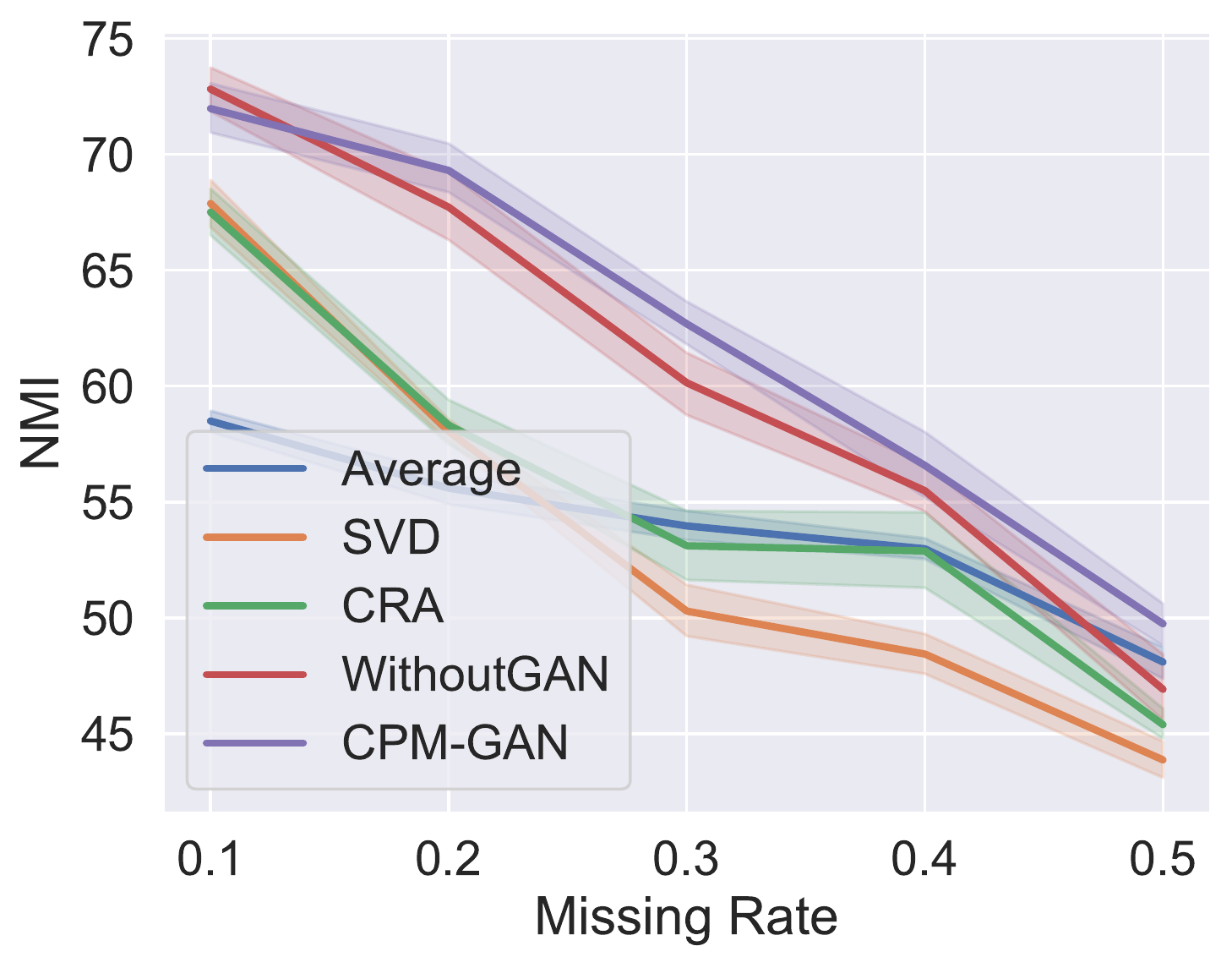}
\end{minipage}}
\centering
\subfigure[3Sources-complete]{
\begin{minipage}[t]{0.32\linewidth}
\centering
\includegraphics[width=2.4in,height = 1.8in]{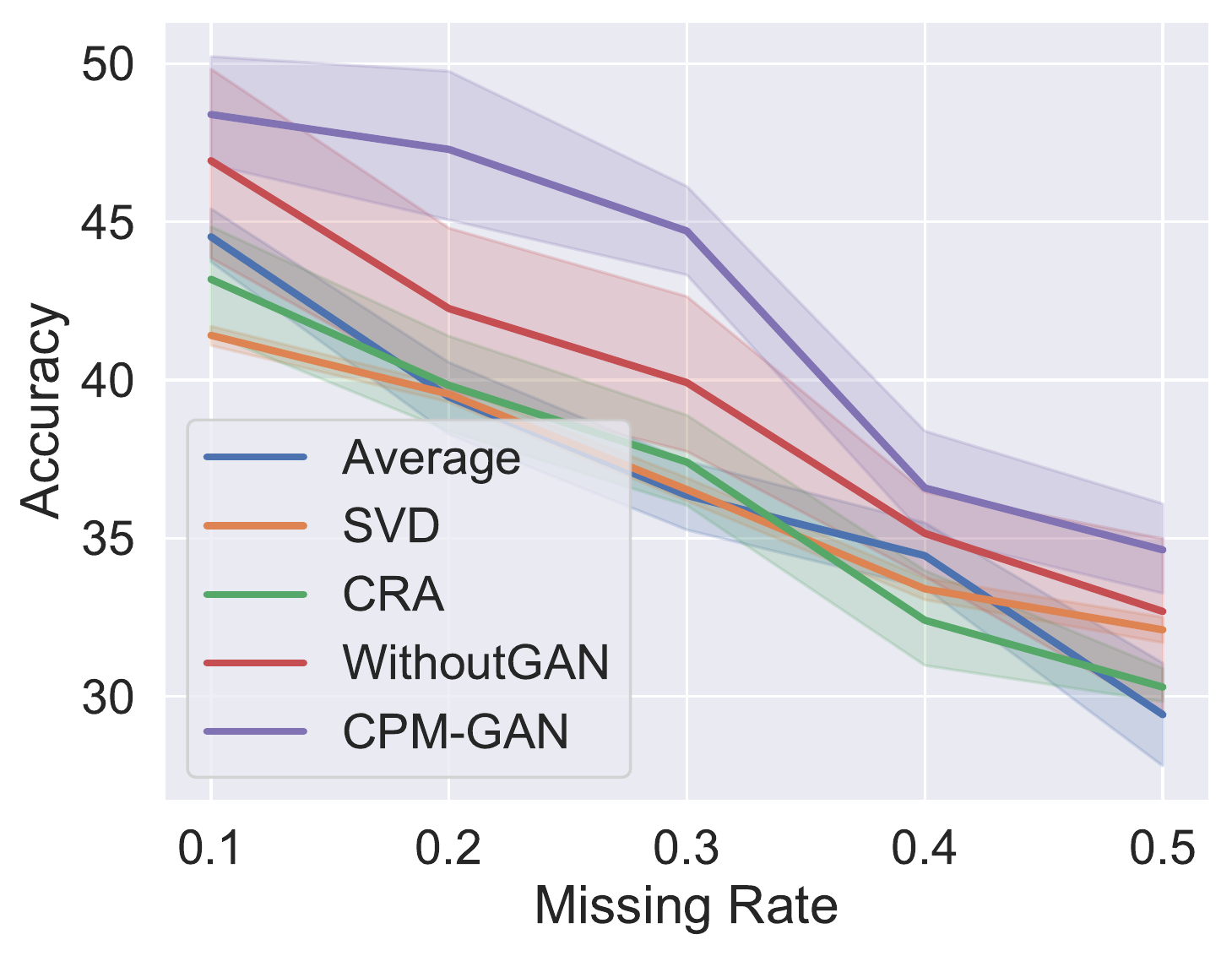}
\includegraphics[width=2.4in,height = 1.8in]{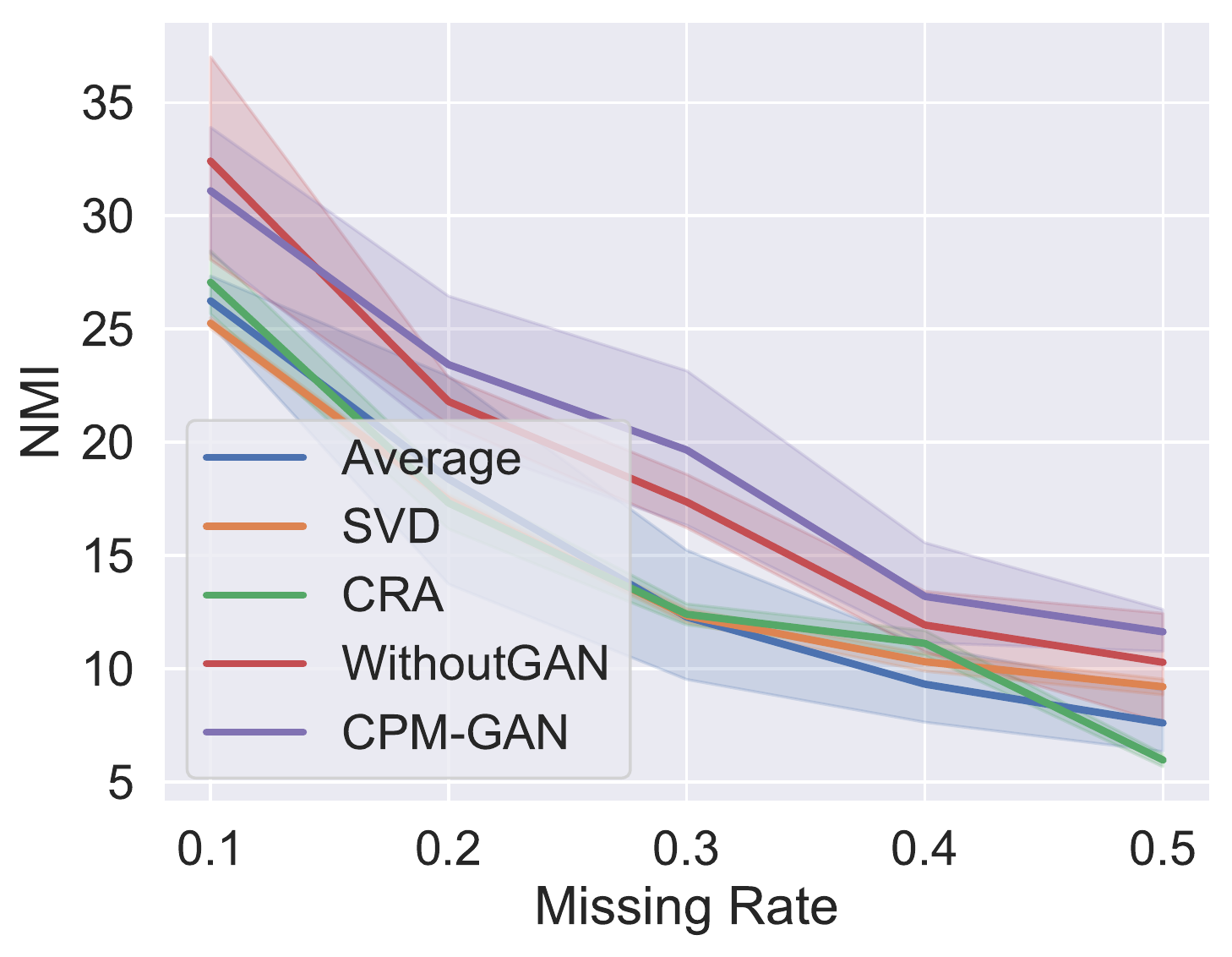}
\end{minipage}}
\centering
\subfigure[Football]{
\begin{minipage}[t]{0.32\linewidth}
\centering
\includegraphics[width=2.4in,height = 1.8in]{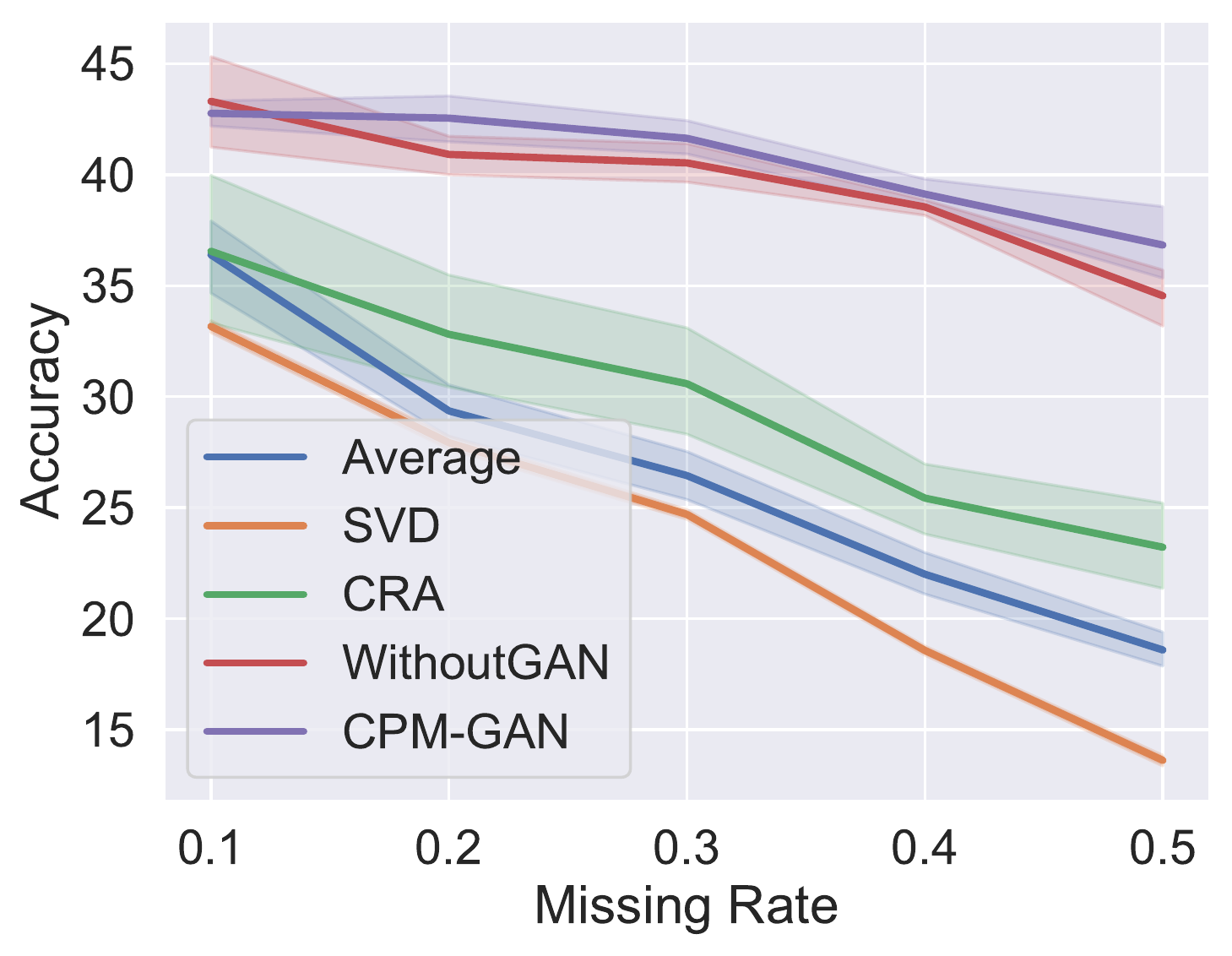}
\includegraphics[width=2.4in,height = 1.8in]{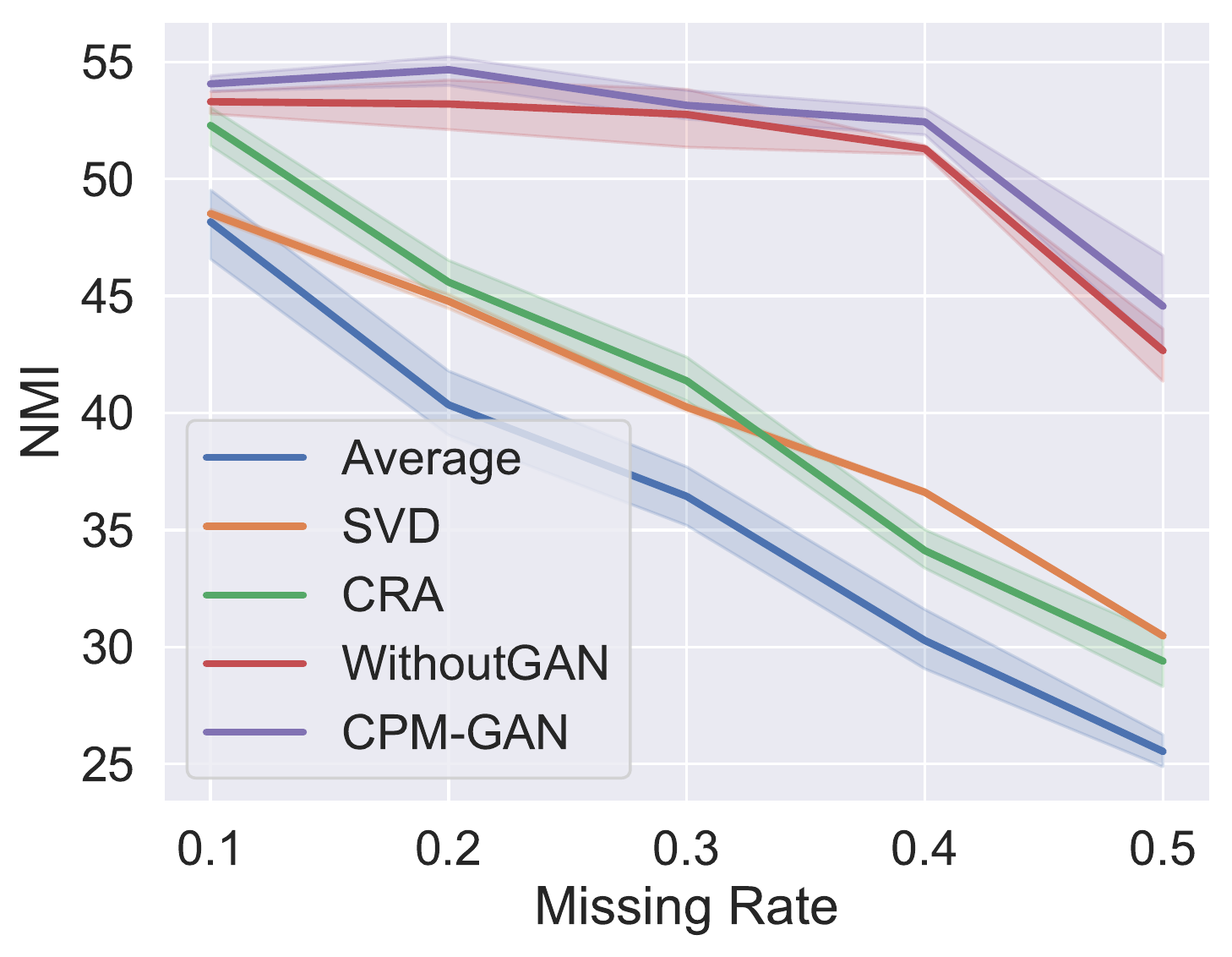}
\end{minipage}}
\centering
\subfigure[Politics]{
\begin{minipage}[t]{0.32\linewidth}
\centering
\includegraphics[width=2.4in,height = 1.8in]{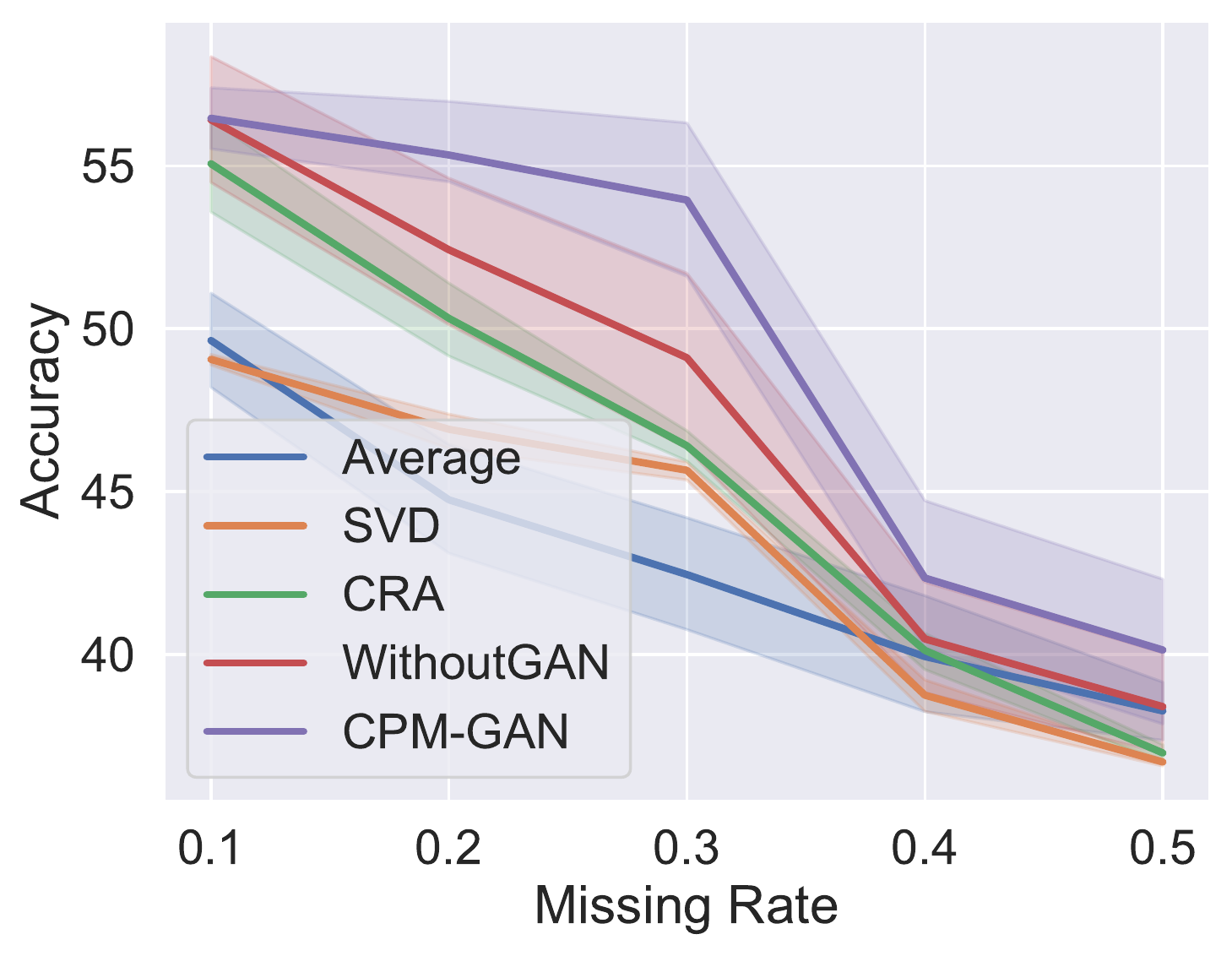}
\includegraphics[width=2.4in,height = 1.8in]{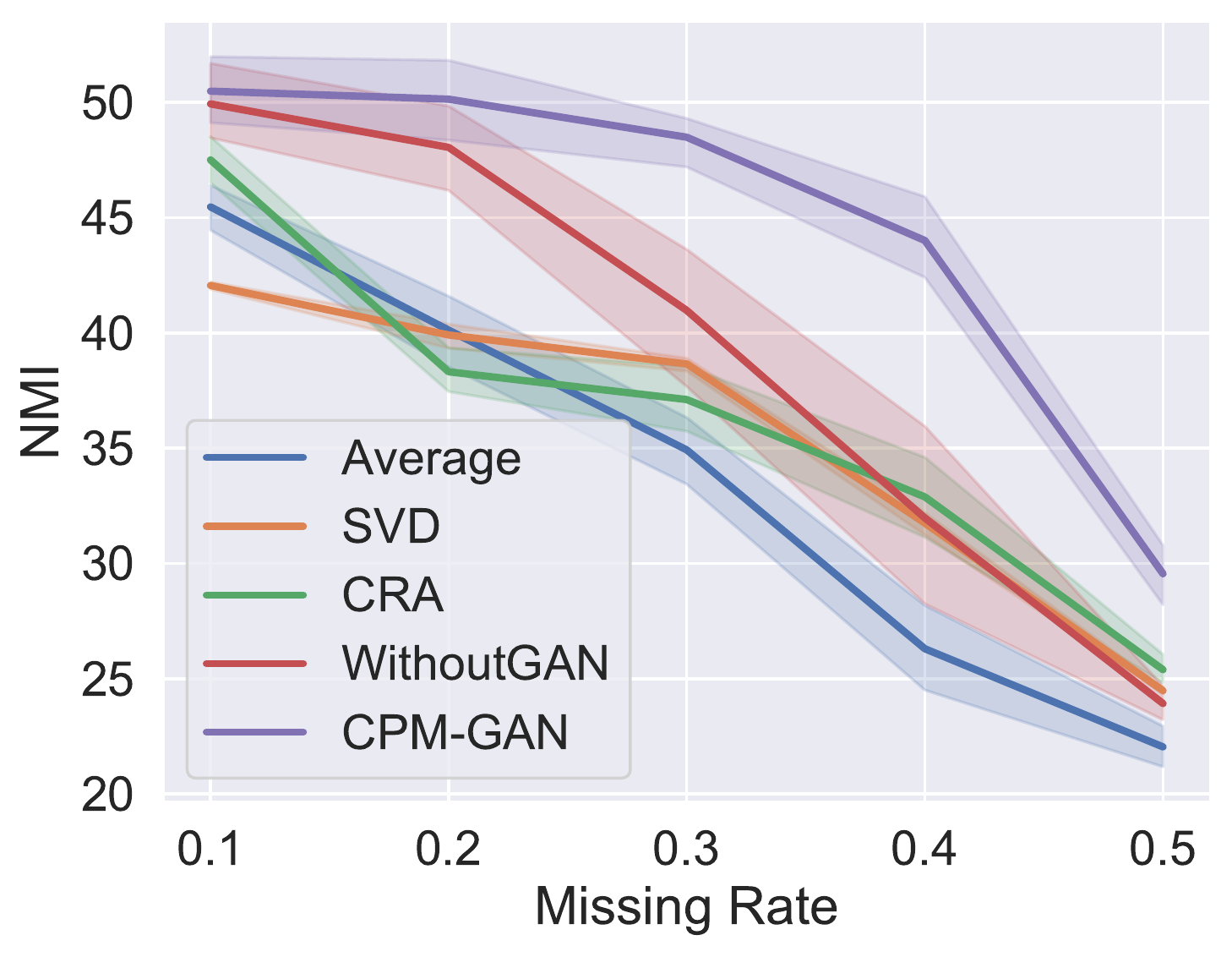}
\end{minipage}}
\caption{Clustering performance comparison under different missing rates ($\eta$).}
\label{fig:clustering}
\end{figure*}

\begin{figure*}[htbp]
\centering
\subfigure[ADNI]{
\begin{minipage}[t]{0.23\linewidth}
\centering
\includegraphics[width=1.6in,height = 1.4in]{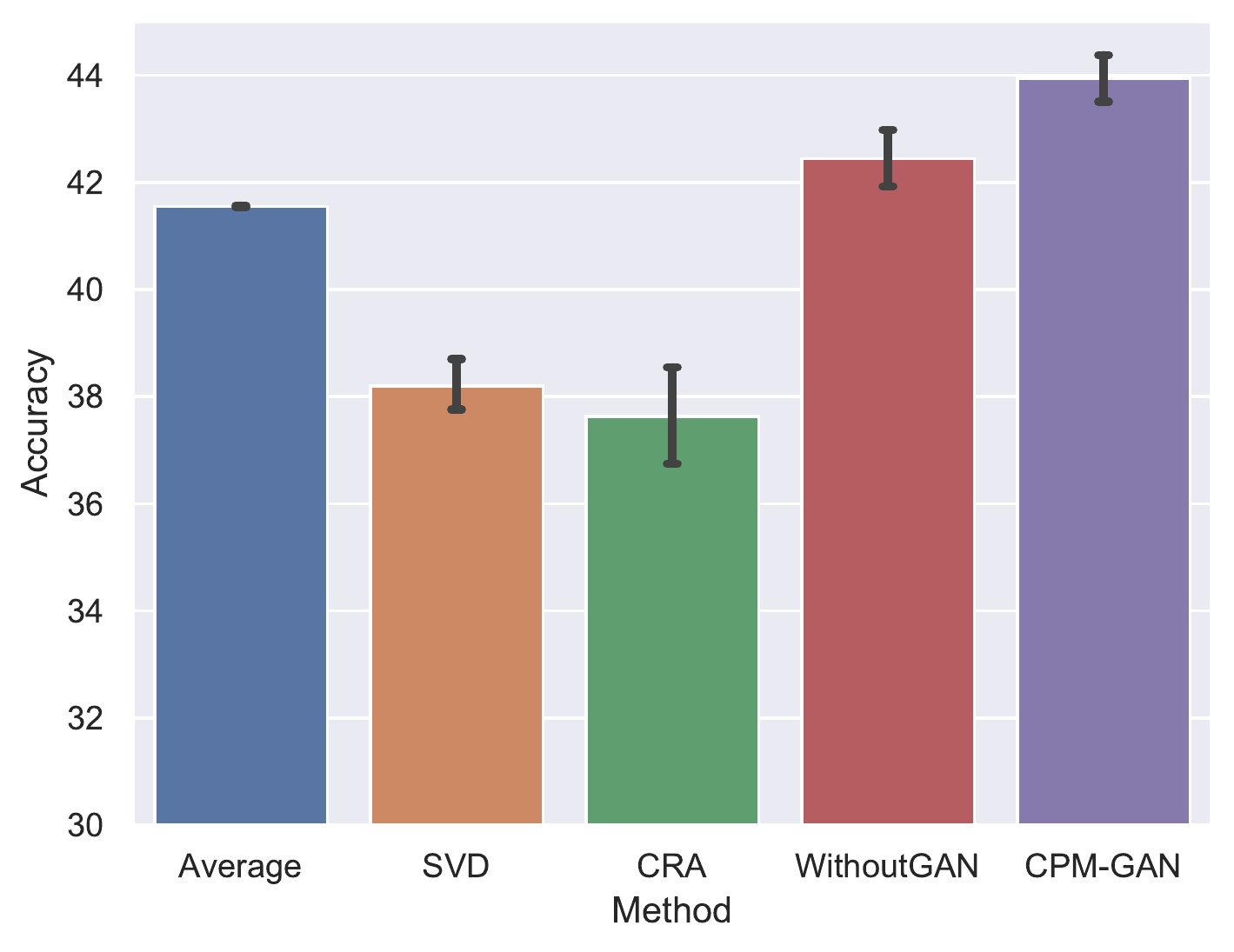}
\end{minipage}
\begin{minipage}[t]{0.23\linewidth}
\centering
\includegraphics[width=1.6in,height = 1.4in]{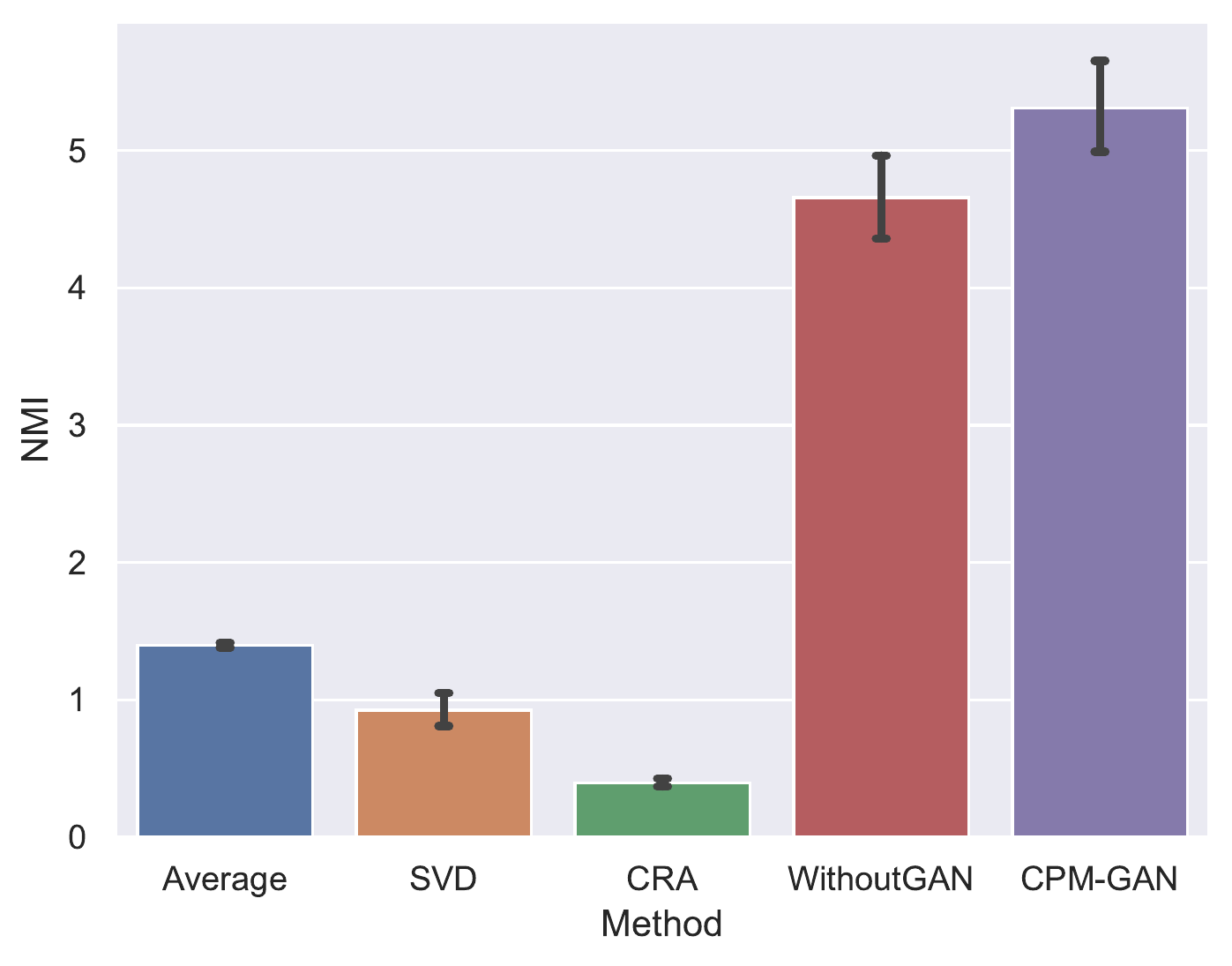}
\end{minipage}
}
\subfigure[3Sources-partial]{
\begin{minipage}[t]{0.23\linewidth}
\centering
\includegraphics[width=1.6in,height = 1.4in]{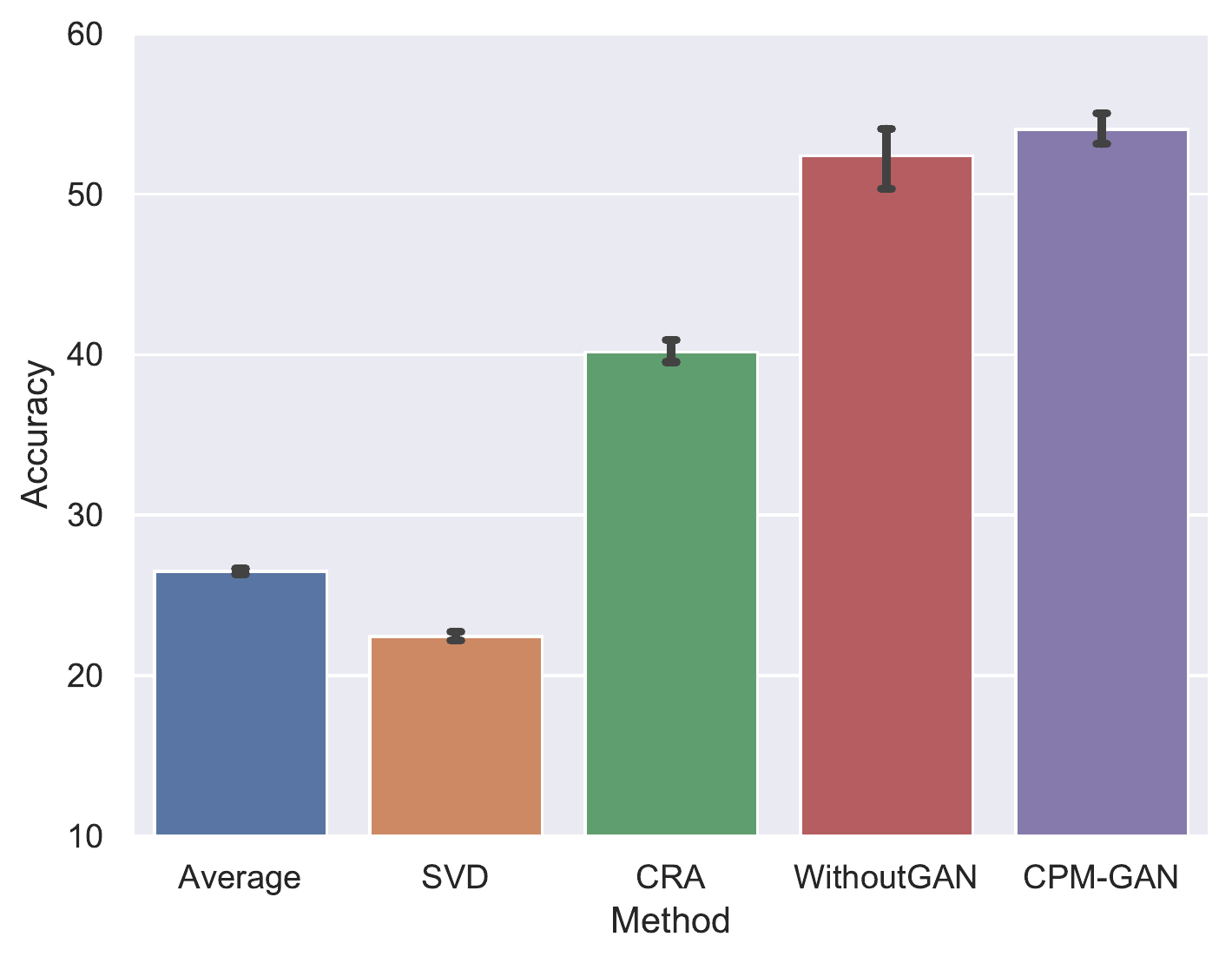}
\end{minipage}
\begin{minipage}[t]{0.23\linewidth}
\centering
\includegraphics[width=1.6in,height = 1.4in]{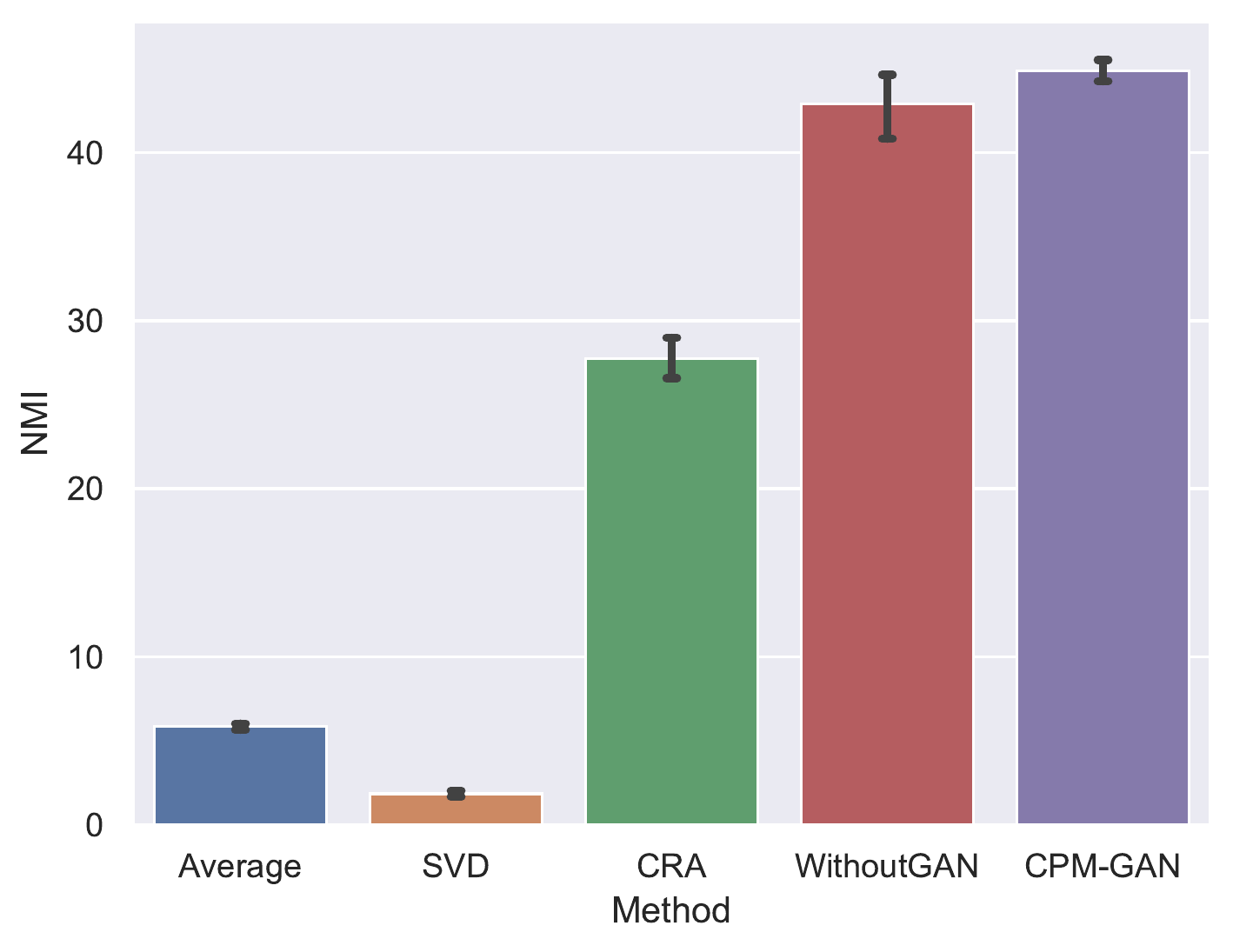}
\end{minipage}
}
\centering
\caption{Clustering performance comparison on real-world missing data.}
\label{fig:clustering-realworld}
\end{figure*}

\begin{figure*}[!ht]

\centering
\subfigure[CUB]{
\begin{minipage}[t]{0.23\linewidth}
\centering
\includegraphics[width=1.6in,height = 1.4in]{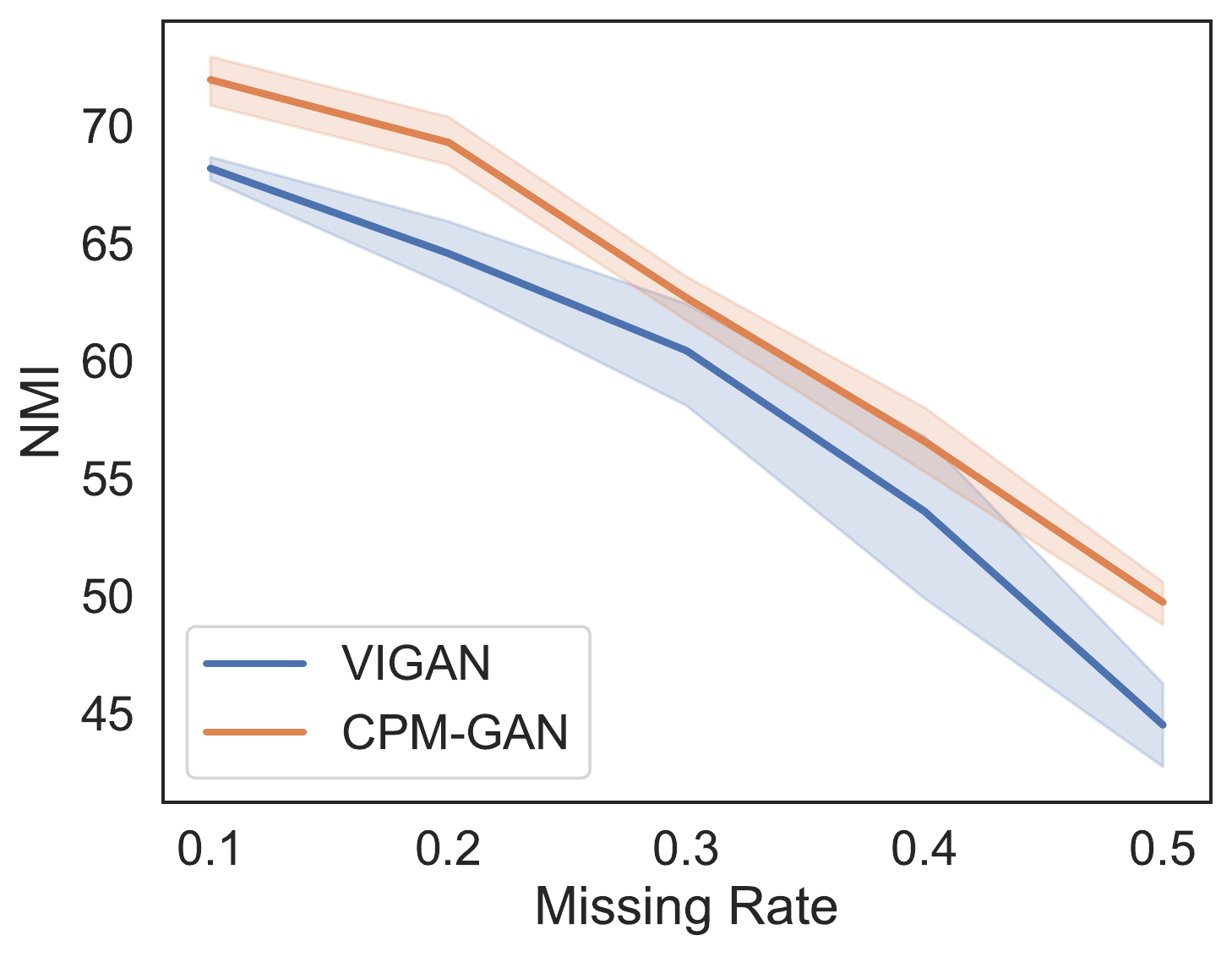}
\end{minipage}
\begin{minipage}[t]{0.23\linewidth}
\centering
\includegraphics[width=1.6in,height = 1.4in]{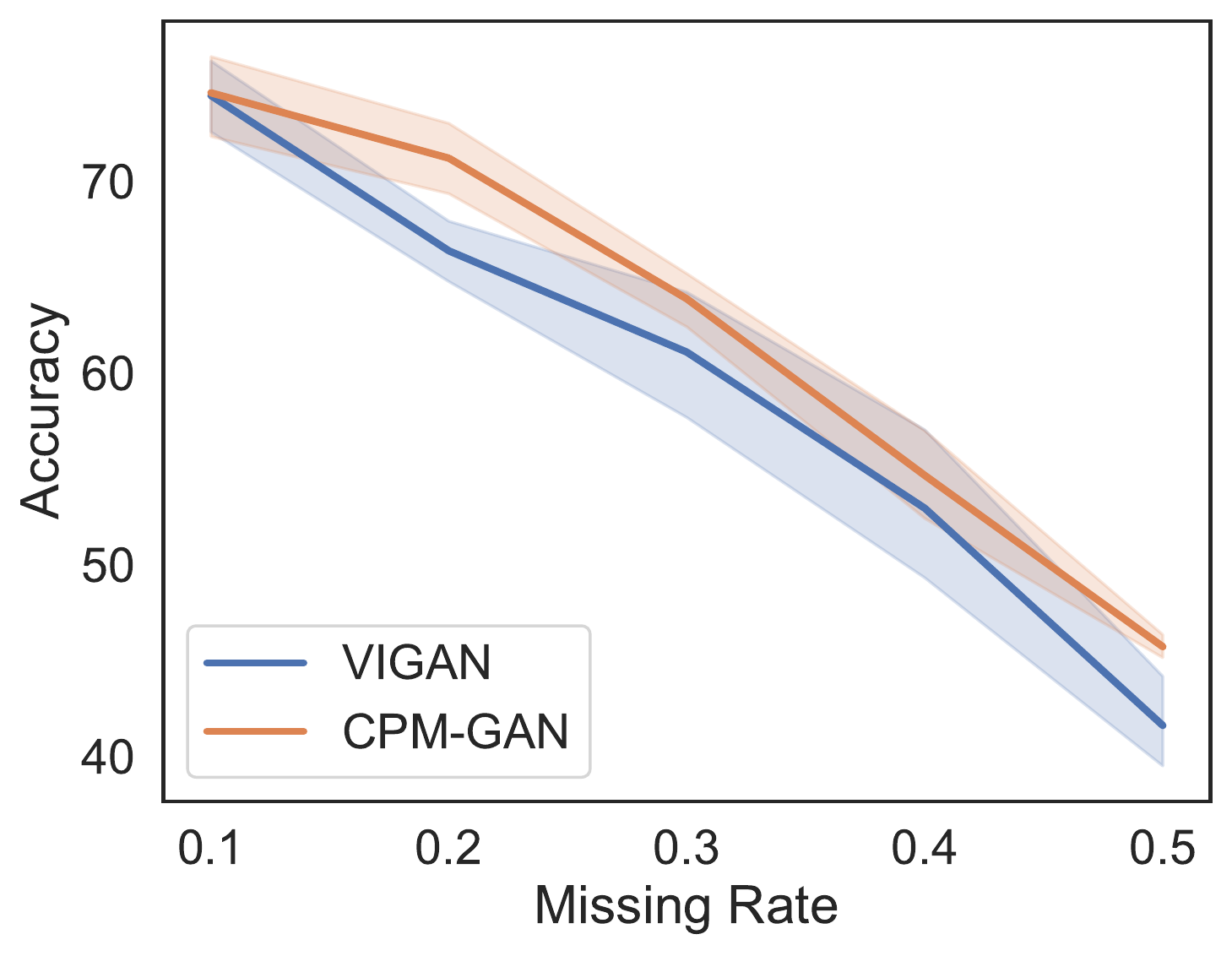}
\end{minipage}
}
\subfigure[Animal]{
\begin{minipage}[t]{0.23\linewidth}
\centering
\includegraphics[width=1.6in,height = 1.4in]{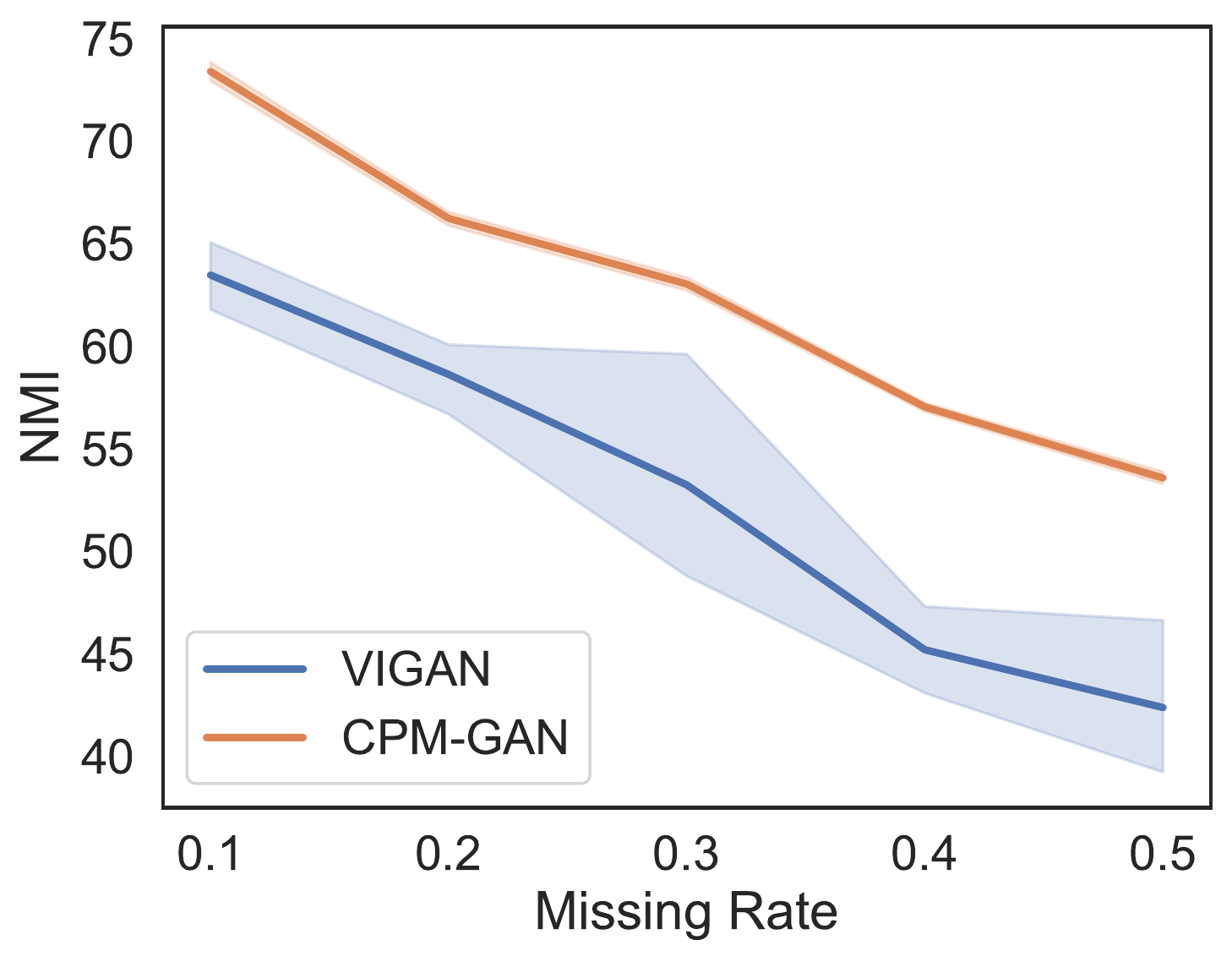}
\end{minipage}
\begin{minipage}[t]{0.23\linewidth}
\centering
\includegraphics[width=1.6in,height = 1.4in]{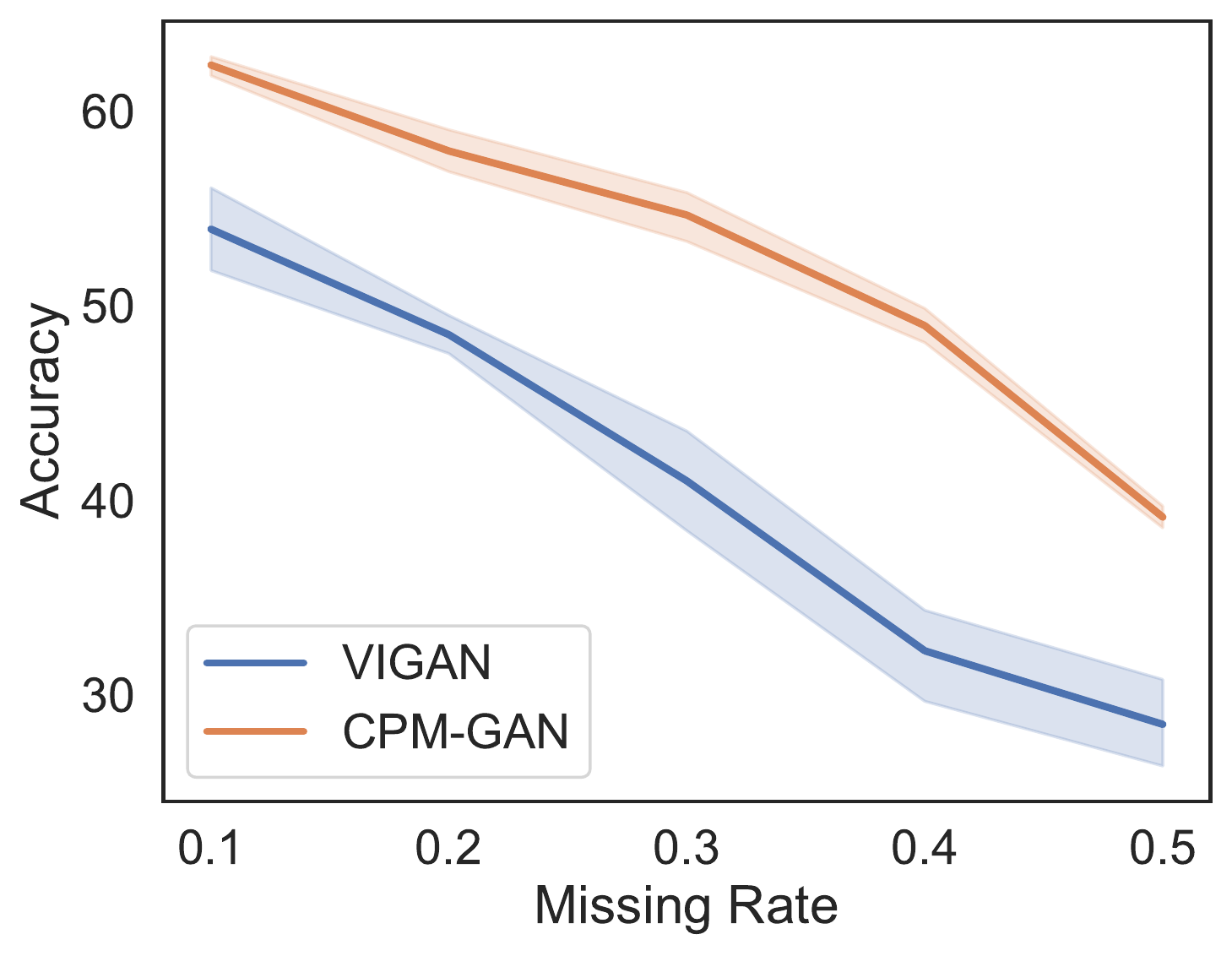}
\end{minipage}
}
\caption{Clustering performance comparison with VIGAN.}
\label{fig:vigan}
\end{figure*}

For all methods, we tune the parameters with five-fold cross validation. For CCA-based methods, we select two views for the best performance. For our CPM-Nets, we set the dimensionality ($K$) of the latent representation from $\{64, 128, 256\}$ and tune the parameter $\lambda $ from the set $\{0.1,1,10\}$ for all datasets. We run each method 10 times and report the average values and standard deviations.

\subsection{Network Architectures and Parameter Settings}
For our CPM-Nets, we employ the fully connected networks equipped with sigmoid activation for all datasets, and ${\ell}_2$-norm regularization is used with the value of the tradeoff parameter being $0.001$. The dimensionalities of the input, hidden and output layers are denoted as $K$, $M$ and $D$, respectively. The dimensionality of the input layer (\emph{i.e.}, the dimensionality of the latent representation) is selected from the set $\{64,128,256\}$. The main architectures are the same with detailed differences described as follows:
\begin{itemize}
\item{
\textbf{Handwritten}. We employ three-layer (\emph{i.e.}, input/hidden/ouput layers) fully connected networks. The dimensionalities of the input, hidden and output layers are $K=64$, $M=200$ and $D$ (240, 76, 216, 47, 64 and 6 respectively for six views).  The learning rate is set as $0.001$.}
\item{
\textbf{CUB}. We employ two-layer (\emph{i.e.}, input/ouput layers) fully connected networks. The dimensionalities of the input and output layers are $K=128$, and $D$ (1024 and 300 respectively for two views).  The learning rate is set as $0.01$.}
\item{
\textbf{Animal}. We employ four-layer (\emph{i.e.}, input/two hidden/ouput layers) fully connected networks. The dimensionalities of the input, hidden and output layers are $K=256$, $M$ (512 and 1024 respectively for two hidden layers) and $D$ (4096 for both views).  The learning rate is set as $0.001$.}

\item{
\textbf{Football}. We employ three-layer (\emph{i.e.}, input/hidden/ouput layers) fully connected networks. The dimensionalities of the input, hidden and output layers are $K=256$, $M=50$ and $D$ (248, 248, 248, 248, 248, 248, 3601, 7814 and 11806 respectively for 9 views).  The learning rate is set as $0.01$.}
\item{
\textbf{Politics}. We employ three-layer (\emph{i.e.}, input/hidden/ouput layers) fully connected networks. The dimensionalities of the input, hidden and output layers are $K =128$, $M=128$ and $D$ (348, 348, 348, 348, 348, 348, 1051, 1047 and 14377  respectively for 9 views).  The learning rate is set as $0.01$.}
\item{
\textbf{3Source-complete}. We employ three-layer (\emph{i.e.}, input/hidden/ouput layers) fully connected networks. The dimensionalities of the input, hidden and output layers are $K =128$, $M=256$ and $D$ (3560, 3631 and 3068 respectively for 3 views).  The learning rate is set as $0.01$.}
\item{
\textbf{ADNI}. We employ three-layer (\emph{i.e.}, input/hidden/ouput layers) fully connected networks. The dimensionalities of the input, hidden and output layers are $K =128$, $M=50$ and $D$ (93 for both views). The learning rate is set as $0.01$.}
\item{
\textbf{3Source-partial}. We employ three-layer (\emph{i.e.}, input/hidden/ouput layers) fully connected networks. The dimensionalities of the input, hidden and output layers are $K =128$, $M=128$ and $D$ (3560, 3631 and 3068 respectively for 3 views).  The learning rate is set as $0.01$.}

\end{itemize}

\subsection{Supervised Experimental Results}
Firstly, we evaluate our algorithm by comparing it with state-of-the-art multi-view representation learning methods, investigating the performance with respect to various missing rates. The missing rate is defined as  $\eta = \frac{\sum_v M_v}{V \times N}$, where $M_v$ indicates the number of samples without the $v$th view. Since datasets may be associated with different numbers of views, samples are randomly selected as missing multi-view ones, and the missing views are randomly selected by guaranteeing that at least one of them is available. As a result, partial multi-view data are obtained with diverse missing patterns. For compared methods, the missing views are filled with average values according to available samples within the same class. From the results in Fig.~\ref{fig:CPMNets}, we have the following observations: (1) Without missing, our algorithm achieves very competitive performance on all datasets, which validates the stability of our algorithm for complete multi-view data. (2) As the missing rate increases, the performance degradations of the compared methods are much larger than that of ours. Taking the results on CUB for example, ours and LMNN obtain an accuracy of 89.48\% and 86.27\%, respectively, while with increasing the missing rate, the performance gap becomes much larger; (3) Our model is rather robust to view-missing data, since our algorithm usually performs relatively promising with heavily missing cases. For example, the performance decline (on Handwritten) is less than 5\% with increasing the missing rate from $\eta = 0.0$ to $\eta = 0.3$. We also note that the standard deviations are relatively small ($<0.01$) on Animal, and the possible reason is that this dataset is much larger than others, producing more stable results.

\begin{figure}[!ht]
\center
\includegraphics[width=3in,height = 1.8in]{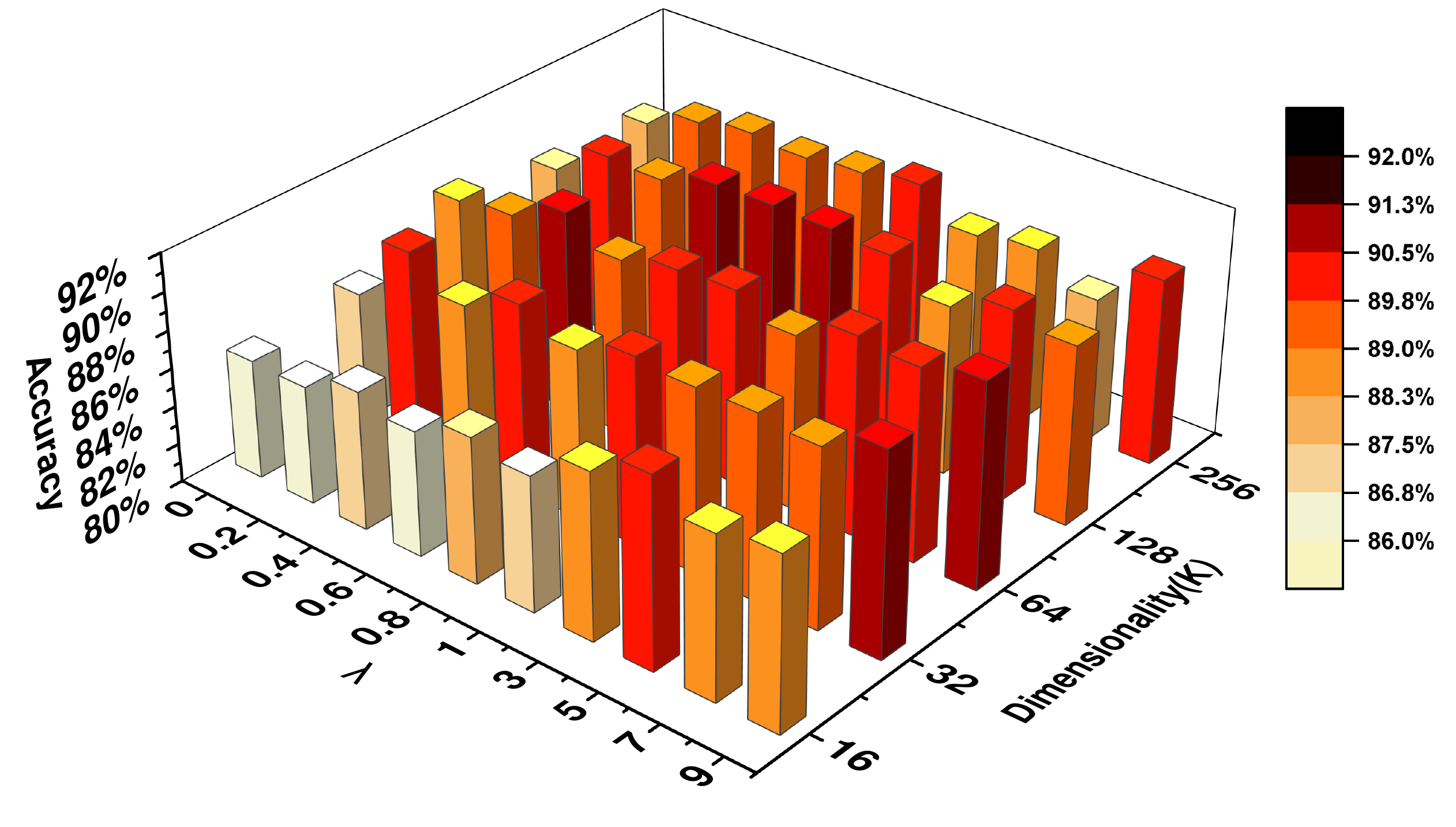}
\caption{Parameter tuning.}
\label{fig:parameter}
\end{figure}
Furthermore, we also fill the missing views with recently proposed imputation method - Cascaded Residual Autoencoder (CRA) \cite{tran2017missing}. Since CRA needs a subset of samples with complete views in training, we set $50\%$ of data as complete-view samples and the remaining are samples with missing views (missing rate $\eta = 0.5$). The comparison results are shown in Fig.~\ref{fig:mean-cra}. It is observed that filling with CRA is generally better than that of using average values due to capturing the correlation of different views. Although the missing views are filled with CRA by using part of samples with complete views, our proposed algorithm still demonstrates clear superiority. According to the experimental results, our model performs as the best on all multi-modal datasets (CUB, 3Sources-complete, Football and Politics), outperforming the second performer with $2.5\%$, $10.2\%$, $36.7\%$, $6.6\%$, respectively. On average, our method achieves $3.9\%$ higher performance than the second performers on multi-feature datasets (Animal and Handwritten), and $14.0\%$ on all multi-modal datasets (CUB, Politics, Football, 3Sources-complete), which steadily verifies the superiority of our model. Our model also performs as the best on real-world partial multi-view data (ADNI and 3Sources-partial), achieving $3.4\%$ and $2.3\%$ higher performance than the second performer, respectively.  

We visualize the representations from different methods on Handwritten to investigate the improvement of CPM-Nets. As shown in Fig.~\ref{fig:tsne}, the subfigures (a)-(c) show the results of  representations obtained in an unsupervised manner. As can be observed, the latent representation from our algorithm reveals the underlying class distribution much better. With introducing label information, the representation from CPM-Nets are further improved, where the clusters are more compact and the margins between different classes become more clear. This validates the effectiveness of using clustering-like loss. It is worth noting that we jointly exploit all samples and views for random view-missing patterns in experiments, demonstrating the flexibility in handling partial multi-view data, while Fig.~\ref{fig:tsne} supports the claim of structured representation.

\begin{figure}[ht]
\center
\includegraphics[width=2.5in,height = 1.5in]{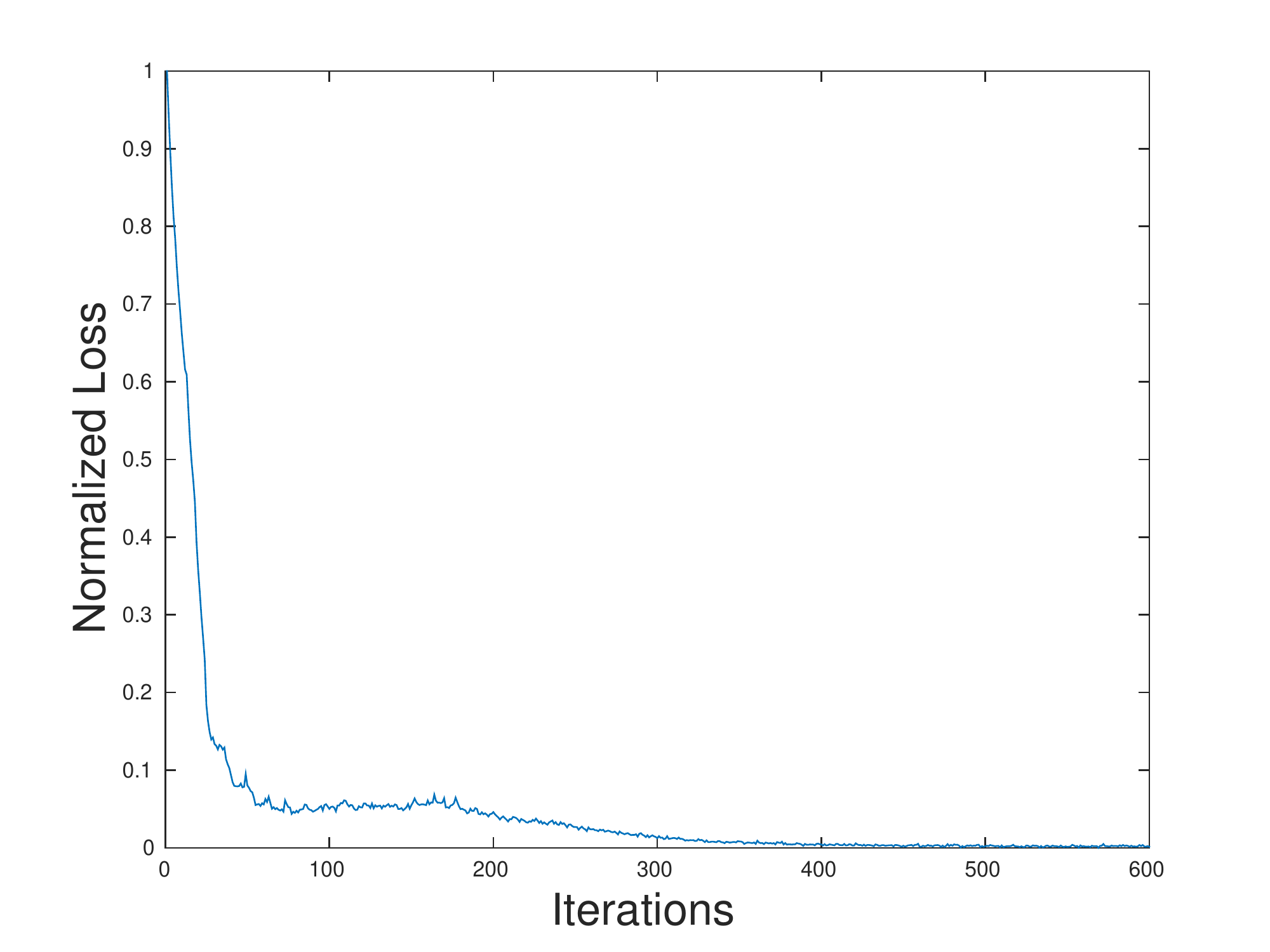}
\caption{Convergence experiment.}
\label{fig:Convergence curve}
\end{figure}

\subsubsection{Parameter Tuning and Convergence}
There are two main parameters in our algorithm, \emph{i.e.}, the tradeoff factor ($\lambda$) and the dimensionality ($K$) of the latent representation. We visualize the parameter tuning on Handwritten (with missing rate $\eta = 0.5$) as shown in Fig.~\ref{fig:parameter}. We tune the parameters from fine-grained sets, \emph{i.e.},  selecting $\lambda$ from $\{0, 0.1, ..., 0.9, 1, 2, ..., 10\}$ and the dimensionality ($K$) from the set $\{16, 32, 64, 128, 256\}$. Five-fold cross-validation is employed with the average values of accuracy reported. It can be observed that: (1) when $\lambda$ = 0 which corresponds to that the label information is not used, it leads to very poor performance; (2) as $\lambda$ gradually increases, the overall trend becomes better, which implies the role of supervisory information. Our model is not sensitive to the trade-off factor. For example, when the dimensionality (K) is fixed as 128, the performance is very promising with $\lambda$ in range 0.6-3; (3) differences between with and without supervisory information can also be found in Fig.~\ref{fig:tsne} (c) and (f), where with given labels the intra-class representation is more compact, and inter-class margin is more obvious.

For relatively small datasets, the model parameters of CPM-Nets can be updated by processing the whole training data. For large datasets, it can be solved through mini-batch training. As shown in Fig.~\ref{fig:Convergence curve}, we empirically investigate the convergence property on the dataset Animal with mini-batch gradient descent. For each mini-batch, 10\% of samples from each class are used.

\subsubsection{Fine-tuning evaluation}
The reason for obtaining fine-tuning is that labels  are used to train encoding networks, while in testing phase labels are not available. Thus there will be a gap between these two phases. To validate the effect of the fine-tuning, we conduct experiments on Handwritten and CUB to compare performance between with- and without- fine-tune procedure. According to Fig.~\ref{fig:finetune}, the performance with fine-tuning will be improved over without fine-tuning for most cases.

\subsection{Unsupervised Experimental Results}
To investigate the proposed \textbf{CPM-GAN} for view-missing data, we compare it with four baselines as follows: (1) \textbf{Average} \cite{average}  simply imputes missing parts with average value of all samples in each view; (2) \textbf{SVD} \cite{hotelling1936relations} is a matrix completion method by iterative soft thresholding of singular value decomposition; (3) \textbf{CRA} \cite{tran2017missing} is composed of a set of stacked residual autoencoders, which can learn complex relationship among data from different modalities; (4) \textbf{CPM-without-GAN} provides the comparative version of the CPM-Nets without adversarial strategy.

For our \textbf{CPM-GAN}, the fully-connected networks of CPM-Nets can be regarded as generators. As for the discriminators, for simplicity, we use the same structure as the generators. For the purpose of discrimination, a sigmoid layer is imposed on the output layer of each discriminator network. We note that promising performance can be expected with relatively shallow networks for most datasets, \emph{i.e.}, the number of network layers is set as $2 \sim 4$ for all datasets. We select the dimensionality ($K$) of the latent representation from $\{64, 128, 256\}$. To be fair, the experimental settings for both \textbf{CPM-GAN} and  \textbf{CPM-without-GAN} are the same. We run each method 10 times and report the average values and standard deviations.

\subsubsection{Imputation Performance}
Our model can jointly learn latent representations and perform missing view imputation. To validate the effectiveness of imputation, we run different methods under different missing rates, where the views are abandoned at random. We evaluate the imputation performance in terms of Normalized Root Mean Square Error (NRMSE) \cite{hotelling1936relations}.
The imputation performance under different missing rates is shown in Fig.~\ref{fig:curves}. The following observations can be drawn from the experiments: (1) When the ratio of missing views increases, the imputation performance of all methods consistently declines; (2) CPM-GAN outperforms all the comparative methods with all missing rates on all six datasets, validating the effectiveness of our model. (3) CPM-GAN consistently outperforms CPM-without-GAN on all datasets, which empirically validates the motivation of adversarial strategy. (4) According to the proposition 2.1, the latent representations will be more complete and versatile due to the more promising imputation results.

\subsubsection{Clustering Performance}
Note that, different from the standard GAN whose inputs for the generator are sampled from a fixed distribution and do not change, the inputs (a.k.a latent representations) in our CPM-GAN are dynamically updated to encode complementary information. To evaluate the latent representations, we conduct clustering with k-means on six datasets. The metrics Accuracy (ACC) and Normalized Mutual Information (NMI) are used to measure the performance.

According to Fig.~\ref{fig:clustering}, we have the following observations: (1) In terms of both ACC and NMI, CPM-GAN and CPM-without-GAN both achieve relatively promising performance compared with all baselines, validating the effectiveness of CPM-Nets. (2) Although simple, directly averaging the observed values for missing data is unreasonable and risky, since the performance from `Average' is quite unpromising; (3) Compared with CPM-without-GAN, CPM-GAN consistently performs better, especially under high missing rate. We also note that CPM-without-GAN sometimes slightly outperforms CPM-GAN under low missing rate (on Football or CUB). A possible reason is that the latent representation from CPM-Nets is stable under low missing rate. While as the missing rate increases, adversarial strategy improves the imputation and thus the latent representation as well;  (4) Although the baselines can also achieve promising performance at low missing rate, clear performance degeneration can be observed under high missing rate. (5) Our model outperforms all compared algorithms on all multi-modal datasets. Taking the missing rate ($\eta$ = 0.5) for example, on average, our method improves the performance $3.36\%$ and $5.11\%$ over the second performers in terms of ACC and NMI, respectively. According to Fig.~\ref{fig:clustering-realworld}, our method still outperforms other comparison methods on the naturally partial multi-view datasets in terms of both ACC and NMI. The above observations further validate the advantages of CPM-Nets in encoding complementary information and the robustness of adversarial strategy.

We also conduct experiments on CUB and Animal (these two-view datasets are suitable for VIGAN) to compare the VIGAN \cite{shang2017vigan} with ours in terms of unsupervised clustering performance. VIGAN needs a part of complete-view samples to train an encoder, and thus we use $10\%$ complete data to train the VIGAN. Though using training data, it can be seen from Fig.~\ref{fig:vigan} that our model still outperforms VIGAN in terms of both Accuracy and NMI.

\section{Conclusions}
We propose a novel algorithm for representation learning on partial multi-view data, which can jointly exploit all samples and views, and is flexible for arbitrary view-missing patterns. Our algorithm focuses on learning a complete and thus versatile representation to handle the complex correlations among multiple views. The common representation also endows flexibility for handling data with an arbitrary number of views and complex view-missing patterns, which is different from existing ad hoc methods. Equipped with a clustering-like classification loss, the learned representation is well structured, making the classifier interpretable. We empirically validate that the proposed algorithm is relatively robust to heavy and complex view-missing data. 

\section*{Acknowledgment}
This work was supported partially by National Natural Science Foundation of China (No. 61976151), the Natural Science Foundation of Tianjin of China (No. 19JCYBJC15200) and the National Key Research and Development Program of China (No. 2019YFB2101901).
{
\bibliographystyle{IEEEtran}
\bibliography{nips16}
}
\vspace{-15 mm}
\begin{IEEEbiography}[{\includegraphics[width=1in,height=1.1in,clip,keepaspectratio]{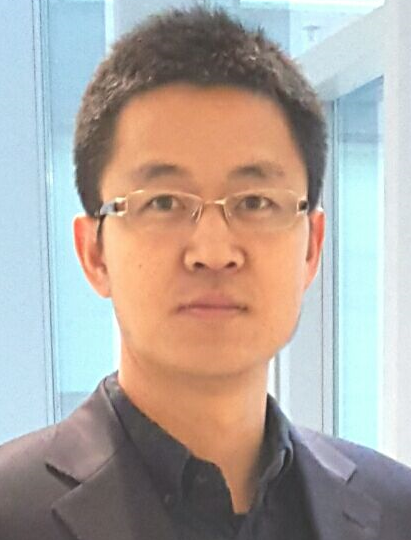}}]{Changqing Zhang}
received the B.S. and M.S. degrees from the College of Computer, Sichuan University, Chengdu, China, in 2005 and 2008, respectively, and the Ph.D. degree in Computer Science from Tianjin University, China, in 2016. He is an associate professor in the College of Intelligence and Computing, Tianjin University. He has been a postdoc research fellow in the Department of Radiology and
BRIC, School of Medicine, University of North Carolina at Chapel Hill, NC, USA. His current research interests include machine learning, computer vision and medical image analysis.
\end{IEEEbiography}

\vspace{-12 mm}
\begin{IEEEbiography}[{\includegraphics[width=1in,height=1.1in,clip,keepaspectratio]{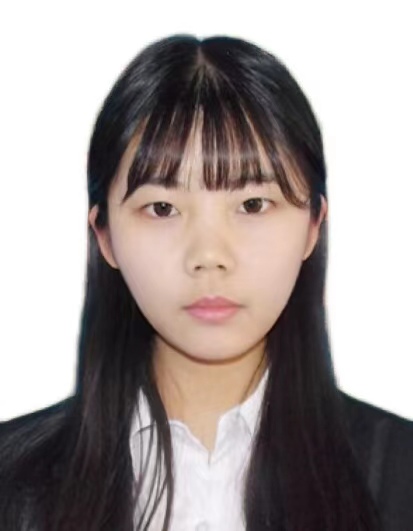}}]{Yajie Cui} received the B.S. degree in computer science and technology from Northeastern University, Qinhuangdao, China, in 2018. She is currently working towards the M.S. degree in the College of Intelligence and Computing, Tianjin University, Tianjin, China. Her research interests include multi-view representation and machine learning.
\end{IEEEbiography}


\begin{IEEEbiography}[{\includegraphics[width=1in,height=1.1in,clip,keepaspectratio]{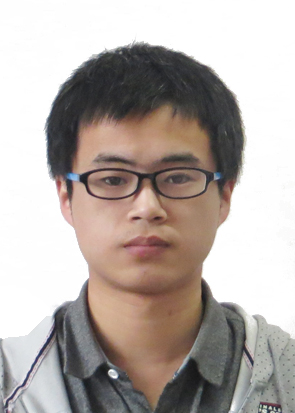}}]{Zongbo Han} received the B.S. degree in electronics and communication engineering from Dalian University of Technology, Dalian, China, in 2019. He is currently working toward the M.S. degree in the College of Intelligence and Computing, Tianjin University, Tianjin, China. His research interests include machine learning and computer vision.
\end{IEEEbiography}

\vspace{-16 mm}
\begin{IEEEbiography}[{\includegraphics[width=1in,height=1.1in,clip,keepaspectratio]{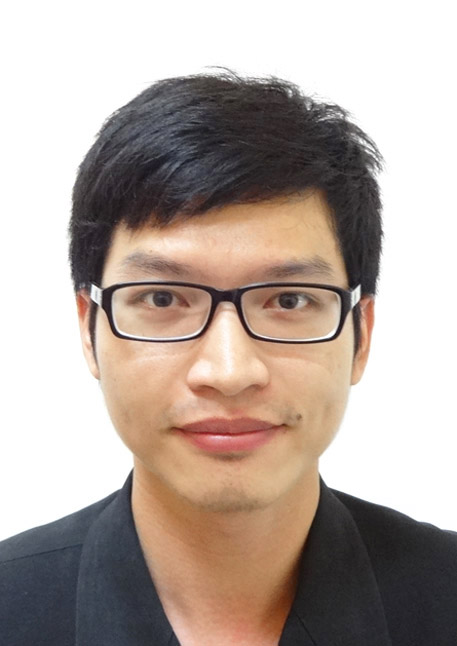}}]{Joey Tianyi Zhou} is currently a scientist, PI and group manager with the Institute of High Performance Computing (IHPC) in Agency for Science, Technology, and Research (A*STAR), Singapore. He is currently leading the AI Group  with more than 30 research staff members.  He is  also holding an adjunct faculty position in National University of Singapore (NUS).   Before working in IHPC, he was a senior research engineer with SONY US Research Center in San Jose, USA.  Joey Zhou received a Ph.D. degree in computer science from Nanyang Technological University (NTU), Singapore. His current interests mainly focus on low-resource machine learning and their applications in natural language processing and computer vision tasks.

Dr. Zhou co-organized ACML'16 workshop on Learning on Big Data workshop and IJCAI'19 workshop on Multi-output Learning, ICDCS'20 workshop on Efficient AI; has served as an Associate/Guest Editor for IEEE Access, IET Image Processing,  IEEE Multimedia, and ACM Transactions on Multimedia Computing, Communications, and Applications (TOMM), Springer Nature Computer Science and TPC Chair in Mobimedia 2020; and received NIPS Best Reviewer Award in 2017.
\end{IEEEbiography}
\vspace{-13 mm}
\begin{IEEEbiography}[{\includegraphics[width=1in,height=1.1in,clip,keepaspectratio]{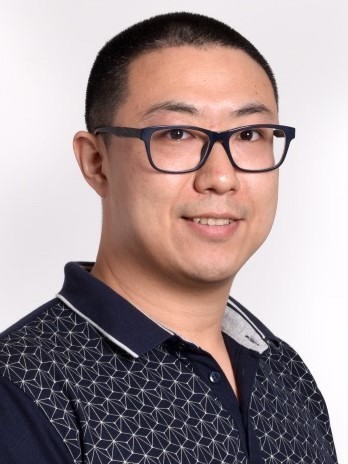}}]{Huazhu Fu} (SM'18) is a Senior Scientist at Inception Institute of Artificial Intelligence, Abu Dhabi, United Arab Emirates. He received his Ph.D. from Tianjin University in 2013, and was a Research Fellow at Nanyang Technological University for two years. From 2015 to 2018, he was a Research Scientist in Institute for Infocomm Research at Agency for Science, Technology and Research. His research interests include computer vision, machine learning, and medical image analysis. He currently serves as an Associate Editor of IEEE TMI, IEEE JBHI, and IEEE Access. He also serves as a co-chair of OMIA workshop and co-organizer of ocular image series challenge (i-Challenge).
\end{IEEEbiography}

\vspace{-15 mm}
\begin{IEEEbiography}[{\includegraphics[width=1in,height=1.1in,clip,keepaspectratio]{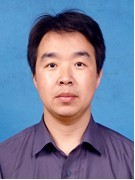}}]{Qinghua Hu} (SM'13) received the B.S., M.S., and Ph.D. degrees from the Harbin Institute of Technology, Harbin, China, in 1999, 2002, and 2008, respectively. He was a Post-Doctoral Fellow with the Department of Computing, Hong Kong Polytechnic University, Hong Kong, from 2009 to 2011. He is currently a Full Professor there. He has authored over 150 journal and conference papers in the areas of granular computing-based machine learning, reasoning with uncertainty, pattern recognition, and fault diagnosis. His current research interests include multi-modality learning, metric learning, uncertainty modeling and reasoning with fuzzy sets, rough sets and probability theory. Prof. Hu was the Program Committee Co-Chair of the International Conference on Rough Sets and Current Trends in Computing in 2010, the Chinese Rough Set and Soft Computing Society in 2012 and 2014, and the International Conference on Rough Sets and Knowledge Technology and the International Conference on Machine Learning and Cybernetics in 2014, and the General Co-Chair of IJCRS 2015. 
\end{IEEEbiography}

\end{document}